\documentclass{article}

\PassOptionsToPackage{numbers}{natbib}
\usepackage[preprint]{neurips_2023}

\usepackage[utf8]{inputenc} 
\usepackage[T1]{fontenc}    
\usepackage{hyperref}       
\usepackage{url}            
\usepackage{booktabs}       
\usepackage{amsfonts}       
\usepackage{nicefrac}       
\usepackage{microtype}      
\usepackage{xcolor}         
\usepackage[pdftex]{graphicx}
\usepackage{subcaption}
\usepackage{array}
\usepackage{amsmath}
\usepackage{float}
\usepackage{cleveref}
\usepackage{multirow}
\usepackage{arydshln}

\hypersetup{
  colorlinks   = true, 
  urlcolor     = blue, 
  linkcolor    = blue, 
  citecolor   = red 
}

\newcolumntype{M}[1]{>{\centering\arraybackslash}m{#1}}
\newcommand{\bb}[1]{\boldsymbol{#1}}
\newcommand{\repo}[0]{\url{https://github.com/wiskott-lab/classification-reconstruction-encoder}}

\title{Classification and Reconstruction Processes in Deep Predictive Coding Networks: Antagonists or Allies?}

\author{%
  Jan Rathjens \\
  Institute for Neural Computation\\
  Faculty of Computer Science\\
  Ruhr University Bochum\\
  44801 Bochum, Germany \\
  \texttt{jan.rathjens@rub.de} \\
  \And
  Laurenz Wiskott \\
  Institute for Neural Computation\\
  Faculty of Computer Science\\
  Ruhr University Bochum\\
  44801 Bochum, Germany \\
  \texttt{laurenz.wiskott@rub.de} \\
}

\begin{document}

\maketitle
\begin{abstract}
Predictive coding-inspired deep networks for visual computing integrate classification and reconstruction processes in shared intermediate layers. Although synergy between these processes is commonly assumed, it has yet to be convincingly demonstrated. In this study, we take a critical look at how classifying and reconstructing interact in deep learning architectures. Our approach utilizes a purposefully designed family of model architectures reminiscent of autoencoders, each equipped with an encoder, a decoder, and a classification head featuring varying modules and complexities. We meticulously analyze the extent to which classification- and reconstruction-driven information can seamlessly coexist within the shared latent layer of the model architectures. Our findings underscore a significant challenge: Classification-driven information diminishes reconstruction-driven information in intermediate layers' shared representations and vice versa. While expanding the shared representation's dimensions or increasing the network's complexity can alleviate this trade-off effect, our results challenge prevailing assumptions in predictive coding and offer guidance for future iterations of predictive coding concepts in deep networks.
\end{abstract}

\section{Introduction}
\label{sec:intro}
Predictive coding is a neuroscientific theory postulating how sensory perception is performed in the brain, with a particular emphasis on visual perception. In the traditional feedforward view of visual processing~\cite{hubel_receptive_1962}, visual stimuli are believed to be processed in a series of stages in a bottom-up fashion, starting from the retina and moving up through various brain areas, each responsible for recognizing increasingly abstract features of the visual input. On the other hand, in predictive coding~\cite{rao_predictive_1999}, each area predicts the activity of its preceding area along the visual pathway, transmitting these predictions through feedback connections in a top-down manner, while feedforward connections carry the difference between predicted and actual activity. Inference and learning are achieved by minimizing this difference. 

Deep learning networks for visual computing often adhere to the feedforward view of visual processing: Images are passed through successive layers in a feedforward manner~\cite{he_deep_2016, huang_densely_2017, tan_efficientnet_2019}. These networks have achieved remarkable success across various computer vision tasks. However, there is also an interest in networks that combine deep learning with predictive coding concepts into so-called deep predictive coding networks (DPCNs)~\cite{lotter_deep_2017, wen_deep_2018, huang_neural_2020, han_deep_2018} for various reasons. Firstly, deep learning networks may unlock the full potential of predictive coding due to their scalability, efficiency, and modularity – properties absent in standard computational models of predictive coding~\cite{hosseini_hierarchical_2020}. Secondly, incorporating predictive coding concepts into deep learning models can potentially transfer desirable features of human visual processing, such as robustness to noise, sample efficiency, and generalization, to deep learning applications~\cite{geirhos_comparing_2018}.
    
Although mathematical formulations of predictive coding concepts exist~\cite{rao_predictive_1999, friston_learning_2003}, DPCNs do not rigorously follow these formulations. Instead, they take loose inspiration from predictive coding concepts and differ in the extent of incorporation~\cite{hosseini_hierarchical_2020}. Nevertheless, a shared characteristic among deep predictive coding networks is the presumed ability to synergistically integrate classification and reconstruction processes into shared intermediate layers. For instance, Lotter et al.~\cite{lotter_deep_2017} argued that their proposed architecture, PredNet, learns useful representations for classification by predicting subsequent video frames. Subsequently, Wen et al.~\cite{wen_deep_2018} and Huang et al.~\cite{huang_neural_2020} introduced DPCN variants concurrently classifying and reconstructing input images.

The hypothesis that classification and reconstruction processes might synergistically enhance performance in deep predictive coding networks was first challenged by an analysis performed on PredNet~\cite{rane_prednet_2020}. Rane et al. added a classification head to the network's topmost layer, shared between the image prediction and classification processes. They showed that incorporating the classification process led to a decline in the quality of the predicted images, while the classification accuracy remained inferior to that of comparable feedforward networks. These findings imply that the classification and reconstruction processes in deep predictive coding networks may not work synergistically but antagonistically, as they negatively impact each other's performance.

In this study, we explore whether the antagony observed in PredNet is unique to its architecture or if such phenomena could also be observable in different DPCNs. Rather than examining other DPCN variants directly, our investigation is grounded on the premise that for DPCNs to achieve synergistic integration of these processes, their intermediate layers must effectively combine classification- and reconstruction-driven information using tools from deep learning, ideally, in a synergistic fashion.

We employ a purposefully designed model architecture that combines an autoencoder with an additional classification head to analyze the dynamics of integrating classification and reconstruction-driven information into a shared intermediate layer. The architecture enables the encoder to be trained to balance classification-driven and reconstruction-driven features. To ensure a comprehensive examination of potential dynamics, our methodology utilizes multiple variations of the model architecture. Each variant incorporates different deep learning tools in various complexities, including fully connected networks, convolutional networks, and vision transformer modules.

The studies on our model architecture reveal a significant challenge in deep learning: Integrating classification and reconstruction-driven information into a shared representation is only possible to a certain extent. Our findings challenge prevailing assumptions in deep predictive coding networks and point to potential future research directions to effectively combine predictive coding principles with deep learning.

\section{Background and Related Work}
\label{sec:background}
\subsection{Predictive Coding}
While the roots of predictive coding trace back to Helmholtz's concept of unconscious inference~\cite{helmholtz_concerning_1866}, the predictive coding framework was formally articulated by Rao and Ballard~\cite{rao_predictive_1999} and later reinterpreted as evidence maximization within hierarchical Gaussian generative models~\cite{friston_learning_2003, friston_theory_2005}. In this framework, an (artificial) neural network operates as a hierarchical generative model characterized by a unique flow of information: Each layer constantly transmits predictions of its preceding layer's neural activity while concurrently receiving feedback on the errors of its predictions. Both inference and learning within this framework occur through minimizing the difference between predicted and actual neural activity. Networks following this predictive coding framework are called predictive coding networks (PCNs).

\begin{figure}[ht]
    \centering
    \includegraphics[width=.8\linewidth]{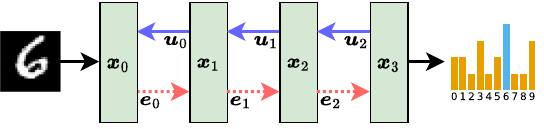}
    \caption{Inference in a PCN. Sensory stimulus predictions are hierarchically generated at the top layer and transmitted to the bottom via feedback connections. Prediction errors travel from the bottom to the top layer through feedforward connections. Each layer iteratively adjusts its values to minimize the sum of all prediction errors.}
    \label{fig:pcn}
\end{figure}

The information flow within PCNs in the context of classification is exemplified in \Cref{fig:pcn}. The neural activity of each PCN's layers $\ell$ is represented by a vector $\bb{x}_\ell$. The bottom layer activity $\bb{x}_0$ represents an input image or its preprocessed version, whereas the top layer activity $\bb{x}_3$ indicates the input image's class label, for instance, as a one-hot encoded vector. Every layer $\ell$ transmits a prediction of its preceding layer's activity $\bb{u}_{\ell - 1} = \bb{W}_{\ell} \phi(\bb{x}_{\ell})$ to its preceding layer $\ell - 1$. In parallel, each layer transmits the prediction error, defined as the difference between predicted and current activity $\bb{e}_\ell = \bb{x}_\ell - \bb{u}_\ell$, to its succeeding layer. 

The operating principle of the PCN is to determine configurations for all $\bb{x}_\ell$ such that the total squared prediction error, $\mathcal{E} = \frac{1}{2} \sum_{\ell}\bb{e}_{\ell}^2$, commonly referred to as the network's energy function, is minimized. This minimization is typically executed through methods like gradient descent. The network incrementally adjusts its configuration in a parallel manner until it reaches a state of minimal error, thereby determining the network's inferred state. 

PCNs exhibit flexibility in their application. For image generation, the top layer activity $\bb{x}_3$ is fixed to a particular class label, allowing the network to optimize its energy function accordingly, resulting in the generation of an image at $\bb{x}_0$. Conversely, for classification, the bottom layer activity $\bb{x}_0$ is fixed to an input image or its preprocessed form. Again, the network optimizes its energy function, yielding the class label at $\bb{x}_3$. Thus, a PCN essentially infers the class label by reconstructing the input image. 

Training the PCN for both functions simultaneously is achieved by fixing the bottom and top layers' activities, letting the network converge to a minimum configuration, and then running gradient descent on the network's energy function with respect to its weight matrices $\bb{W}$ \cite{millidge_2022_predictive_survey}.

The PCN in \Cref{fig:pcn} is an example of a forward PCN, characterized by prediction flow from the class label to the input image. However, there are alternative setups for PCNs in classification contexts. For instance, backward PCNs operate oppositely, with predictions moving from the input image to the class label~\cite{whittington_approximation_2017, sun_predictive-coding_2020}. Nevertheless, forward PCNs are considered to be more intuitive~\cite{millidge_2022_predictive_survey} and serve as the main inspiration for DPCNs, particularly when utilized for inference.

\subsection{Deep Predictive Coding Networks}
DPCNs take inspiration from the predictive coding framework and implement it in varying degrees with tools from deep learning~\cite{hosseini_hierarchical_2020}. In alignment with PCNs, DPCNs maintain a dynamic state or representation at each layer facilitated by recurrent connections. They incorporate top-down and bottom-up processing, aiming to reconstruct an image at the lowest layer and classify it at the highest layer. However, unlike PCNs, DPCNs typically employ sequential updates of representations through alternating top-down and bottom-up sweeps instead of parallel updates. Furthermore, DPCNs often relax the constraint of learning solely from locally available information, instead utilizing end-to-end training via backpropagation through time.

A notable early example of a DPCN was presented by Lotter et al.~\cite{lotter_deep_2017}. Their proposed architecture PredNet, designed for next-frame video sequence prediction, was the first to incorporate complex deep learning modules such as convolutional LSTM units into the predictive coding framework. The network focuses on minimizing the absolute error between the predicted and actual next frames. The architecture has the flexibility to be trained to minimize all prediction errors or to focus solely on the bottom-most error for reconstruction. Although not explicitly trained for classification, the representations learned by this network are argued to be transferable to classification tasks.

Wen et al.~\cite{wen_deep_2018} specifically designed a DPCN for image classification, drawing on the information processing principles of predictive coding. Interestingly, this DPCN focuses solely on classification training without explicitly aiming to minimize prediction errors. Despite this, the authors observed that tying weights between the top-down and bottom-up processes could facilitate accurate reconstruction of the input image. Moreover, even without tied weights, the network's optimization process led to reconstructing the bottom-layer image. This approach was further refined in subsequent research~\cite{han_deep_2018}, aligning the information flow within the network more closely with standard PCNs.

Huang et al.~\cite{huang_neural_2020} introduced a DPCN designed to concurrently optimize reconstruction error at the bottom layer and classification performance at the top layer. While this network integrates feedforward and feedback mechanisms, it notably does not incorporate the concept of prediction errors among its intermediate layers. As a result, the forward connections in this DPCN handle a diverse spectrum of information, going beyond the conventional focus on prediction error typically seen in PCNs.

\subsection{Synergy and Antagony}
Analogously to PCNs, one rationale behind developing DPCNs is the belief that classification and reconstruction processes can mutually benefit each other, creating a synergistic effect. Along these lines, the literature has documented instances where beneficial interactions between these two processes have been observed in PCNs, DPCNs, and deep learning in general. However, there are also reported cases where these processes exhibit antagonistic effects.

Predictive coding posits that learning to predict sensory input is pivotal for developing rich representations of the world~\cite{millidge_2022_predictive_survey}. The prediction conceptualized as input reconstruction, alongside the notion that a rich representation is inherently classifiable, indicates synergy between reconstruction and classification. In PCNs applied to classification, class labels are inferred by predicting or reconstructing input images, reinforcing this synergistic view.

Nevertheless, Sun et al.~\cite{sun_predictive-coding_2020} reported that predictive coding may face challenges in balancing these processes. Depending on the operational mode of the network, it may excel in either generation or classification but not both.

In the context of DPCNs, Lotter et al.~\cite{lotter_deep_2017} argued that such architectures can learn representations transferable to classification tasks. Wen et al.~\cite{wen_deep_2018} suggested that the predictability of input inherently enhances representation quality, as representations can be utilized for input reconstruction and classification. The authors argued that the iterative nature of the two processes yielded smaller yet accurate networks, potentially outperforming their larger counterparts. Huang et al.~\cite{huang_neural_2020} contended that the reconstruction process imparts an inductive bias, contributing to network robustness, while Alamia et al.~\cite{alamia_role_2023} indicated its utility in handling noise-perturbed images.

Conversely, Rane et al.~\cite{rane_prednet_2020} demonstrated that optimizing for classification could diminish the quality of predicted frames and vice versa. Sun et al~\cite{sun_predictive-coding_2020} criticized that the DPCN of~\cite{wen_deep_2018} did not effectively synthesize images, as the architecture was primed with an initial feedforward pass.

In broader deep learning applications, reconstruction is a common training technique for models in a variety of applications. Models are initially trained to reconstruct a specific form of input, such as images or sentences, tailored to the application. Subsequently, the model is adapted for downstream tasks like classification, with fine-tuning of weights to optimize task-specific performance.

In natural language processing, notable instances include masking-based reconstruction in BERT~\cite{devlin_bert_2019} and autoregressive reconstruction of input sequences in GPT~\cite{brown_language_2020}. He et al.~\cite{he_masked_2022} demonstrated the application of an autoencoder, based on a Vision Transformer (ViT)~\cite{dosovitskiy_image_2020}, trained to reconstruct masked image patches. These representations were found to be effectively transferable to classification tasks, especially under high masking ratios and fine-tuning scenarios, suggesting a synergy between reconstruction and classification.

Notably, the synergy between classification and reconstruction typically assumed in deep learning diverges from that in PCNs or DPCNs. While reconstruction may be utilized for pretraining, it is negligible during inference. In contrast, PCNs and DPCNs explicitly reconstruct an image as a crucial step in inferring its class, integrating reconstruction directly into the inference mechanism.

\section{Methods}
\label{sec:methods}
Our methods are grounded in the understanding that intermediate layers' representations in PCNs and DPCNs are simultaneously utilized for two tasks. Firstly, these representations carry information to classify the input image at the top layer. Secondly, they carry information for reconstructing an input image at the bottom layer. Hence, unrelated to implementation details of PCNs or DPCNs, these networks integrate classification- and reconstruction-driven information into shared intermediate layers. 

Several possibilities arise when merging the two types of information within shared representations:

\begin{itemize}
    \item Integration could lead to a synergistic effect wherein reconstruction-driven information positively impacts classification accuracy and vice versa. For example, the reconstruction process could function as a regularizer during training or offer valuable features for more effective classification. At the same time, classification-driven information could provide meaningful information for the reconstruction process. 
    \item The integration may result in a competitive or antagonistic relationship between the two kinds of information. Here, the dominance of one type adversely affects the performance of the other. For instance, reconstruction-driven information might diminish classification accuracy and vice versa, suggesting a need to balance between these competing sources of information.
    \item The two kind of information could coexist in the shared representation without exerting substantial impact or interference on each other. This scenario implies a level of independence or separation between the two types of information within the shared representations.
\end{itemize}

While our study is primarily motivated by exploring the dynamics of integrating classification and reconstruction-driven information into shared representation in DPCNs, we do not to utilize DPCNs or their inspiration PCNs directly for several reasons. Firstly, PCNs are challenging to train effectively~\cite{kinghorn_preventing_2023}, as they often converge to suboptimal local optima, have computationally demanding training requirements~\cite{zahid_predictive_2023}, and are hard to scale~\cite{hosseini_hierarchical_2020}. Secondly, DPCNs are variable in their implementation, and the lack of a standardized or unique architecture for DPCNs complicates their use for consistent exploration in our context.

Given these challenges, we opt for a more foundational exploration, employing a purposefully designed model architecture based on tools from deep learning. The model architecture is designed and trained to explicitly integrate classification and reconstruction-driven information into a shared representation of an intermediate layer. Thus, our study yields results that go beyond the scope of DPCNs but are of particular interest to these networks.

\subsection{Model Architecture}
\begin{figure}[ht]
    \centering
    \includegraphics[width=.8\linewidth]{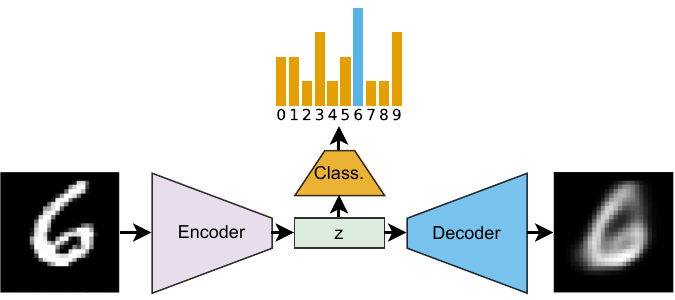}
    \caption{Classification-Reconstruction Encoder (CRE). The CRE comprises a decoder and a classifier connected to the latent representation $z$. The encoder can be optimized to encode an input image in either a classification- or reconstruction-driven manner.}
    \label{fig:mhe}
\end{figure}

Our model architecture is reminiscent of an autoencoder featuring an additional classification head connected to the latent variables. Without claiming originality, we refer to this architecture as a Classification Reconstruction Encoder (CRE); see \Cref{fig:mhe}. The classifier and the decoder share the representation of the hidden or latent $\bb{z}$-layer. In our study, this layer is of pivotal interest since the encoder must explicitly integrate classification and reconstruction-driven information into a shared representation here since the representation is utilized as input for both the classifier and decoder.

During training, the decoder is optimized to reconstruct images from $z$ by minimizing the mean squared error (MSE) loss $L_{\text{MSE}}$ between the reconstructed and the input images. The classifier is optimized to classify images from $z$ by minimizing the cross entropy (CE) loss $L_{\text{CE}}$ between the predicted and true labels of the images. The encoder's loss function $L$ is a weighted combination of the two losses:

\begin{equation}
    L = \lambda L_{\text{MSE}} + (1 - \lambda) L_{\text{CE}}, \quad \lambda \in [0, 1]
    \label{eq:encoder_loss}
\end{equation}

The encoder's role is to combine classification- and reconstruction-driven information within the shared $\mathbf{z}$-layer. We achieve a balance between these two types of information in $\bb{z}$ by adjusting the $\lambda$-value, scaling $L_{\text{CE}}$ and $L_{\text{MSE}}$. Setting $\lambda$ to 0 directs the encoder to optimize $\bb{z}$ solely for classification. In this configuration, the encoder and classifier resemble a conventional classification network. Consequently, the decoder is tasked with reconstructing the image from a representation driven purely by classification. In contrast, a $\lambda$-value of 1 steers the encoder to optimize $\bb{z}$ purely for reconstruction purposes, essentially transforming the encoder-decoder combination into a standard autoencoder. Here, the classifier is optimized for image classification based on a representation driven solely by reconstruction.

Notably, when $\lambda$ is either 0 or 1, the encoder is not trained to integrate these two types of information in $\mathbf{z}$. Nonetheless, these extreme values of $\lambda$ provide benchmark scenarios for evaluating the integration dynamics. Furthermore, they allow us to assess the transferability of representations, i.e., how well representations optimized for one task can be utilized by a different task. When $\lambda$ is set to a value between 0 and 1, the encoder integrates classification and reconstruction information into $\mathbf{z}$. The degree of emphasis on either $L_{\text{CE}}$  or $L_{\text{MSE}}$ is modulated based on the value of $\lambda$. As a result, the encoder variably prioritizes one type of information over the other, contingent on the specific setting of $\lambda$.

In deep learning, various factors such as architecture design, dataset selection, hyperparameter tuning, and the choice of training algorithms can significantly influence how information is represented, whether driven by classification or reconstruction tasks. These factors may, therefore, also affect the integrability of the two kinds of information. To gain a comprehensive view of the integrability of classification- and reconstruction-driven information into shared representations in deep learning architectures, we base the CRE's modules, i.e., the encoder, decoder, and classifier, on different common deep learning architecture variants. These variants include a fully connected (FC)-based CRE, a convolutional neural network (CNN)-based CRE, and a Vision Transformer (ViT)-based CRE~\cite{dosovitskiy_image_2020}. 

Moreover, we implement diverse configurations within these modules. For instance, we modify the dimensions of the latent space and adjust the complexity of the modules by altering their layer or parameter count. A comprehensive description of all the models utilized is provided in the supplementary material and is accessible on GitHub\footnote{\repo}.
Finally, we train the models on different datasets. Namely, MNIST~\cite{lecun_gradient-based_1998}, FashionMNIST~\cite{xiao_fashion-mnist_2017}, and CIFAR-10~\cite{krizhevsky_learning_2009}.


We adopt the ViT-based CRE from the ViT-based autoencoder suggested by He et al.~\cite{he_masked_2022}. The original approach involved randomly masking a certain percentage of input patches and training the autoencoder to reconstruct these masked patches only. The masking process was shown to increase the transferability of reconstruction-driven representations to classification and is, thus, particularly interesting for our study. To build upon this, we introduce an additional masking variant, v2, alongside the original method (we now refer to as v1). V2 trains the CRE to reconstruct the entire input image. This modification in v2 allows for a more equitable comparison of reconstruction performance across different CRE variants. For both v1 and v2, we opted for a masking percentage of 66\%, as the authors showed that this ratio yields robust classification performance.

Furthermore, instead of fine-tuning weights for classification after training the model for reconstruction, we train all components (encoder, decoder, and classifier) simultaneously. In the original paper, the authors suggested two classification techniques that are similar in classification performance. These include appending a classification token to the latent representation and transmitting an averaged latent representation to the classifier. We opt for an averaging approach to confirm that classification- and reconstruction-driven information are coherently integrated.

To benchmark the quality of learned representations, we train a number of classifiers and decoders on the principal components (PCA) of the respective training datasets and on random but fixed linear projections (RP) of the training input.

The methodology for training our models is as follows: Each configuration of the CRE is trained with an Adam optimizer on a specific dataset using a range of values for the parameter $\lambda$ until convergence is achieved. Given that $\lambda$ is a continuous parameter, we strategically select values that span the most significant regions of its range. We choose values that broadly span the spectrum between the extreme values of $\lambda$, i.e., $0$ and $1$, and values close to these extremes. Specifically, the $\lambda$-values we use are $0.0$, $0.01$, $0.05$, $0.1$, $0.2$, $0.4$, $0.6$, $0.8$, $0.9$, $0.95$, $0.99$, and $1.0$. To ensure robustness and reliability in our findings, each CRE configuration is trained multiple times, ranging from five to ten runs. This approach yields a dataset comprising approximately 10,000 trained model instances, providing a solid foundation for evaluating the training efficacy and the impact of varying $\lambda$-values on the model performance.

\section{Results}
\label{sec:results}
\subsection{Performance Evaluation}
\begin{figure}[ht]
    \centering
    \includegraphics[width=1.\linewidth]{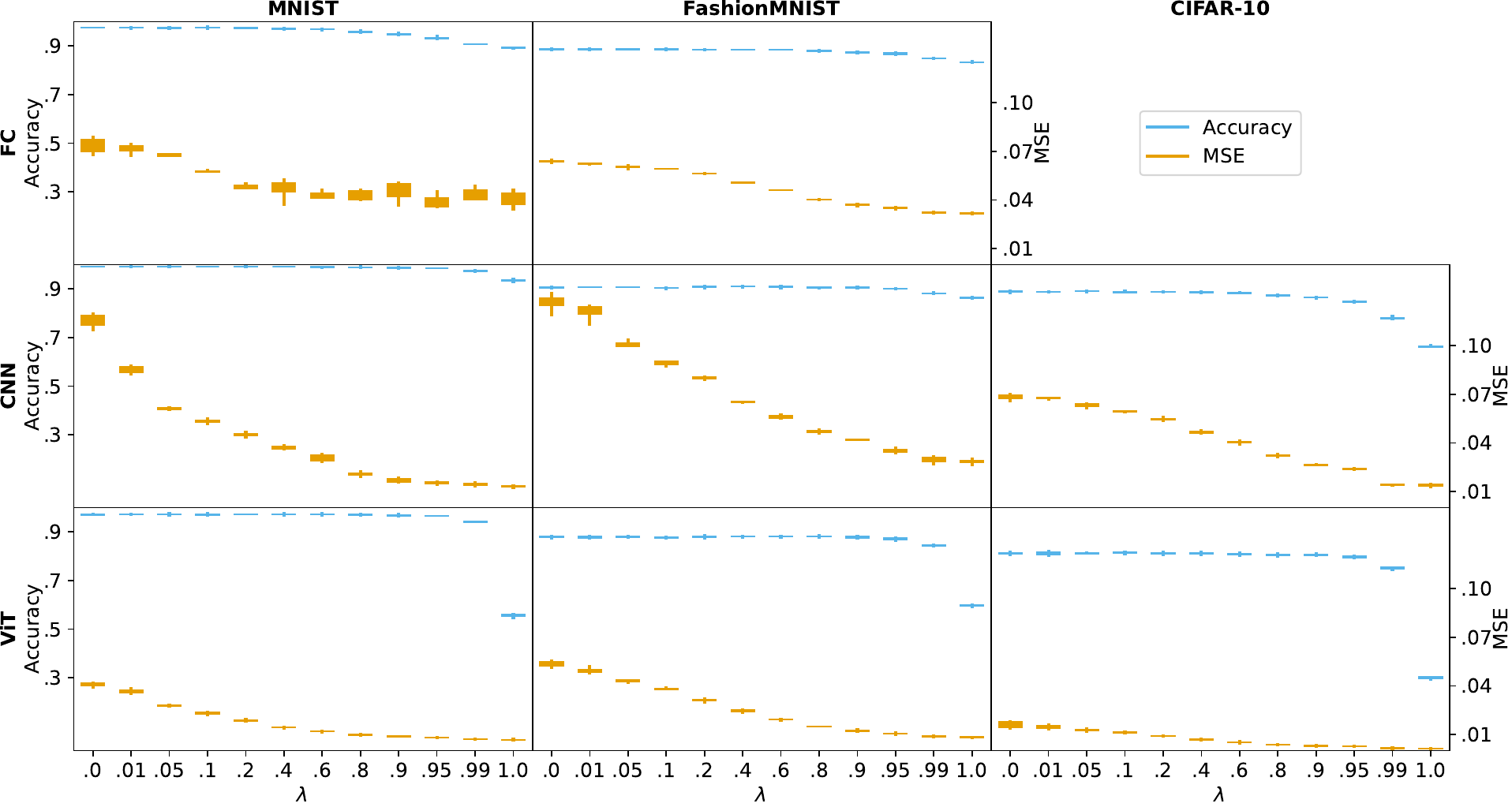}
    \caption{Performances w.r.t. $\lambda$. The box plots of the classification (blue) and reconstruction (orange) performance for different variants of trained CREs on several datasets are displayed.}
    \label{fig:combined_tradeoffs}
\end{figure}

\Cref{fig:combined_tradeoffs} showcases box plots illustrating the performance distributions across different $\lambda$-values for all three variants of the CRE with select configurations on various datasets. Notably, we did not train the FC-based variant on the CIFAR-10 dataset, as fully connected architectures face considerable challenges with this dataset. 

Our experimental results display a consistent pattern. The models reach peak classification and reconstruction performances at the extreme values of $\lambda$. Specifically, $\lambda = 0$ yields the best classification results, while $\lambda = 1$ leads to the most effective reconstruction. On the other hand, these extreme $\lambda$-values also correspond to the lowest performances for the opposite metric – poor reconstruction at $\lambda = 0$ and suboptimal classification at $\lambda = 1$.

We also note a monotonic relationship between $\lambda$-values and performance: As $\lambda$ increases, reconstruction performance improves, whereas classification performance deteriorates. The reconstruction performance of the FC-based CRE on the MNIST dataset slightly deviates from this relationship as the pattern is less consistent than in the other plots. 

Interestingly, classification performance significantly drops only when $\lambda$ is close to one, whereas the decrease in reconstruction performance is more linear. Hence, balanced performance for both classification and reconstruction is noted at $\lambda = 0.9$, though it is not optimal for either.

The manifestation of this pattern varies across different architectures and datasets. For instance, the CNN-based CRE shows a significant decrease in reconstruction performance as $\lambda$ nears zero, while the ViT-based CRE experiences a sharp decline in accuracy as $\lambda$ approaches one. Remarkably, these specific CRE-variants also demonstrate the best classification, respectively, reconstruction performance on all datasets.

The consistent pattern across architectures and datasets highlights a general trade-off between integrating classification and reconstruction information: Enhancing one aspect tends to weaken the other without synergistic gains. This trade-off is especially evident when transferring representations from one task to another, such as using purely reconstruction-driven representations for classification purposes and vice versa.

\subsection{Visual Analysis}
We visually examine the latent space representations of the CRE's $\mathbf{z}$-layer and sample reconstructions for distinct $\lambda$-values to understand the trade-off effect better.

\begin{figure}[ht]
\centering
    \begin{subfigure}[t]{0.28\linewidth}
        \centering
        \includegraphics[width=1.0\linewidth]{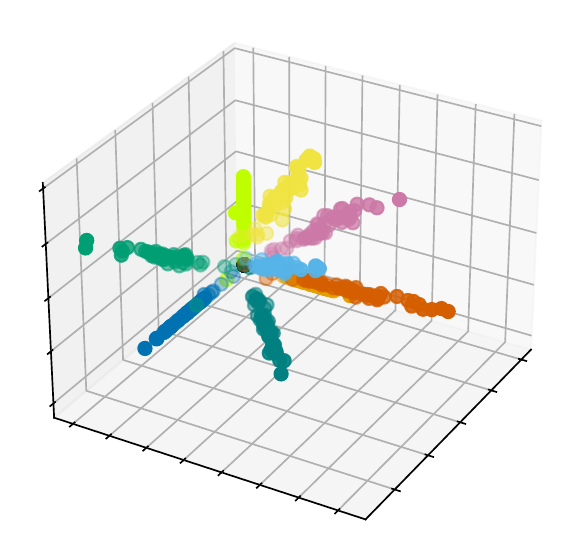}
        \includegraphics[width=1.0\linewidth]{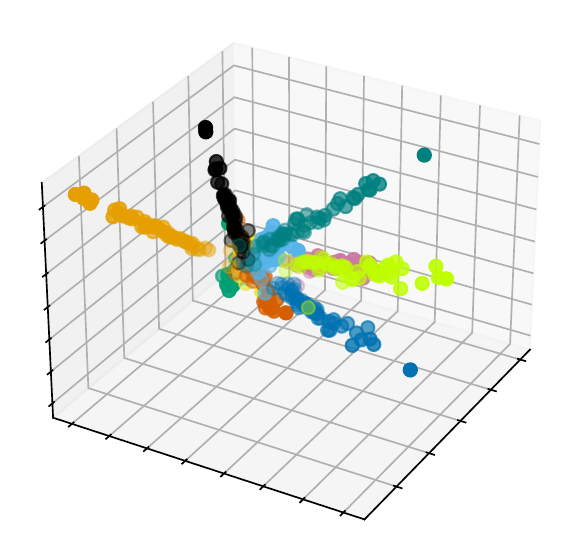}
        \caption{$\lambda = 0.0$}
        \label{fig:3d_00}
    \end{subfigure}
    \begin{subfigure}[t]{0.28\linewidth}
        \includegraphics[width=1.0\linewidth]{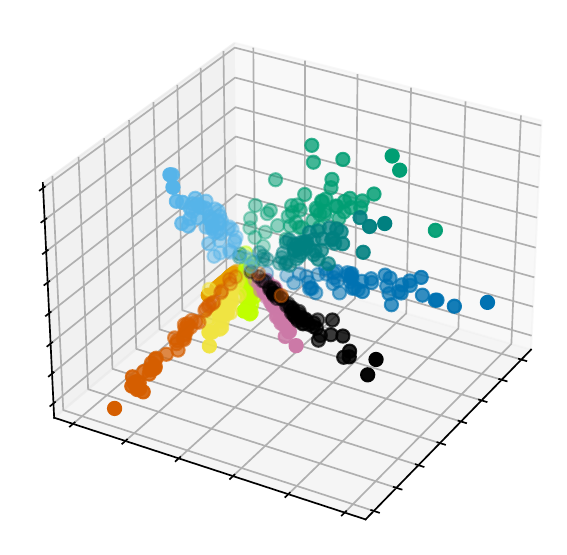}
        \includegraphics[width=1.0\linewidth]{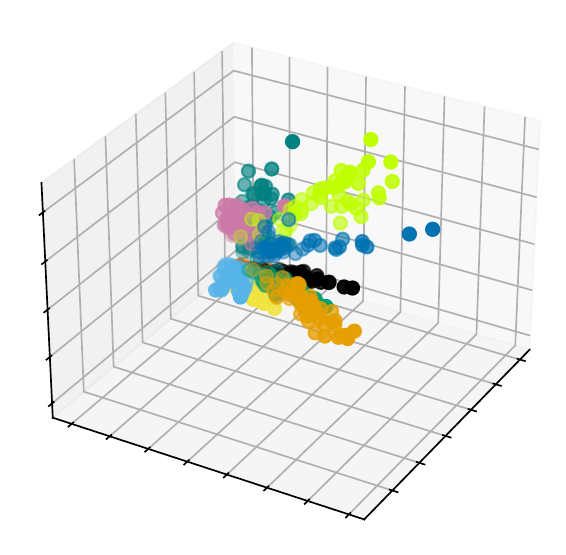}
        \caption{$\lambda = 0.2$}
        \label{fig:3d_02}
    \end{subfigure}
    \begin{subfigure}[t]{0.28\linewidth}
        \centering
        \includegraphics[width=1.0\linewidth]{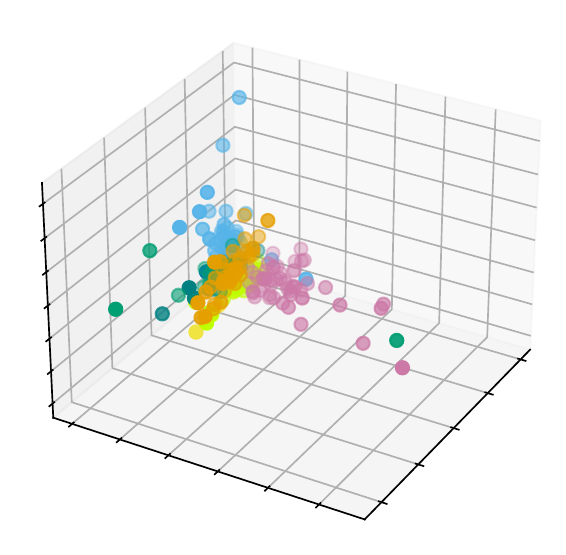}
        \includegraphics[width=1.0\linewidth]{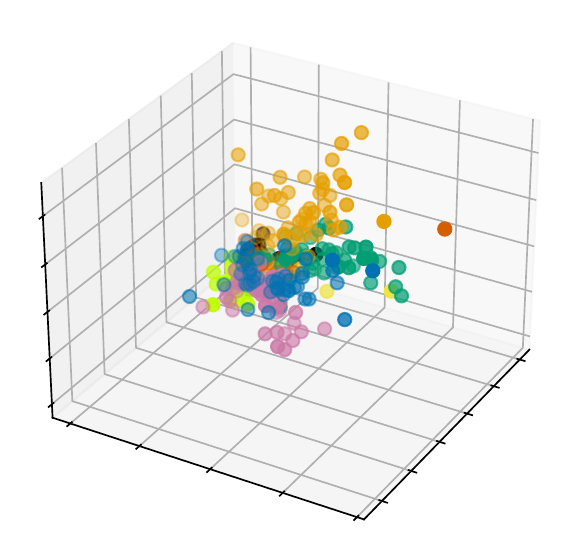}
        \caption{$\lambda = 1.0$}
        \label{fig:3d_10}
    \end{subfigure}
   \caption{Visualization of the latent space across different $\lambda$-values. Each plot displays 3D coordinates of input images in the latent space. An exemplary FC-based CRE for each $\lambda$-value with a 3-dimensional latent space was used to generate each plot. Each color represents one class. We depict fifty instances from each class. The top row showcases the MNIST dataset, and the bottom row showcases the FashionMNIST dataset.}
   \label{fig:latent_spaces}
\end{figure}

For a genuine visualization of the latent space, we select FC-based models with a latent space constrained to three dimensions. Figure~\ref{fig:latent_spaces} showcases typical latent space configurations for various $\lambda$-values, with each color denoting one of ten classes. The arrangement of the figure is such that the top row illustrates the latent spaces for the MNIST dataset, and the bottom row shows those of the FashionMNIST dataset. 

In both datasets, a consistent pattern is evident. Within the latent spaces for $\lambda = 0.0$., classes manifest as nearly straight lines emanating from a central point. The angular separation between points appears to be maximized, resembling a star-shaped configuration. As we move to $\lambda=0.2$, the overall structure remains similar to the $\lambda = 0.0$ scenario, with classes extending as straight lines from a central hub but with a noticeable increase in scatter among the points. In contrast, at $\lambda=1.0$, there is a significant shift in the spatial arrangement. Here, classes are no longer characterized by linear formations but instead appear as clusters, exhibiting varying degrees of overlap.

\begin{figure}[ht]
    \begin{subfigure}[t]{0.246\linewidth}
        \centering
        \includegraphics[width=1.0\linewidth]{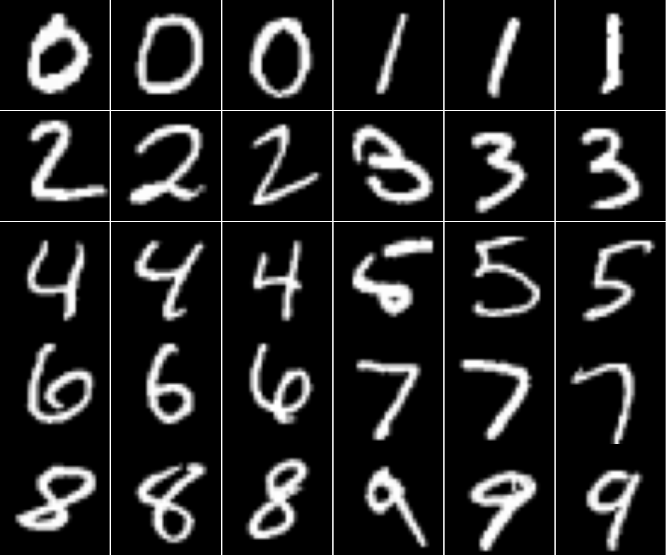}
        \includegraphics[width=1.0\linewidth]{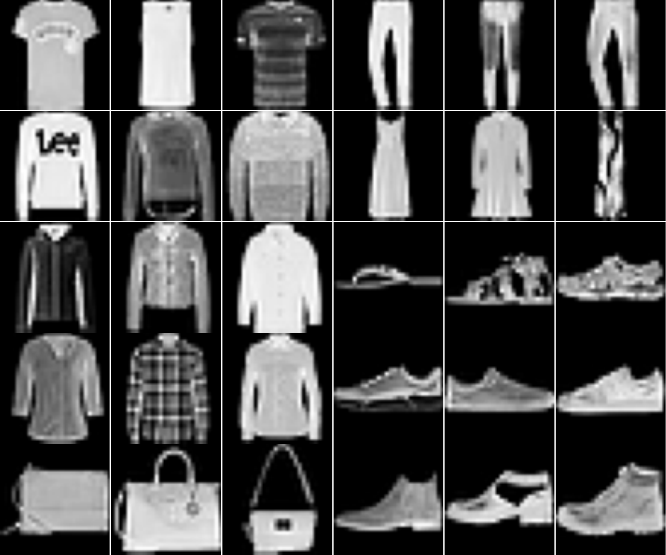}
        \caption{Input images}
        \label{fig:input_images}
    \end{subfigure}
    \begin{subfigure}[t]{0.246\linewidth}
        \includegraphics[width=1.0\linewidth]{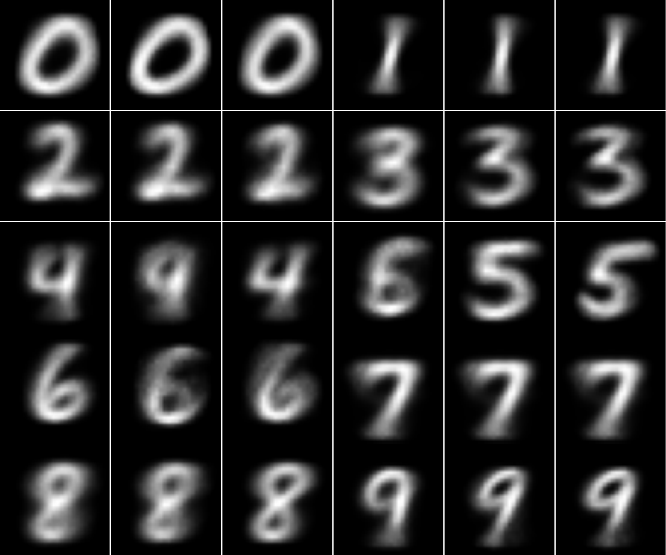}
        \includegraphics[width=1.0\linewidth]{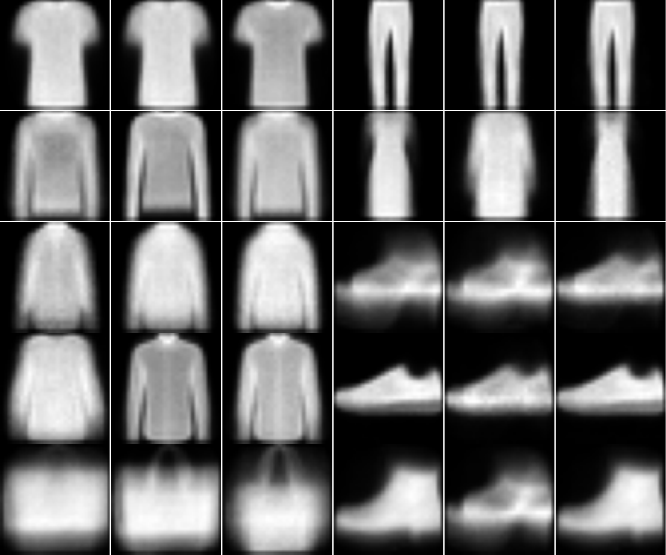}
        \caption{$\lambda = 0.0$}
        \label{fig:recons_00}
    \end{subfigure}
    \begin{subfigure}[t]{0.246\linewidth}
        \centering
        \includegraphics[width=1.0\linewidth]{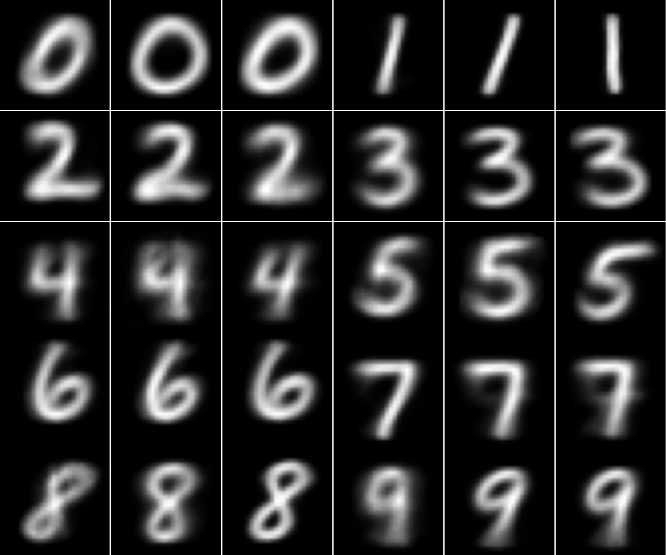}
        \includegraphics[width=1.0\linewidth]{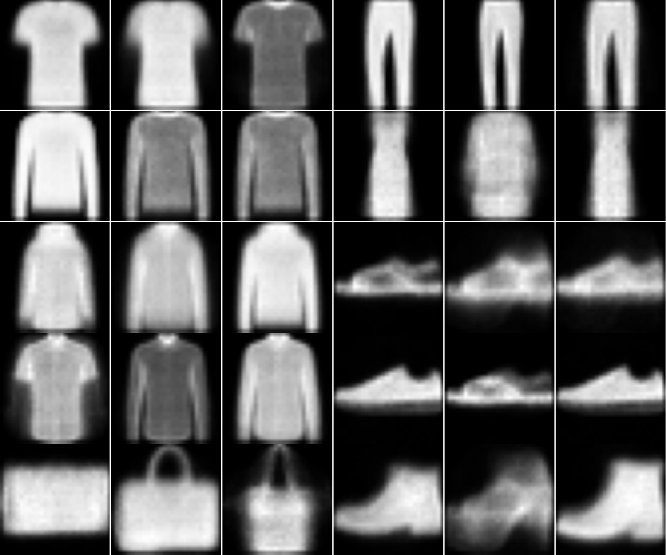}
        \caption{$\lambda = 0.2$}
        \label{fig:recons_02}
    \end{subfigure}
    \begin{subfigure}[t]{0.246\linewidth}
        \centering
        \includegraphics[width=1.0\linewidth]{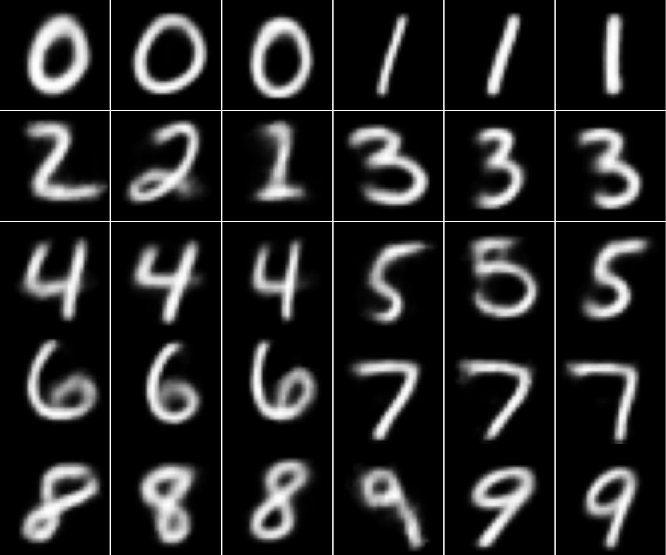}
        \includegraphics[width=1.0\linewidth]{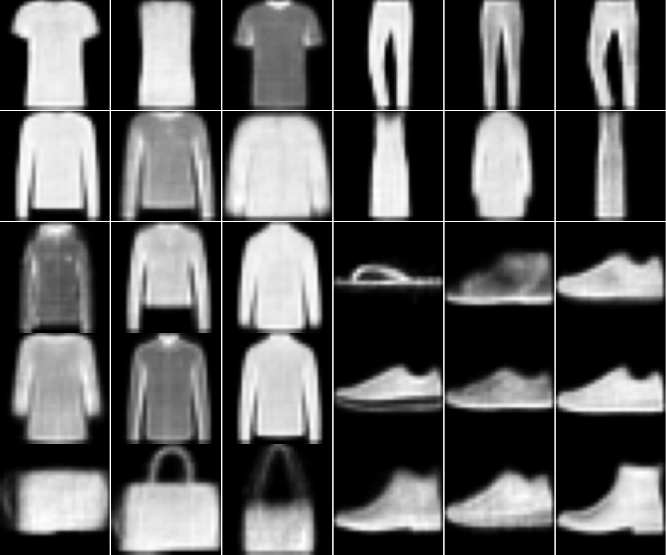}
        \caption{$\lambda = 1.0$}
        \label{fig:recons_10}
    \end{subfigure}
    \caption{Visualization of reconstructions across different $\lambda$-values. The left column shows the input images. The other columns display exemplary reconstructions w.r.t. their respective $\lambda$-value. An exemplary FC-based CRE for each $\lambda$-value with a 3-dimensional latent space was used to generate each plot. The top row showcases the MNIST dataset, and the bottom row showcases the FashionMNIST dataset.}
    \label{fig:reconstructions}
\end{figure}

\Cref{fig:reconstructions} presents side-by-side comparisons of input images and their corresponding reconstructions for the same models depicted in \Cref{fig:latent_spaces}.  At $\lambda = 0 $, the reconstructions for each class instance are similar, lacking the distinct details present in the input images; instead, prototypical examples for each input image are reconstructed. When $\lambda = 0.2$, some details of the input images are captured, while others are disregarded, resulting in a mix of prototypical and detailed reconstructions. At $\lambda = 1.0$, the models exhibit a significant improvement in capturing details.

Visualizing the latent space and sampling reconstructions offers insight into the trade-off effect. Firstly, classification- and reconstruction-driven prefer different configurations in a three-dimensional latent space. Classification-driven representations favor a star-shaped configuration, while reconstruction-driven representations prefer point clouds. Combining both preferences is only feasible to a certain extent, as the lines become less straight or the point clouds become less diverse. This trend is likely to extend to higher-dimensional latent spaces as well. Secondly, integrating classification-driven information into representations tends to result in a loss of visual detail. The decoder primarily reconstructs based on prototypical examples derived from classification-driven representations, leading to a loss in finer details.

\subsection{Alleviating the Trade-Off Effect}
It is plausible that increasing the CRE's complexity or expanding the latent space's dimensions may mitigate the observed trade-off effect. For example, by enhancing the capability of the classifier component, the model might sustain high classification accuracy even when representations are primarily reconstruction-driven. Similarly, increasing the dimensionality of $\bb{z}$ would give the encoder a larger 'canvas' for information encoding. This expansion may allow for a more effective combination of classification-driven and reconstruction-driven information. To explore this hypothesis, we compare the trade-off effect between CREs with constant size of $\bb{z}$ but varying complexities and CREs with fixed complexity but differing size of $\bb{z}$.

\begin{figure}[ht]
    \centering
    \includegraphics[width=1.0\linewidth]{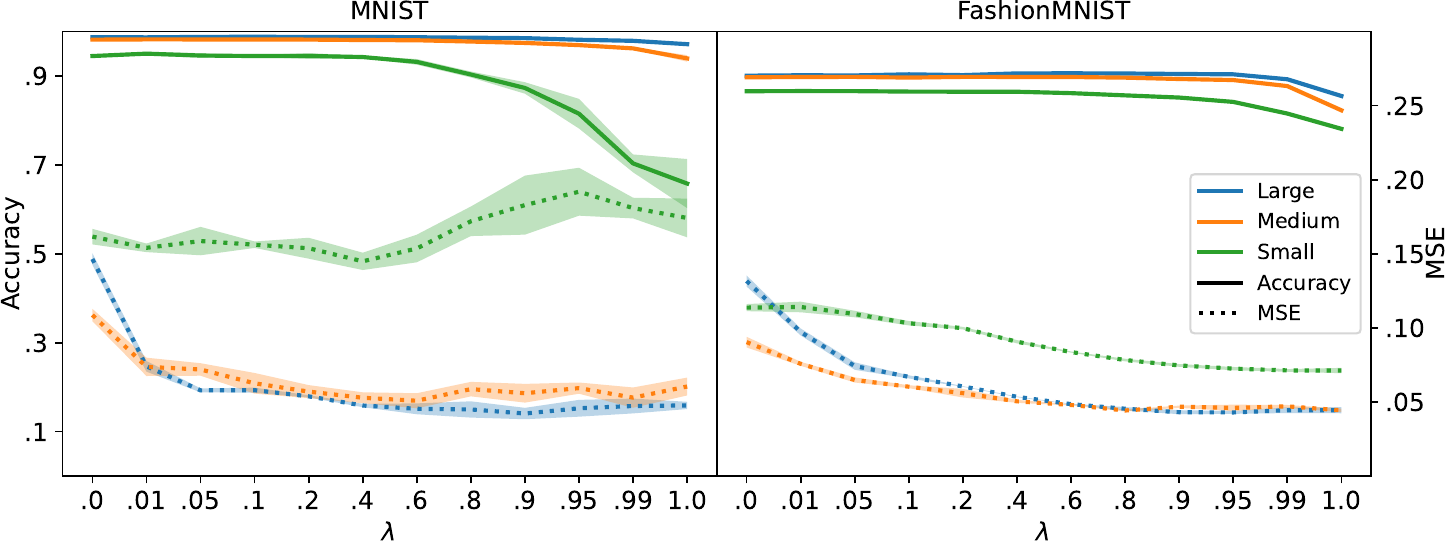}
    \caption{Trade-off effect by network complexity: Plots show the average classification (solid lines) and reconstruction (dashed lines) performances of FC-based CREs of varying complexities with their 95\% confidence interval. Complexity varies by layer count and parameter number. Green represents the least complex CRE, orange the medium, and blue the most complex. All have a 16-dimensional latent space. Left plot: MNIST; Right plot: FashionMNIST.}
    \label{fig:fc_complexities}
\end{figure}

\Cref{fig:fc_complexities} illustrates how varying network complexities influence the trade-off effect in FC-based CREs on the MNIST and FashionMNIST datasets. The large variant of the model incorporates four layers in both the encoder and the decoder and two layers in the classifier. The encoder and decoder in the medium variant consist of two layers, and the small variant further reduces this to one layer. The classifier in these versions is similarly adjusted. In the medium variant, the classifier consists of two layers, though the intermediate layer is reduced in size compared to the large variant. The classifier in the small variant consists of one layer.

In both MNIST and FashionMNIST datasets, with one exception, all CRE variants exhibit a trade-off effect aligning with results shown in \Cref{fig:combined_tradeoffs}. For classification, more complex networks, as expected, achieve higher accuracy at $\lambda=0$, though the difference between medium and large CREs is minimal. As $\lambda$ increases, a monotonous decrease in performance is observed across all CRE complexities while maintaining their relative performance order. However, the drop in accuracy is more pronounced or occurs at lower $\lambda$ values in less complex CREs. For instance, on the MNIST dataset, the large CRE only experiences a minor accuracy reduction at $ \lambda$-values close to one. The medium CRE shows a slightly more significant accuracy drop at similar $ \lambda$-values. The small CRE exhibits a substantial decrease in accuracy starting from around $\lambda = 0.6$.

The comparison of reconstruction performance across different complexities presents a more convoluted scenario than classification performance. Analogously to classification, at $\lambda = 1$, more complex models show superior reconstruction performance, which generally decreases as $\lambda$ decreases, with one exception. However, unlike classification, the relative performance rankings among complexities are inconsistent. For instance, in both MNIST and FashionMNIST datasets, the medium CRE outperforms the larger model at $\lambda = 0.0$. Furthermore, the small CRE in MNIST does not follow a monotonous decrease in performance with decreasing $\lambda$, contrasting with the more predictable pattern observed in classification.

This suggests that features optimal for classification and their respective architectural complexity do not invariably translate to superior reconstruction performance. Thus, even though more complex architectures theoretically possess greater capability, their effectiveness in reconstruction tasks is not guaranteed.

\begin{figure}[ht]
    \centering
    \includegraphics[width=1.0\linewidth]{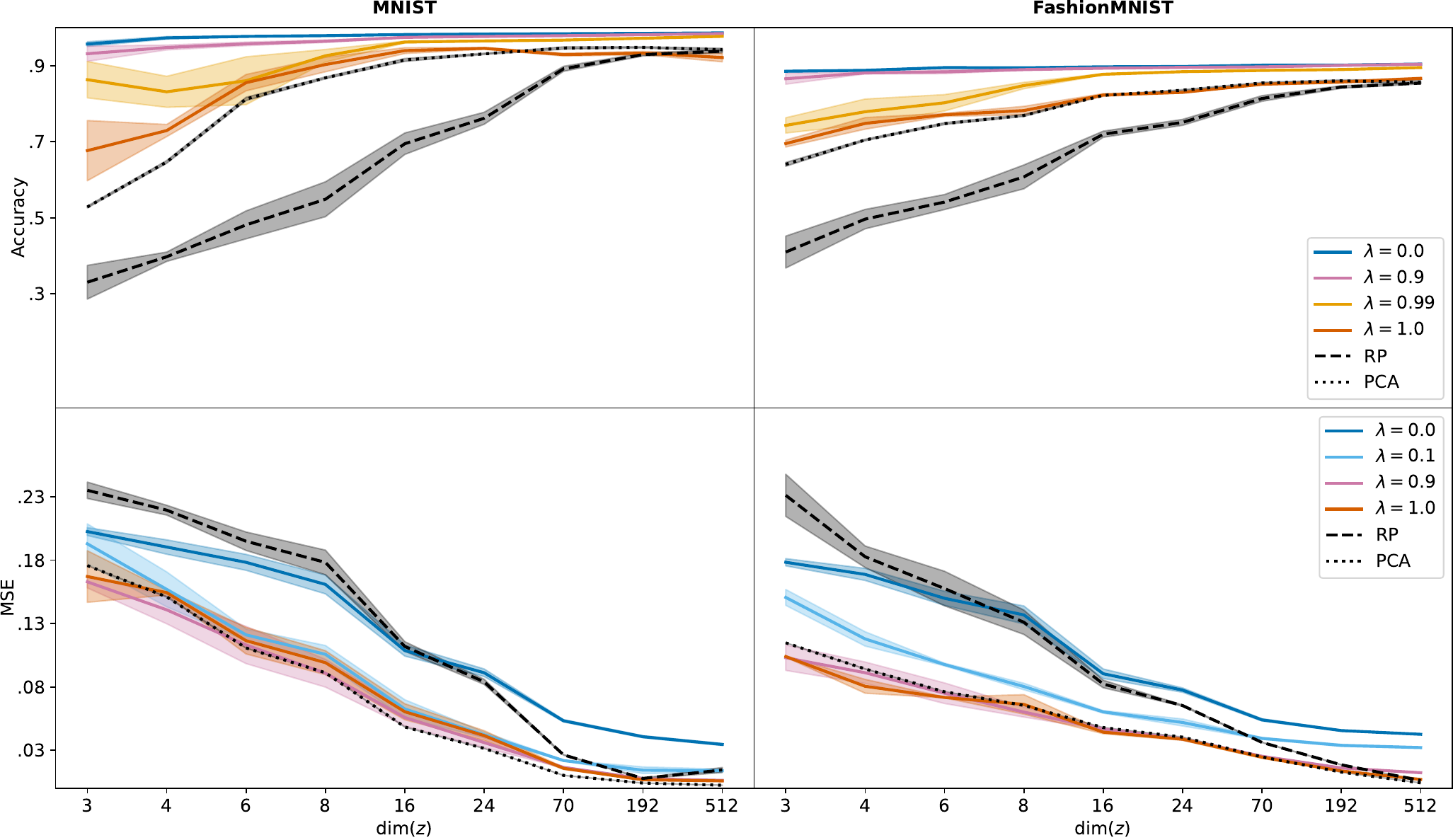}
    \caption{Trade-Off effect by dimensionality of the latent space: Plots show the average classification (top row) and reconstruction (bottom row) performance of FC-based CREs on MNIST (left column) and FashionMNIST(right column) for varying $\lambda$-values, within a 95\% confidence interval. The performances on random projections RP-based and PCA-based representations are also included. }
    \label{fig:dims_fc}
\end{figure}

\Cref{fig:dims_fc} outlines how the trade-off effect changes with increasing latent dimension for FC-based CREs on MNIST and FashionMNIST. We display varying $\lambda$-values and benchmark these against PCA- and RP-based representations. Note that augmenting the latent dimensions leads to higher model complexity. This complexity arises from the dependency of the parameter count in layers, which are directly connected to the latent layer, on the size of the latent layer itself. Therefore, while the plots provide valuable indications of the influence of varying dimensions, not all observed effects can be solely attributed to changes in the number of latent dimensions.

In the plots, at each specific dimension, the trade-off effect is reflected in the distance between one particular $\lambda$-value to its top-performing counterpart, i.e., for classification, the distance to $\lambda = 0$ and for reconstruction, the distance to $\lambda=1$. A pronounced disparity between two lines signifies a significant trade-off effect. Conversely, a narrower gap indicates a more modest trade-off impact.

In the context of classification, our observations for both MNIST and FashionMNIST datasets reveal a notable trend. Consistent with our prior results, the performance rank ordering of $\lambda$-values remains stable across dimensions. Notably, as $\text{dim}(\bb{z})$ increases, there is a general improvement in accuracy. However, for $\lambda=0$, maximum or near maximum performance is already attained at a low number of dimensions, e.g., $\text{dim}(\bb{z}) = 4$, and does not significantly enhance with additional dimensions. Thus, the trade-off reflected in the gap to $\lambda=0$ diminishes with a larger latent space. Interestingly, while all $\lambda$-values smaller than one close or almost close the gap to $\lambda=0$ after some dimension threshold, this pattern is significantly less pronounced for $\lambda = 1$, PCA-, and RP-based representations.

Regarding reconstruction performance, the trends we observe show notable differences from those in classification. Unlike classification, where maximum performance is reached for some variant of representations at a low number of dimensions, reconstruction efficiency consistently improves as dimensions increase, regardless of the representation type. Furthermore, the performance ranking among different representation types is less stable than in classification. For instance, the best reconstructions in low dimensions are typically achieved with $\lambda = 1$ or $\lambda = 0.99$. However, at higher dimensions, PCA-based representations often perform as well as or better than these $\lambda$-values. Additionally, while RP-based representations consistently rank lowest in classification, their performance in reconstruction is markedly different. They tend to underperform at lower dimensions, but as dimensions increase, they surpass primarily classification-driven representations. 

Our evaluation of the impact of architecture complexity and latent space size supports our hypothesis regarding the mitigation of the trade-off effect: Increasing either model complexity or the size of the latent space indeed diminishes the trade-off effect to a negligible level. This effect is especially apparent in representations that are primarily driven by reconstruction while incorporating a bit of classification-driven information, i.e., $\lambda \sim 0.9$. Beyond a certain complexity or latent space threshold, these models achieve performance that is comparable to, or nearly as good as, models trained with extreme $\lambda$-values. \Cref{fig:dims_fc} implies an explanation for this effect. A sufficient amount of classification-driven information requires less space for encoding than reconstruction-driven information. While the classification still slightly increases with more latent dimensions, the benefit is significantly smaller than for the reconstruction process. Therefore, when these two processes compete for resources, a balance favoring reconstruction while still accounting for classification to some extent, e.g., $\lambda=0.9$, leads to satisfactory performance.

\subsection{Transferring Representations}
We have observed a particularly evident decline in classification performance when transferring representations, i.e., when $\lambda=1$. We now want to evaluate the masking techniques v1 (reconstructing masked patches only) and v2 (reconstructing full image) on the ViT-based CRE as they are considered to boost the transferability of reconstruction-driven representations~\cite{he_masked_2022}.

\begin{figure}[ht]
    \centering
    \includegraphics[width=1.0\linewidth]{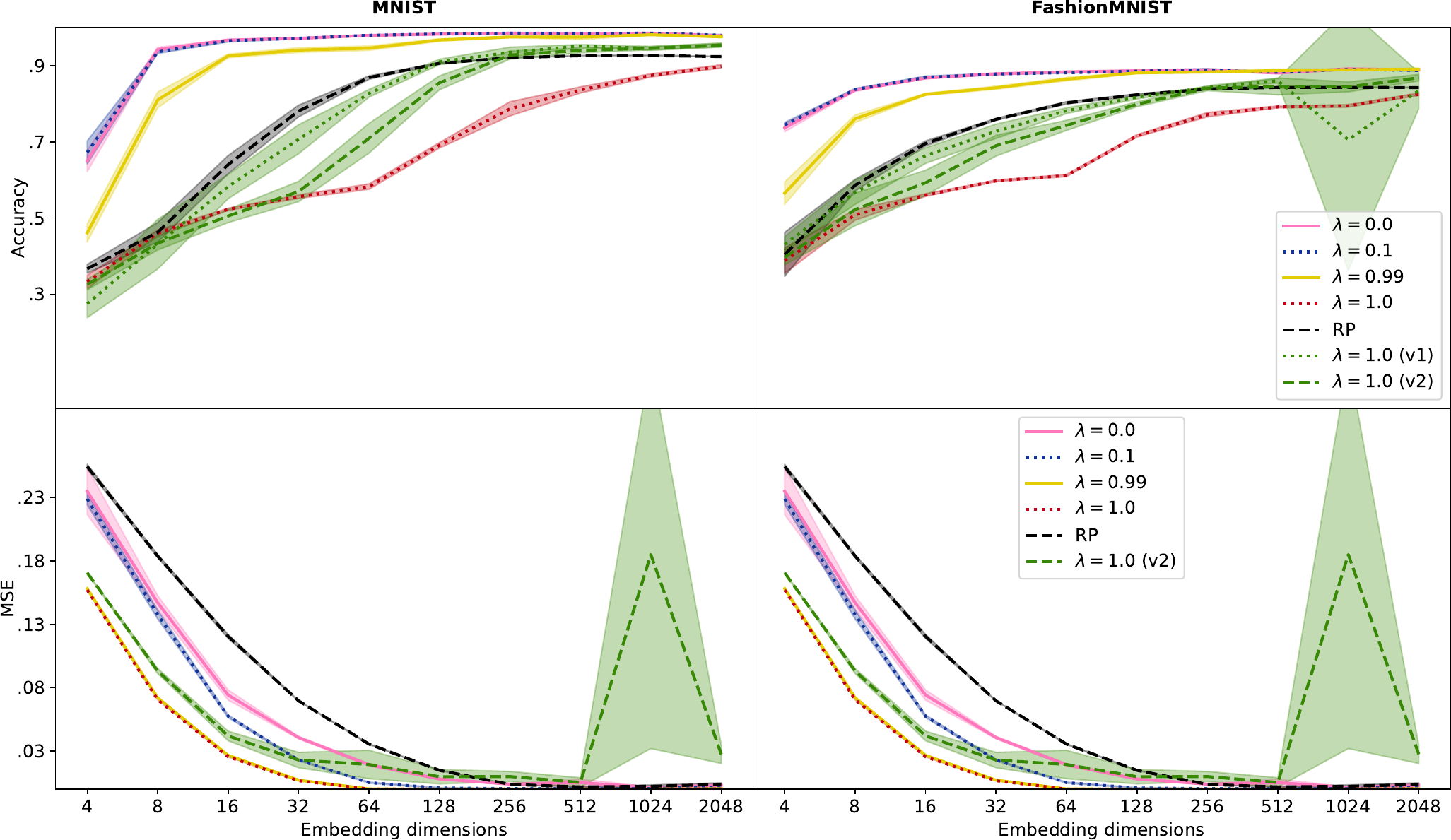}
    \caption{Trade-off effect by embedding dimensions: Plots show the average classification (top row) and reconstruction (bottom row) performance of ViT-based CREs on MNIST (left column) and FashionMNIST(right column) for v1 and v2. As reference, the performances of PCA-based representation and varying $\lambda$-values-based representations are shown. All performance are plotted within a 95\% confidence interval.}
    \label{fig:dims_vit}
\end{figure}

\Cref{fig:dims_vit} illustrates how v1- and v2-based representation with $\lambda  = 1$ behave in the context of our analysis of the trade-off effect when applied to the MNIST and FashionMNIST datasets. This figure compares these representations' classification and reconstruction performances across various embedding dimensions. As references, we include the performances of RP-based representations and those emerging from our standard training at different $\lambda$-values. Note that in contrast to \Cref{fig:dims_fc}, enhancing the embedding dimensions expands not only the number of latent dimensions but also the size of each layer within the encoder and decoder. Consequently, the figure simultaneously demonstrates the impact of augmenting the latent space and the architecture's complexity.

In general, we obtain similar results as for the FC-based CRE in \Cref{fig:dims_fc} for all representations. Both classification and reconstruction performance scale with the embedding dimensions, while mostly maintaining a rank ordering in terms of performance, though training becomes unstable for some masking-based representations in high embedding dimensions.

Focusing on v1 and v2 in classification performance, we successfully reproduce the results of He et al.~\cite{he_masked_2022}, i.e., the masking process significantly increases the transferability of reconstruction-driven representations compared to scenarios without masking. Furthermore, our study verifies that reconstructing only the masked patches of an image, as done in v1, is more beneficial for accuracy than reconstructing the entire image, which is the approach of v2. Interestingly, masking outperforms RP only after surpassing a certain dimensional threshold. In general, however, both masking techniques underperform compared to training the architecture with a small amount of classification-driven data, such as in cases where $\lambda = 0.99$.

Assessing the reconstruction performance of masked-based representations, we observe that v2's performance is inferior to that of its unmasked counterpart. Notably, while v2 does outperform classification-based representations at lower embedding dimensions, it is surpassed by purely classification-based representations after crossing a certain dimensional threshold. This shift in performance dynamics is especially marked when incorporating small amounts of reconstruction-driven information, e.g., $\lambda = 0.1$. Note that we excluded technique v1 from the bottom plots as this approach focuses on reconstructing only masked portions of images and thus fails to reconstruct whole images. 

In contrast to the FC-based decoder in \Cref{fig:dims_fc}, the ViT-based decoder achieves perfect reconstructions (MSE equals 0) at specific embedding dimensions, particularly when $\lambda = 1$. We primarily attribute this difference to the architectural features of these systems. In particular, ViT-based CREs are no longer constrained by the need to compress data once embedding dimensions reach or exceed 64 since the total number of latent dimensions is equal to the number of embedding dimensions times the number of image patches. 

Our evaluation of masking-based representations indicates problems transferring reconstruction-based representations to classification in the context of shared representations. While the masking techniques increase accuracy, they do so at the price of reconstruction performance. In addition, incorporating classification-driven information into primarily reconstruction-driven objectives, e.g., $\lambda=0.99$, outperforms the masking techniques for both classification and reconstruction.

\section{Discussion and Conclusion}
\label{sec:discussion_and_conclusion}
In this study, we have explored the interplay between classification- and reconstruction-driven information in shared representations. Our experiments across various architectures unveil a consistent pattern: integrating these two types of information necessitates a trade-off. Neither did the reconstruction process benefit from classification-driven information nor did reconstruction-driven information benefit the classification process. Prioritizing one kind of information diminishes the efficacy of the other. 

We have found that this trade-off effect can be alleviated by increasing the network's complexity or expanding the dimensions of the shared representations, suggesting that classification- and reconstruction-driven information compete for available resources. Consequently, while the two kinds of information can coexist in a shared representation once the resources are large enough, they do not integrate synergistically.

Our findings have several implications for the development and understanding of DPCNs. The trade-off effect suggests reassessing prevailing DPCN models, specifically their mechanisms for handling this phenomenon. Such reevaluation resonates with the insights of Rane et al.~\cite{rane_prednet_2020}, who observed antagonistic effects in the DPCN variant PredNet~\cite{lotter_deep_2017}. Moreover, the identified trade-off effect provides insights for future DPCN designs, suggesting the need for strategic considerations to balance and integrate classification and reconstruction-driven information. 

The interference of reconstruction-driven information with classification-driven information raises questions about the viability of using the MSE for low-level representation predictions in the predictive coding framework. Future research should thus focus on elucidating what should be predicted and how. Inspiration for answers could be drawn from the human visual system. For instance, as suggested by Breedlove et al.~\cite{breedlove_generative_2020}, fewer neurons are active during image synthesis in the visual cortex than in forward processing. This implies that the predictive process in the brain generates less detailed images, a concept that could inform adaptations in DPCN methodologies.

Regarding PCNs, our results may not apply directly to these architectures since we used tools from deep learning instead of the predictive coding framework. However, given that PCNs either excel at reconstruction or classification but not both~\cite{millidge_2022_predictive_survey, sun_predictive-coding_2020}, similar effects may likely affect these architectures, and reconstruction and classification-driven information do not synergize in these architectures as much as presumed. Therefore, future work could analyze if the trade-off or similar effects also occur in PCNs.

A critical insight from our study is the challenge inherent in the predictive coding framework's reliance on unsupervised signals to learn rich representations~\cite{millidge_2022_predictive_survey}. In our experiments, this challenge was exemplified by the scenario where the classifier was trained exclusively with reconstruction-driven information, resulting in notably poor classification performance. Moreover, when we implemented techniques to enhance the transfer of reconstruction-driven representations to the classification task, we observed an improvement in accuracy but at the cost of diminished reconstruction quality. This outcome underscores the need for future research to explore strategies for effectively extracting classification-driven information in an unsupervised setting while preserving the integrity of reconstruction-driven information.

Our work presents certain limitations that open avenues for future research. Firstly, while we followed the methodology from He et al.~\cite{he_masked_2022} for our ViT-based CRE, we believe that the classification performance can be greatly improved. For instance, not passing an average of the encoder's output tokens to the classifier but instead connecting the classifier to each of the encoder's output tokens should benefit accuracy. In this regard, classification and reconstruction processes could be integrated more effectively into the architecture.

Secondly, despite our efforts to utilize a range of deep learning tools resulting in various CRE variants, the possibility of tools capable of achieving a synergistic effect in representing the two kinds of information remains. Future work may be targeted at finding alternatives to our utilized encoders, potentially allowing a successful integration of the two types of information. 

Finally, we did not address two critical properties of DPCNs, which may affect the synergy between reconstruction and classification processes. To begin with, we did not explicitly consider the distribution of classification- and reconstruction-driven information across all layers of a hierarchical network. Instead, we focused on integrating the two kinds of information into a singular layer. Furthermore, we focused on static representations, overlooking the dynamic nature of the information flow in DPCNs characterized by alternating bottom-up and top-down sweeps. Investigating how the two properties influence the interaction of classification and reconstruction processes could yield insightful results. 
  
In conclusion, our research illuminates the intricate challenges of combining classification and reconstruction processes within a deep learning context. We have identified a notable trade-off effect between classification- and reconstruction-driven information in the shared representations. This effect can be alleviated to a certain extent by enhancing the network's complexity or increasing the size of the representations' dimensions, although not synergistically. While our results are relevant to the broader context of deep learning, they are particularly pertinent to DPCNs and, potentially, PCNs. Our findings suggest reconsideration and possible adaptations to DPCN methodologies, paving the way for future explorations. However, a thorough understanding of the interactions of combining classification and reconstruction processes in dynamic processing remains an open field for further research.

\section*{Acknowledgements}
We thank Bruno Olshausen for his valuable and insightful comments on this project.

\medskip
\small{
\bibliography{references}
}
\clearpage
\appendix
\label{supplement}
\section{Implementation Details}
We implemented and trained all models using Python (v. 3.9.6) and PyTorch (v. 2.0.1). Unless otherwise mentioned, the default PyTorch parameters were employed. Detailed implementations, trained models, and methods for reproducing our results can be found in our code repository.\footnote{\repo}

\subsection{Architectural Details}
Architectural details of all used CRE-variants are summarized in Tables \ref{tab:model_summary_fc}, \ref{tab:model_summary_cnn}, and \ref{tab:model_summary_vit}. The variable $n$ denotes the dimension of the latent representations in FC-based CREs and the embedding dimensions in ViT-based CREs. We built the FC- and CNN-based CREs following deep learning conventions for image processing. We adapted ViT-based CRE from the authors of the ViT-based masked autoencoder \cite{he_masked_2022}. \footnote{\url{https://github.com/facebookresearch/mae}}.

\begin{table}[H]
\centering
\tiny
\begin{tabular}[t]{llr}
\toprule
\textbf{Module} & \textbf{Layer (type)} & \textbf{Output Shape} \\
\midrule
Encoder &  Flatten & 784  \\
& Flatten            & 784             \\
& Linear             & $n$              \\
& ReLU               & $n$             \\
\hline
Decoder & Linear   & 784     \\
& Tanh              & 784       \\
& Unflatten          & 1x28x28  \\
\hline
Classifier & Linear  & 10              \\
\toprule
\end{tabular}
\quad 
\begin{tabular}[t]{lll}
\toprule
\textbf{Module} & \textbf{Layer (type)} & \textbf{Output Shape} \\
\midrule
Encoder & Flatten            & 784         \\
&Linear            & 616   \\
&ReLU               & 616        \\
&Linear             & $n$        \\
&ReLU               & $n$      \\
\hline
Decoder &Linear             &  616   \\
&ReLU        & 616     \\
&Linear              & 784  \\
&Tanh               & 784     \\
&Unflatten         &  1x28x28     \\
\hline
Classifier &Linear            & 12  \\
&ReLU              & 12  \\
& Linear            & 10  \\
\toprule
\end{tabular}
\quad
\begin{tabular}[t]{llr}
\toprule
\textbf{Module} & \textbf{Layer (type)} & \textbf{Output Shape}  \\
\midrule
Encoder &Flatten             & 784         \\
&Linear              & 716            \\
&ReLU                & 716             \\
&Linear              &  684            \\
&ReLU                & 684          \\
&Linear              & 616            \\
&ReLU                & 616          \\
&Linear              & $n$        \\
&ReLU                & $n$        \\
\hline
Decoder &Linear     &  616           \\
&ReLU               &  616       \\
&Linear             & 684         \\
&ReLU               & 684           \\
&Linear             & 716            \\
&ReLU               & 716            \\
&Linear             & 784     \\
&Tanh               & 784       \\
&Unflatten          & 1x28x28        \\
\hline
Classifier & Linear  & 64      \\
&ReLU              &  64             \\
&Linear             & 10                \\
\toprule
\end{tabular}
\caption{FC-based CREs' layers. From left to right: small, medium, large FC-based CRE. $n$ represents the number of latent dimensions}
\label{tab:model_summary_fc}
\end{table}

\begin{table}[H]
\centering
\tiny
\begin{tabular}[t]{llr}
\toprule
\textbf{Module} &\textbf{Layer (type)} & \textbf{Output Shape}\\
\midrule
Encoder &Conv2d              & 16x28x28        \\
&BatchNorm2d         & 16x28x28     \\
&MaxPool2d            & 16x14x14      \\
&ReLU              & 16x14x14          \\
&Conv2d               & 16x14x14       \\
&BatchNorm2d          & 16x14x14      \\
&MaxPool2d            & 16x8x8              \\
&ReLU              & 16x8x8         \\
&Conv2d              & 16x8x8         \\
&BatchNorm2d         & 16x8x8         \\
&MaxPool2d          & 16x4x4            \\
&ReLU                & 16x4x4          \\
\hline
Decoder & ConvTranspose2d     & 16x8x8     \\
&ReLU               & 16x8x8             \\
&ConvTranspose2d     & 16x14x14     \\
&ReLU               & 16x14x14          \\
&ConvTranspose2d     & 1x28x28          \\
&Tanh                & 1x28x28         \\
\hline
Classifier  & Flatten             & 256 \\
&Linear              & 10  \\
\toprule
\end{tabular}
\quad
\begin{tabular}[t]{llr}
\toprule
\textbf{Module} & \textbf{Layer (type)} & \textbf{Output Shape}\\
\midrule
Encoder &Conv2d              &  64x32x32       \\
&Dropout2d           & 64x32x32           \\
&BatchNorm2d        & 64x32x32          \\
&ReLU           & 64x32x32           \\
&Conv2d              & 64x32x32    \\
&Dropout2d           & 64x32x32     \\
&BatchNorm2d         & 64x32x32      \\
&ReLU                & 64x32x32         \\
&Conv2d              & 128x32x32   \\
&Dropout2d          & 128x32x32         \\
&BatchNorm2d        & 128x32x32    \\
&MaxPool2d          & 128x16x16\\
&ReLU             & 128x16x16       \\
&Conv2d             & 128x16x16  \\
&Dropout2d          & 128x16x16     \\
&BatchNorm2d        & 128x16x16    \\
&MaxPool2d          & 128x8x8         \\
&ReLU               & 128x8x8     \\
&Conv2d             & 128x8x8    \\
&Dropout2d          & 128x8x8       \\
&BatchNorm2d        & 128x8x8       \\
&MaxPool2d          & 128x4x4        \\
&ReLU               & 128x4x4    \\
\hline
Decoder&ConvTranspose2d    & 128x8x8        \\
&ReLU               & 128x8x8            \\
&ConvTranspose2d    & 128x16x16     \\
&ReLU               & 128x16x16         \\
&ConvTranspose2d    & 64x32x32       \\
&ReLU               & 64x32x32         \\
&ConvTranspose2d    & 64x32x32     \\
&ReLU               & 64x32x32           \\
&ConvTranspose2d    & 3x32x32       \\
&Tanh               & 3x32x32]         \\
\hline
Classifier &Conv2d             & 256x4x4       \\
&Dropout2d          & 256x4x4              \\
&BatchNorm2d        & 256x4x4       \\
&ReLU               & 256x4x4            \\
&Conv2d             & 256x4x4       \\
&Dropout2d         & 256x4x4             \\
&MaxPool2d          & 256x2x2          \\
&BatchNorm2d        & 256x2x2          \\
&ReLU               & 256x2x2            \\
&Flatten            & 1024                \\
&Linear             &  10              \\
\toprule
\end{tabular}
\caption{CNN-based CREs' layers. From left to right: small, large CNN-based CRE}
\label{tab:model_summary_cnn}
\end{table}

\begin{table}[H]
\centering
\tiny
\begin{tabular}[t]{llr}
\toprule
\textbf{Module} & \textbf{Layer (type)} & \textbf{Output Shape}  \\
\midrule
       Encoder  & Conv2d           & 4x4x4    \\       
    &Identity         & 16x4       \\      
    &PatchEmbed       & 16x4       \\
    \hdashline
    Block x 1& LayerNorm        & 5x4   \\          
    &Linear          & 5x12           \\ 
    &Identity         & 4x5x1          \\
    &Identity         & 4x5x1          \\
    &Linear           & 5x4             \\
    &Dropout          & 5x4               \\
    &Attention       & 5x4               \\
    &Identity        & 5x4               \\
    &Identity        & 5x4          \\
    &LayerNorm       & 5x4           \\  
    &Linear          & 5x16             \\ 
    &GELU            & 5x16             \\
    &Dropout         & 5x16               \\
    &Identity        & 5x16             \\
    &Linear          & 5x4            \\
    &Dropout         & 5x4            \\
    &Mlp            & 5x4              \\ 
    &Identity        & 5x4              \\
    &Identity        & 5x4                \\
    \hdashline    
    &LayerNorm       & 5x4              \\
    \hline
    Decoder &Linear          & 5x4   \\
    \hdashline
    Block x 1 &LayerNorm       & 17x4      \\         
    &Linear    & 17x12             \\
    &Identity        & 4x17x1           \\
    &Identity        & 4x17x1         \\
    &Linear          & 17x4             \\
    &Dropout         & 17x4            \\
    &Attention       & 17x4             \\
    &Identity        & 17x4              \\
    &Identity        & 17x4           \\
    &LayerNorm       & 17x4            \\
    &Linear          & 17x16             \\
    &GELU            & 17x16        \\
    &Dropout         & 17x16          \\  
   & Identity        & 17x16          \\
   & Linear          & 17x4            \\
   & Dropout         & 17x4             \\
   & Mlp            & 17x4             \\
   & Identity        & 17x4             \\
   & Identity        & 17x4  \\
   \hdashline
   & LayerNorm       & 17x4      \\       
   & Linear          & 17x49       \\     
    \hline
   Classifier &  BatchNorm1d     & 4               \\   
    & Linear          & 10               \\
    \bottomrule
\end{tabular}
\begin{tabular}[t]{llr}
        \toprule
        \textbf{Module} & \textbf{Layer (type)} & \textbf{Output Shape} \\
        \midrule
        Encoder &Conv2d & 32x16x16 \\
        &Identity & 256x32 \\
        &PatchEmbed & 256x32 \\
        \hdashline
        Block x 7 &LayerNorm & 65x32  \\
        &Linear & 65x96  \\
        &Identity & 8x65x4  \\
        &Identity & 8x65x4 \\
        &Linear & 65x32  \\
        &Dropout & 65x32  \\
        &Attention & 65x32 \\
        &Identity & 65x32 \\
        &DropPath & 65x32 \\
        &LayerNorm & 65x32\\
        &Linear & 65x128  \\
        &GELU & 65x128 \\
        &Dropout & 65x128 \\
        &Identity & 65x128 \\
        &Linear & 65x32  \\
        &Dropout & 65x32\\
        &Mlp & 65x32 \\
        &Identity & 65x32 \\
        &DropPath & 65x32\\
        \hdashline
        &LayerNorm & 65x32\\
        \hline

        Decoder &Linear & 65x32\\    
        \hdashline
        Block x 7 &LayerNorm & 257x32 \\
        &Linear & 257x96  \\
        &Identity & 8x257x4  \\
        &Identity & 8x257x4  \\
        &Linear & 257x32 \\
        &Dropout & 257x32  \\
        &Attention & 257x32  \\
        &Identity & 257x32  \\
        &Identity & 257x32 \\
        &LayerNorm & 257x32 \\
        &Linear & 257x128  \\
        &GELU & 257x128 \\
        &Dropout & 257x128  \\
        &Identity & 257x128 \\
        &Linear & 257x32  \\
        &Dropout & 257x32 \\
        &Mlp & 257x32  \\
        &Identity & 257x32  \\
        &Identity & 257x32  \\
        \hdashline
        &LayerNorm & 257x32  \\
        &Linear & 257x12 \\
        \hline
        Classifier & BatchNorm1d & 32  \\
        &Linear & 10  \\
        \bottomrule
    \end{tabular}
\caption{ViT-based CREs’ layers. From left to right: small, large CNN-based ViT. The small version is exemplified with four embedding dimensions}
\label{tab:model_summary_vit}
\end{table}

\subsection{Training}
\paragraph{Datasets and Model Variants:} 
We trained all FC-based variants (small, medium, large) on MNIST and FashionMNIST. For the size of the latent representation we used $n \in \{3, 4, 6, 8, 16, 24, 70, 192, 512\}$. Additionally, we trained small CNN-based and small ViT-based on MNIST and FashionMNIST. For the embedding dimension of the small ViT-based CREs, we used $n \in \{4, 8, 16, 32, 64, 128, 256, 512, 1024, 2048\}$. Furthermore, we trained large CNN-based and large ViT-based CREs on CIFAR-10.

\paragraph{Parameter Configuration and Model Count:} 
For each model configuration, we trained variants with the following values for the $\lambda$-parameter: $\lambda \in \{0, 0.01, 0.05, 0.1, 0.2, 0.4, 0.6, 0.8, 0.9, 0.95, 0.99, 1.0\}$. We ensured that at least five models were trained for each configuration, culminating in approximately 10000 models in total.

\paragraph{Training Duration and Optimization:}
The number of training epochs varied between 100 and 700, depending on the model configuration and the dataset used. We assigned individual Adam optimizers from PyTorch, with default settings, to each module of the CRE (encoder, decoder, classifier). The step sizes for these optimizers ranged from 0.0001 to 0.001. Our optimization strategy involved parallel training of modules, as it demonstrated superior performance in terms of both classification accuracy and reconstruction quality. 

\paragraph{Data Augmentation and Regularization Techniques:} 
For the CIFAR-10 dataset, we implemented random rotation as a data augmentation strategy. Additionally, dropout regularization was set to 0.3 for CNN-based CREs and droppath regularization to 0.2 for ViT-based CREs. We also stored the all metrics after each epoch and only used the best performing versions for our evaluations.

\section{Additional Results}
\subsection{Additional Performance Results}
\begin{figure}[ht]
    \begin{subfigure}[t]{0.246\linewidth}
        \centering
        \includegraphics[width=1.0\linewidth]{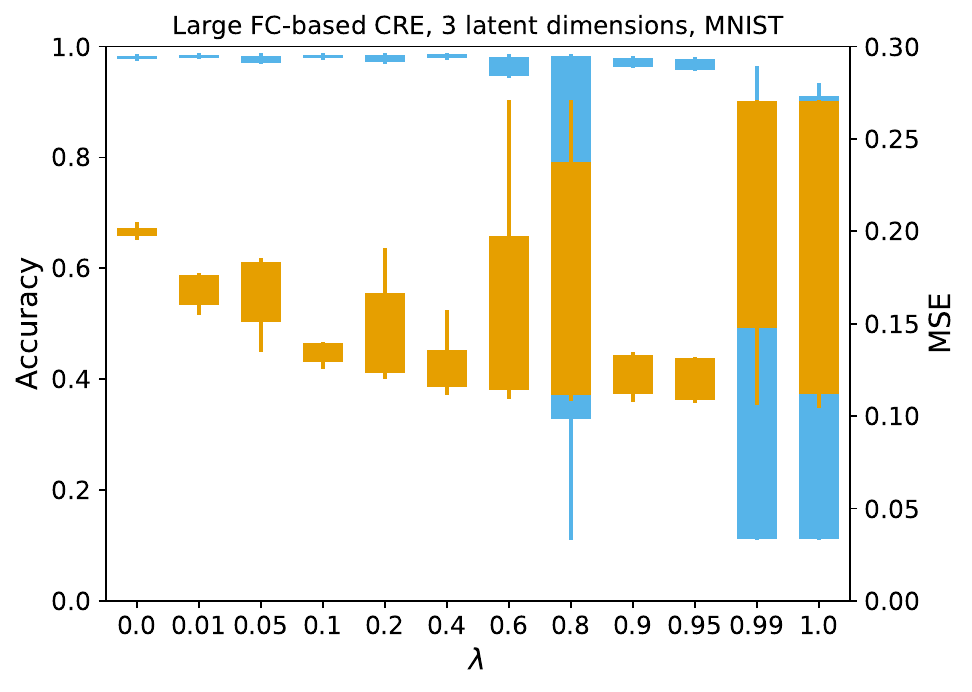}
        \includegraphics[width=1.0\linewidth]{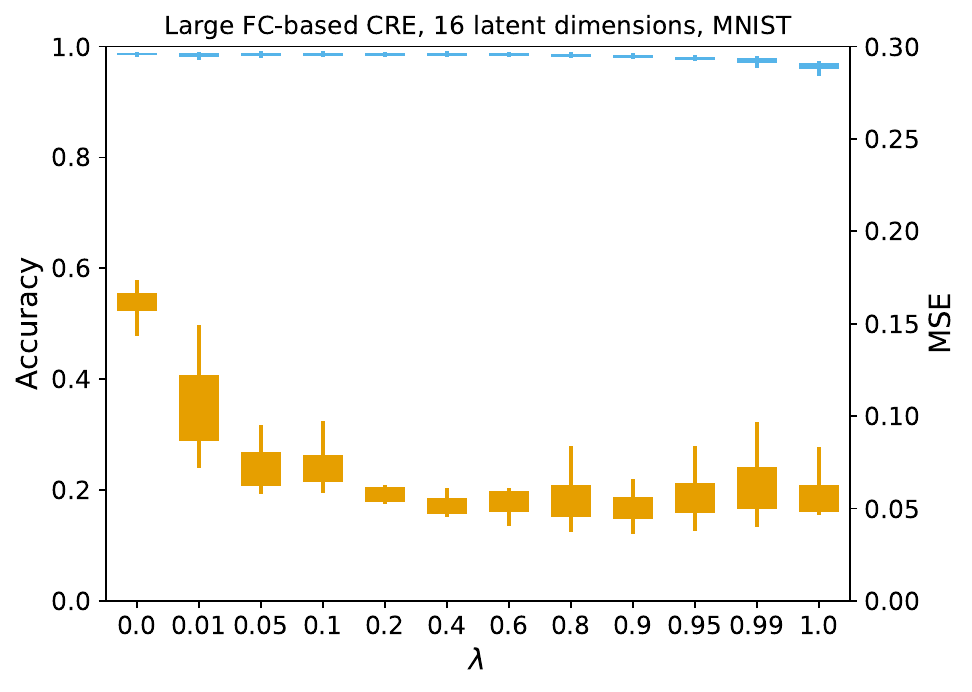}
        \includegraphics[width=1.0\linewidth]{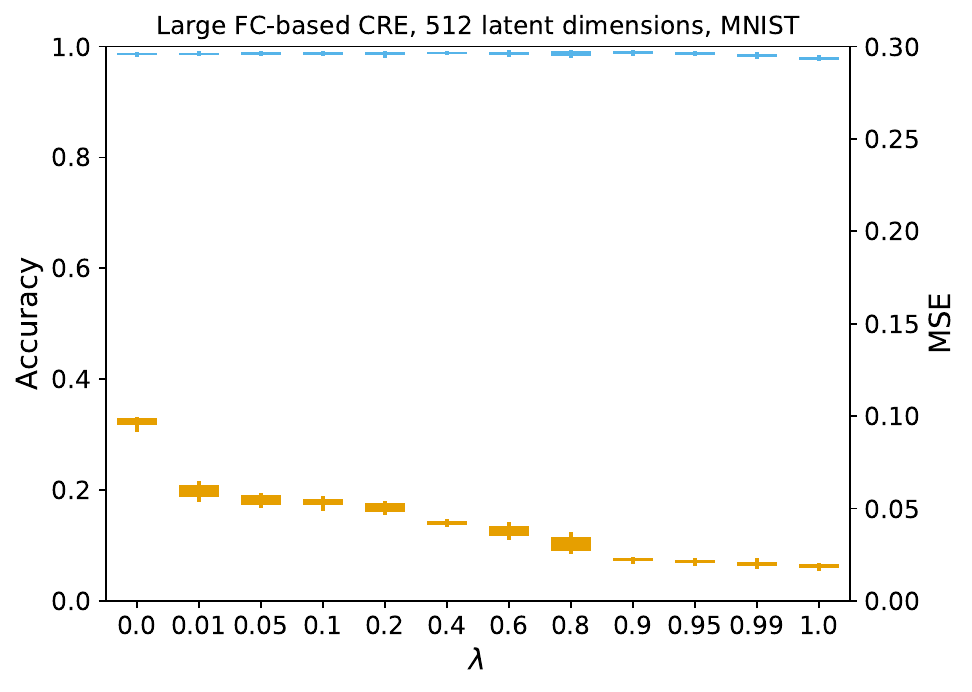}
        \includegraphics[width=1.0\linewidth]{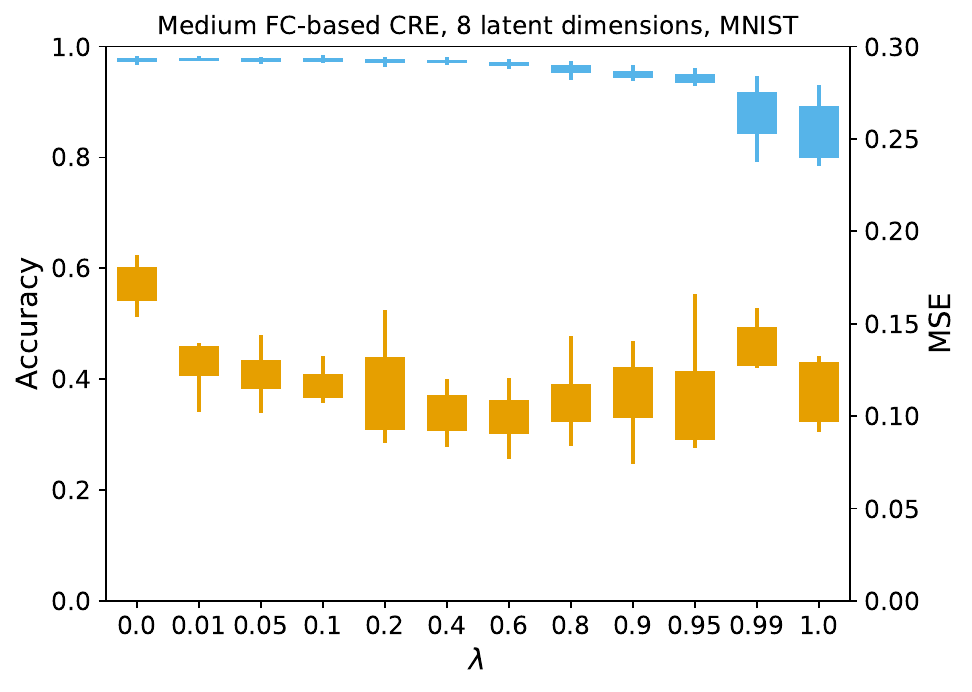}
        \includegraphics[width=1.0\linewidth]{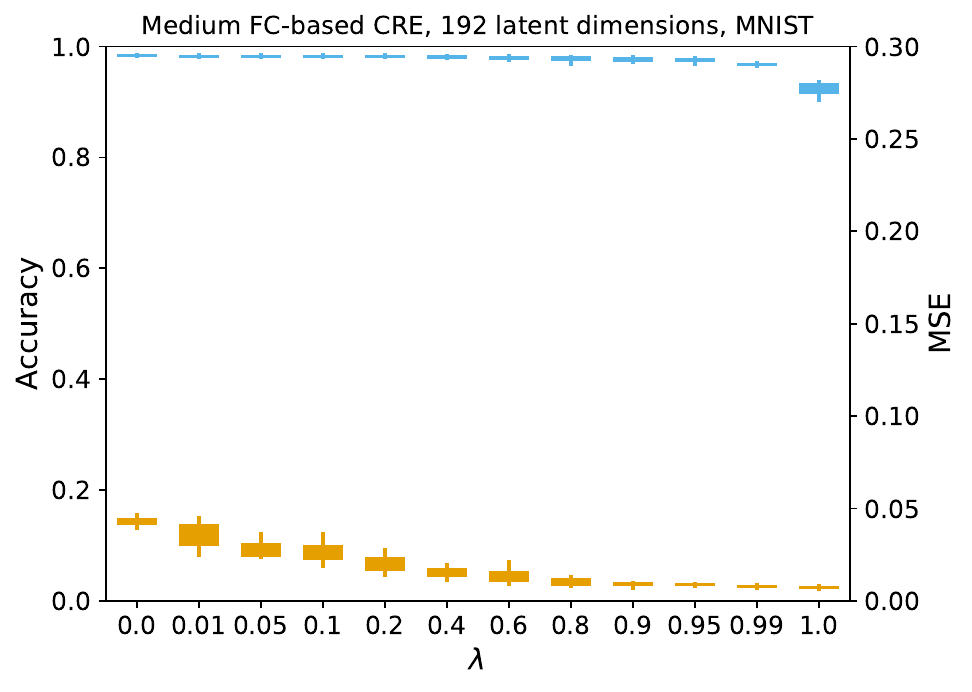}
         \includegraphics[width=1.0\linewidth]{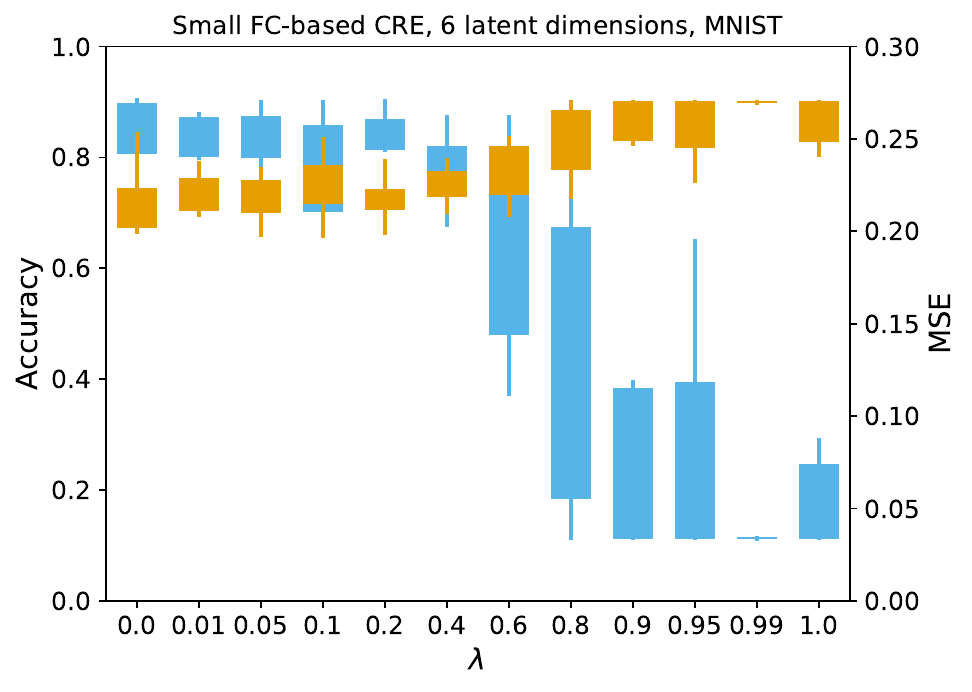}
        \includegraphics[width=1.0\linewidth]{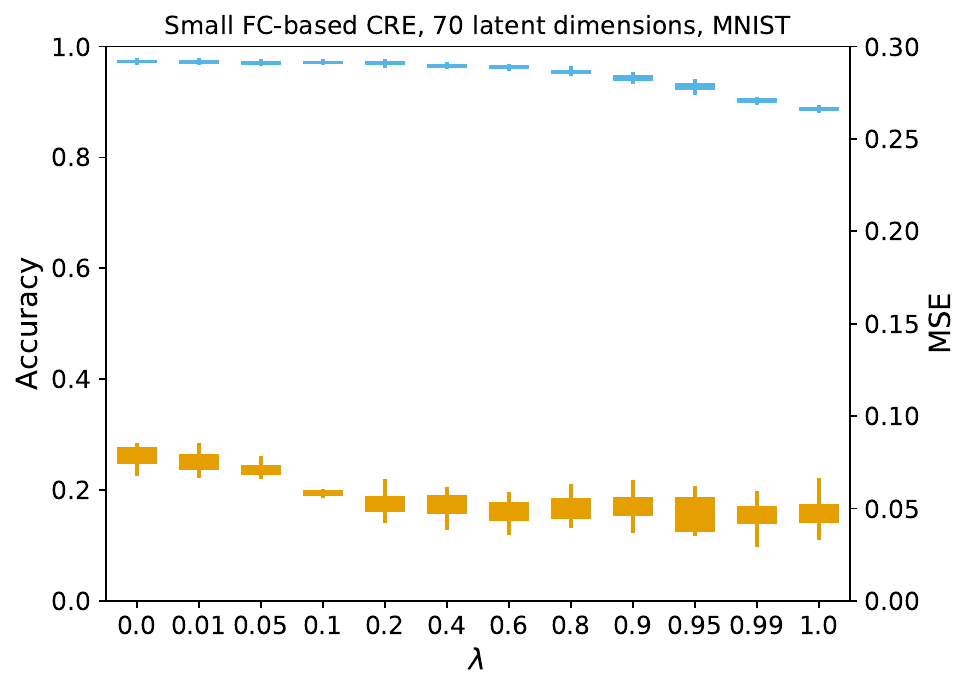}
    \end{subfigure}
    \begin{subfigure}[t]{0.246\linewidth}
        \includegraphics[width=1.0\linewidth]{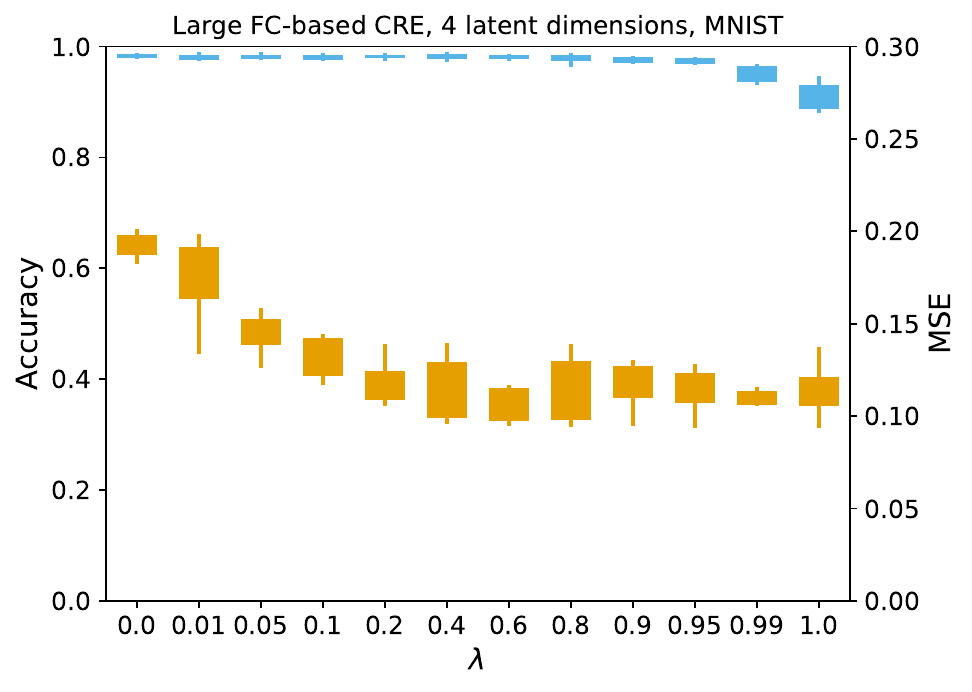}
        \includegraphics[width=1.0\linewidth]{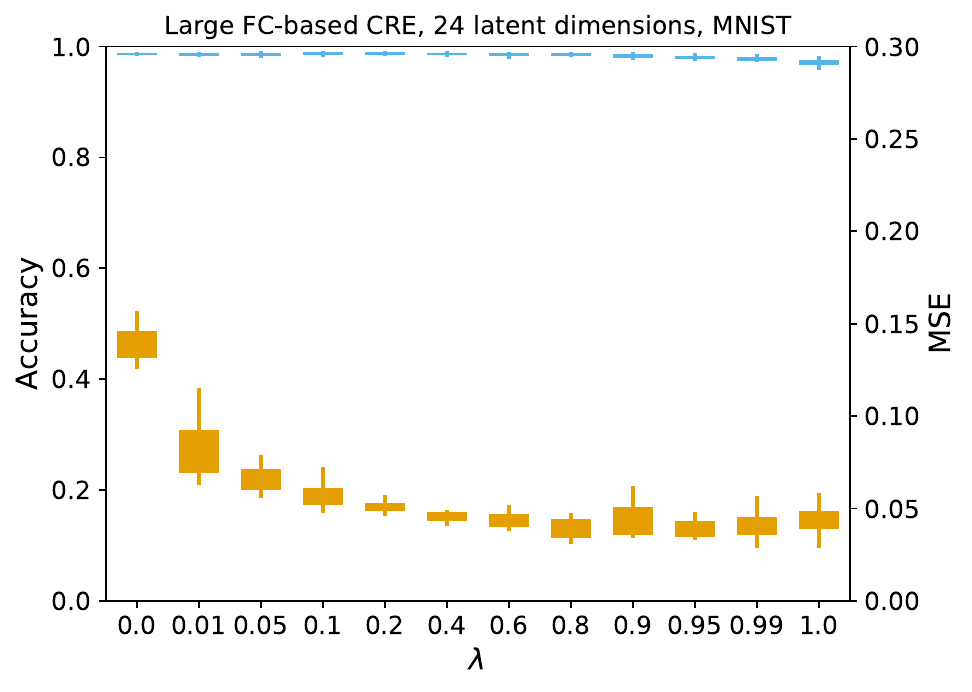}
        \includegraphics[width=1.0\linewidth]{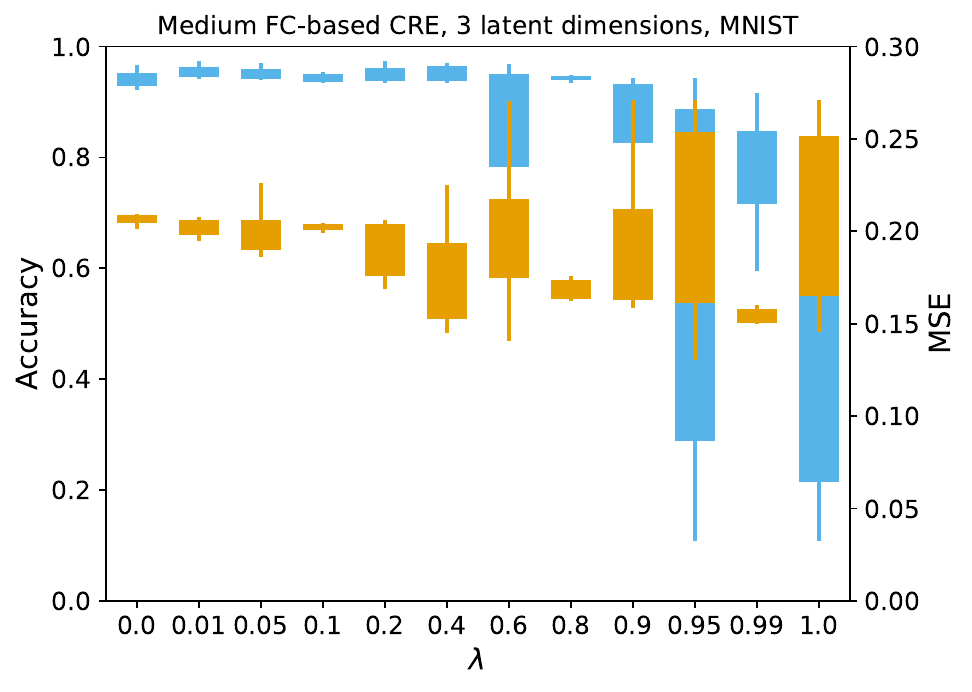}
        \includegraphics[width=1.0\linewidth]{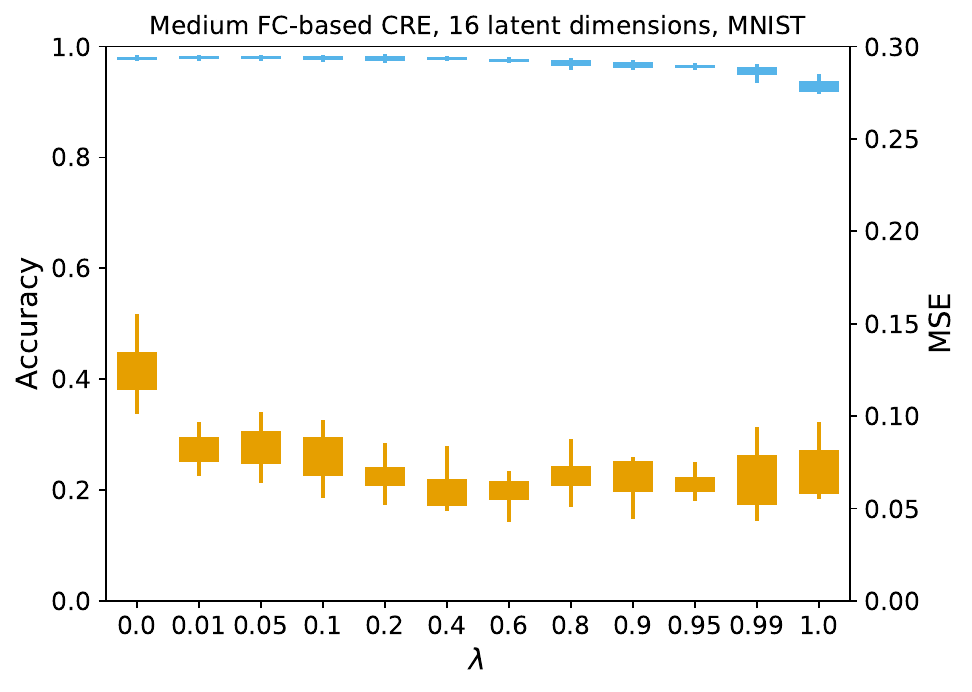}
        \includegraphics[width=1.0\linewidth]{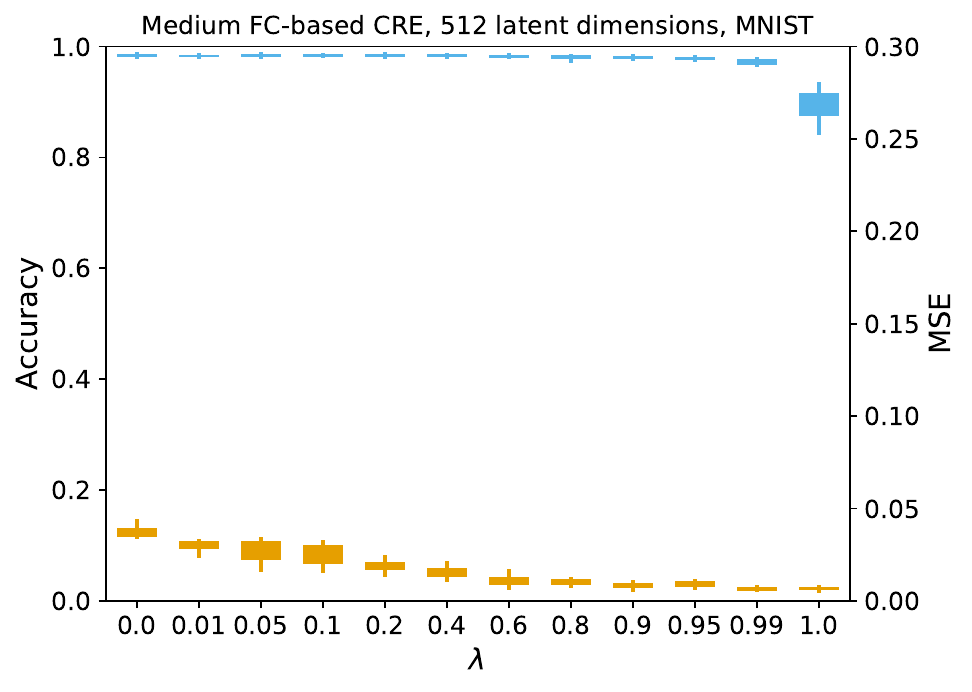}
        \includegraphics[width=1.0\linewidth]{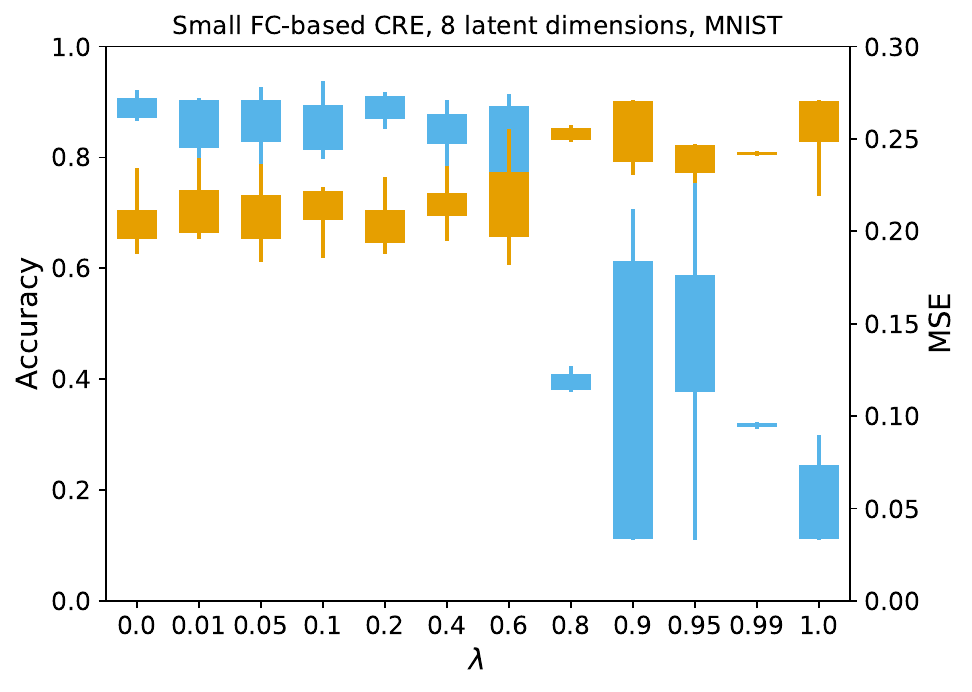}
        \includegraphics[width=1.0\linewidth]{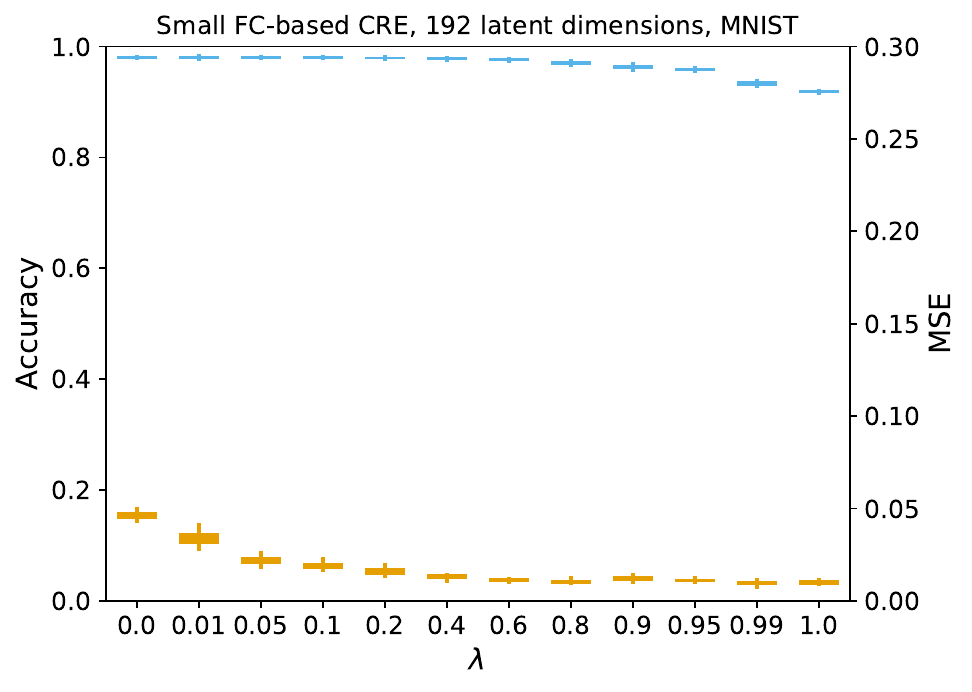}

    \end{subfigure}
    \begin{subfigure}[t]{0.246\linewidth}
        \centering
        \includegraphics[width=1.0\linewidth]{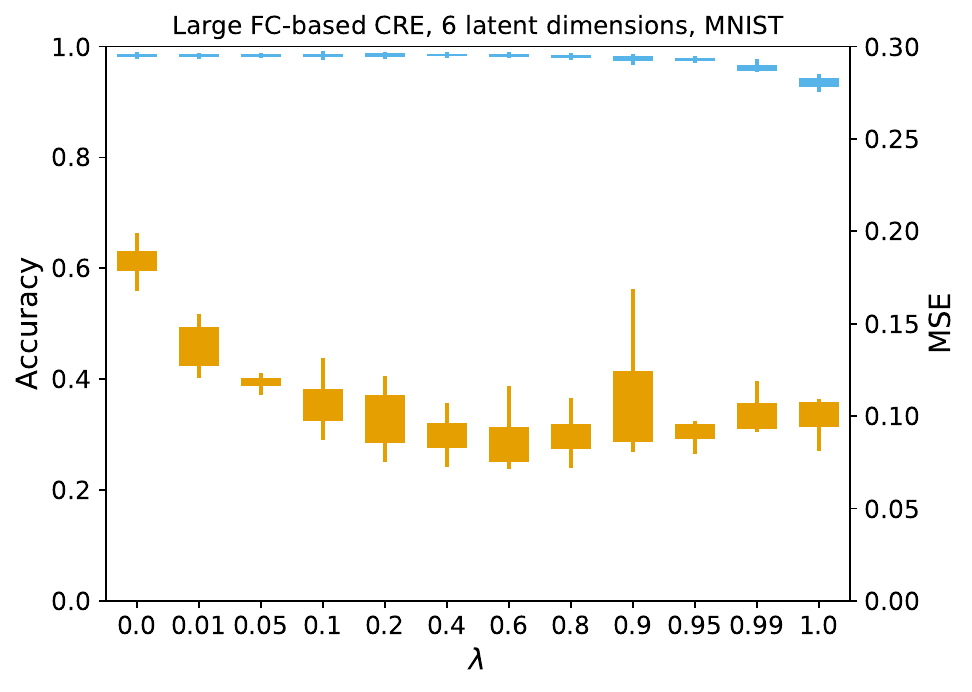}
        \includegraphics[width=1.0\linewidth]{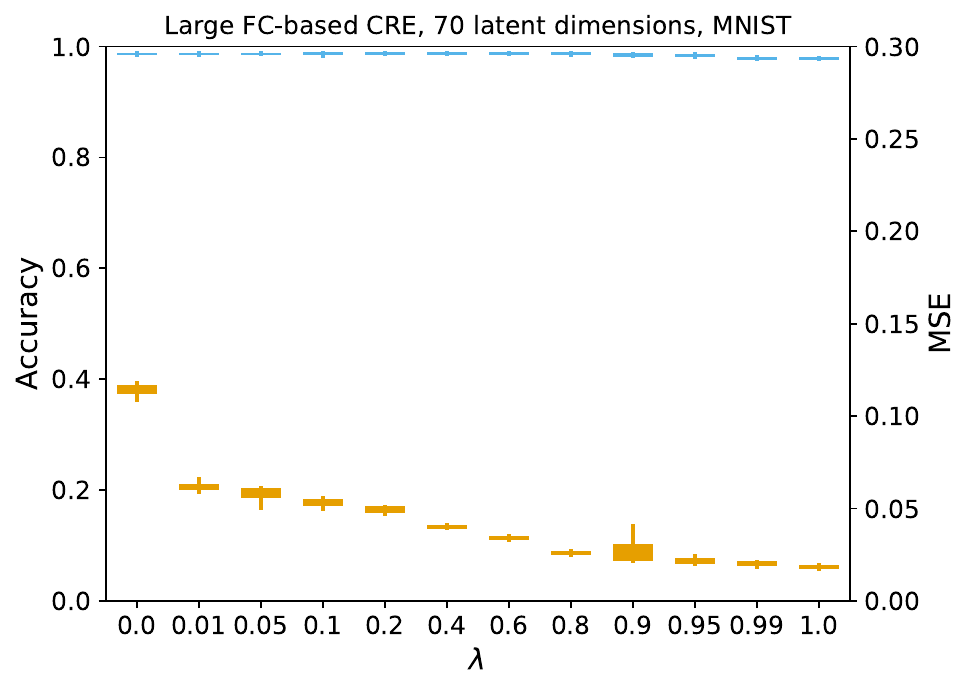}
        \includegraphics[width=1.0\linewidth]{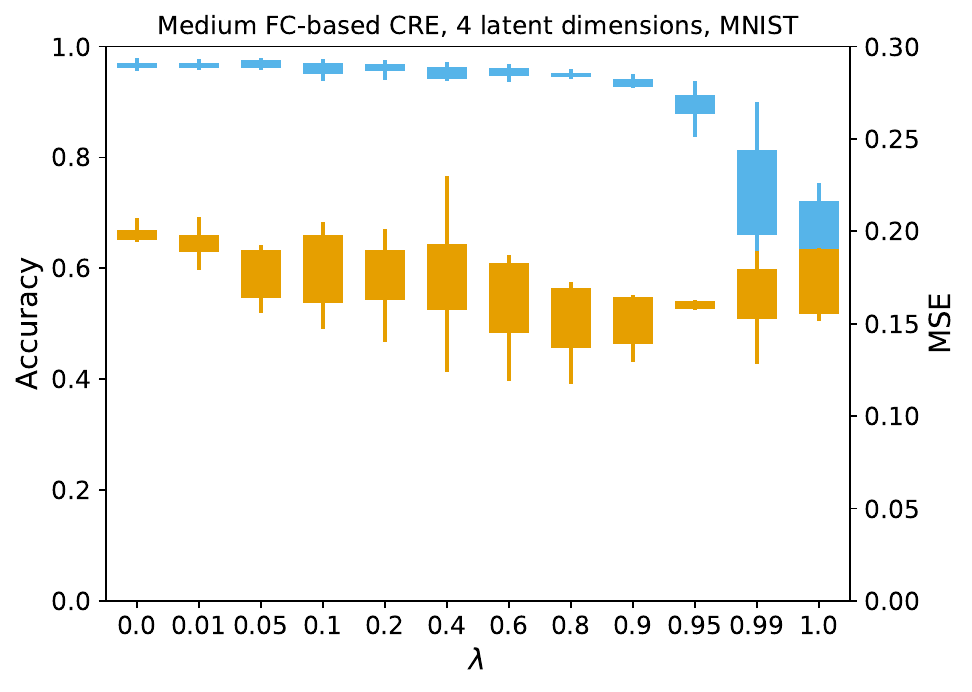}
        \includegraphics[width=1.0\linewidth]{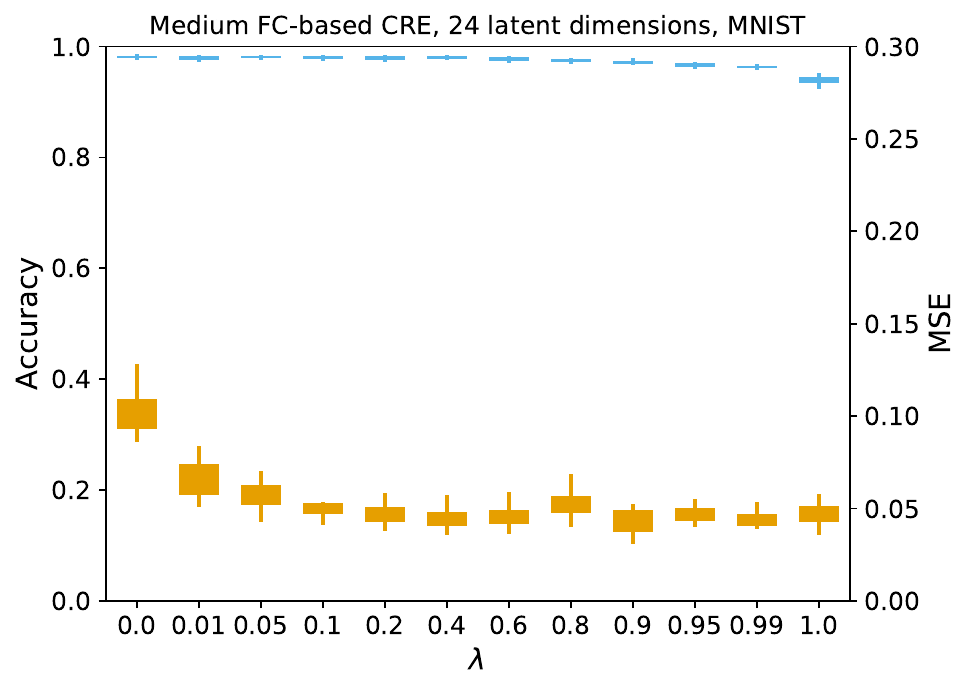}
        \includegraphics[width=1.0\linewidth]{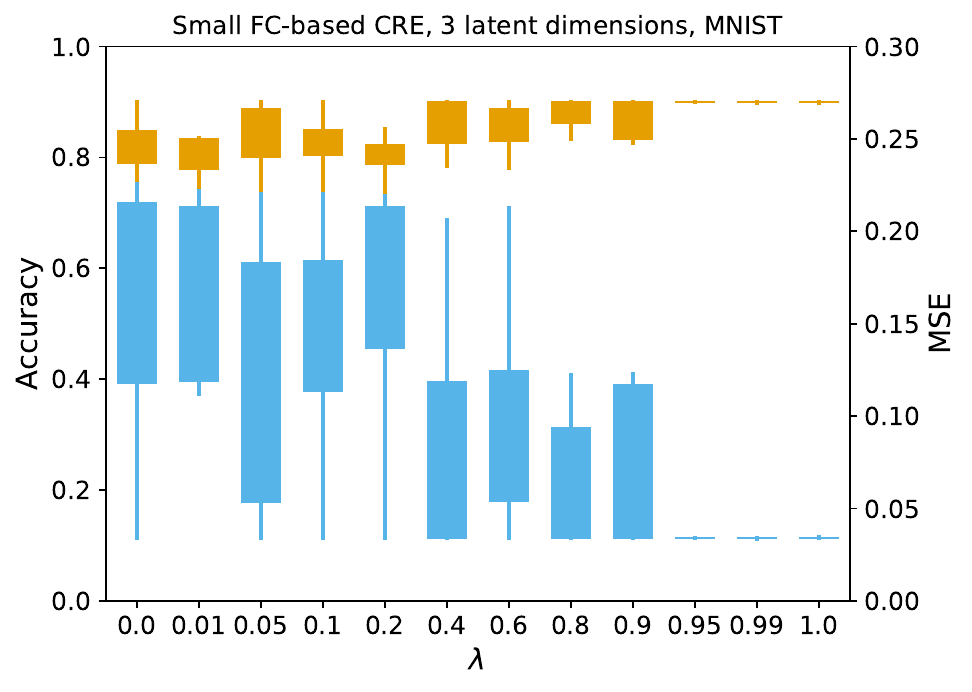}
        \includegraphics[width=1.0\linewidth]{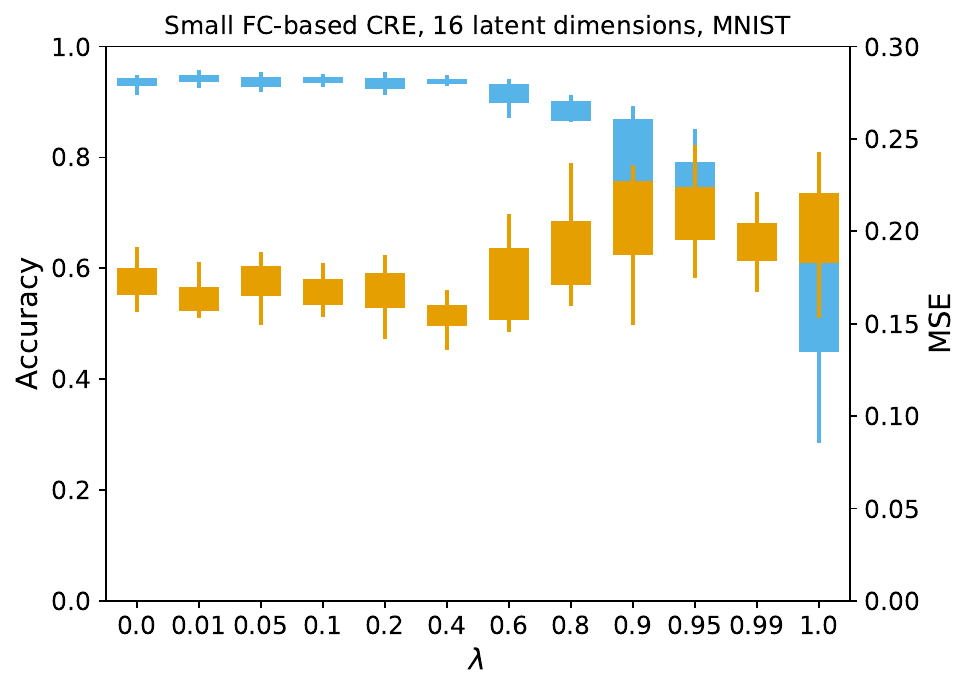}
        \includegraphics[width=1.0\linewidth]{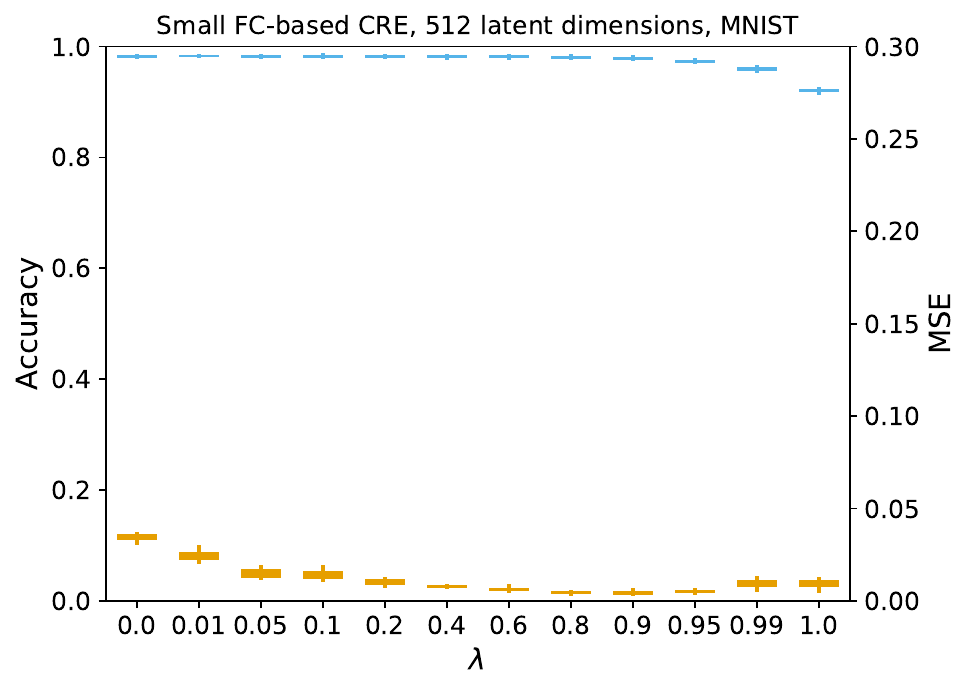}
    \end{subfigure}
    \begin{subfigure}[t]{0.246\linewidth}
        \centering
        \includegraphics[width=1.0\linewidth]{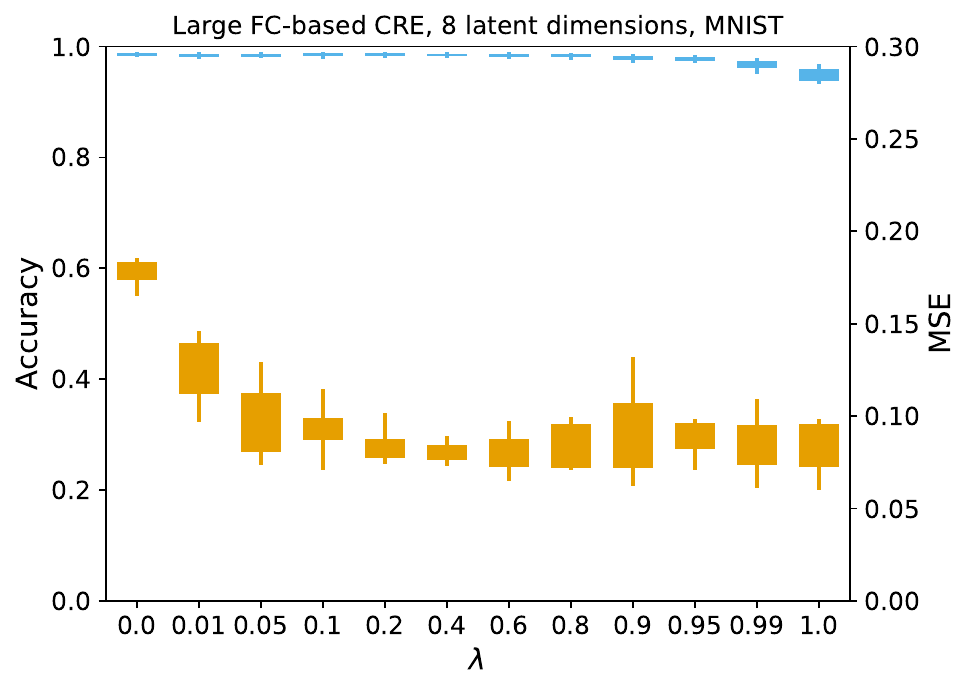}
        \includegraphics[width=1.0\linewidth]{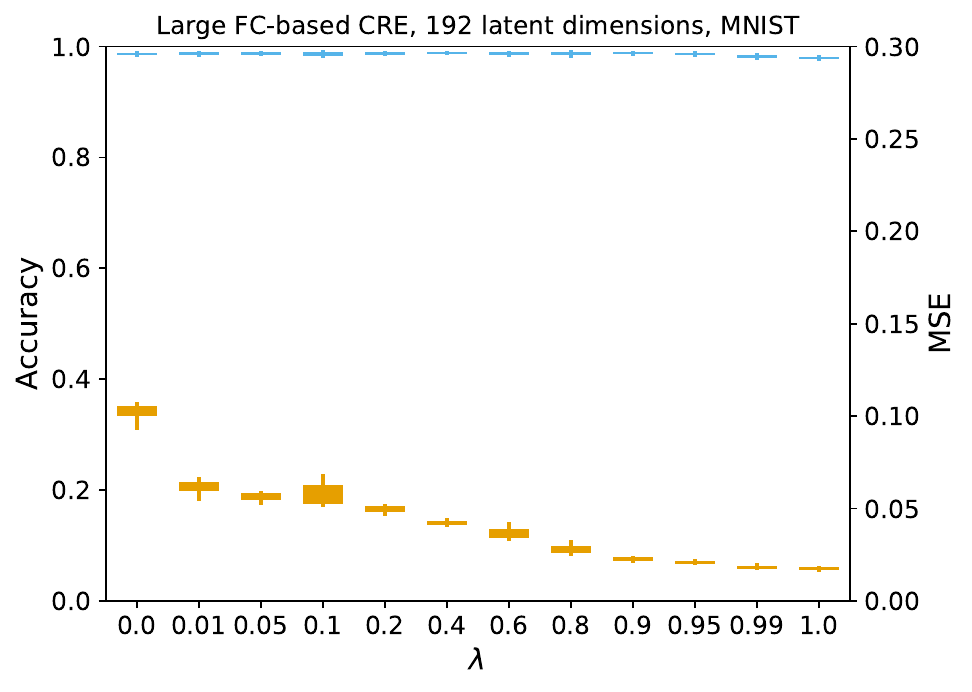}
        \includegraphics[width=1.0\linewidth]{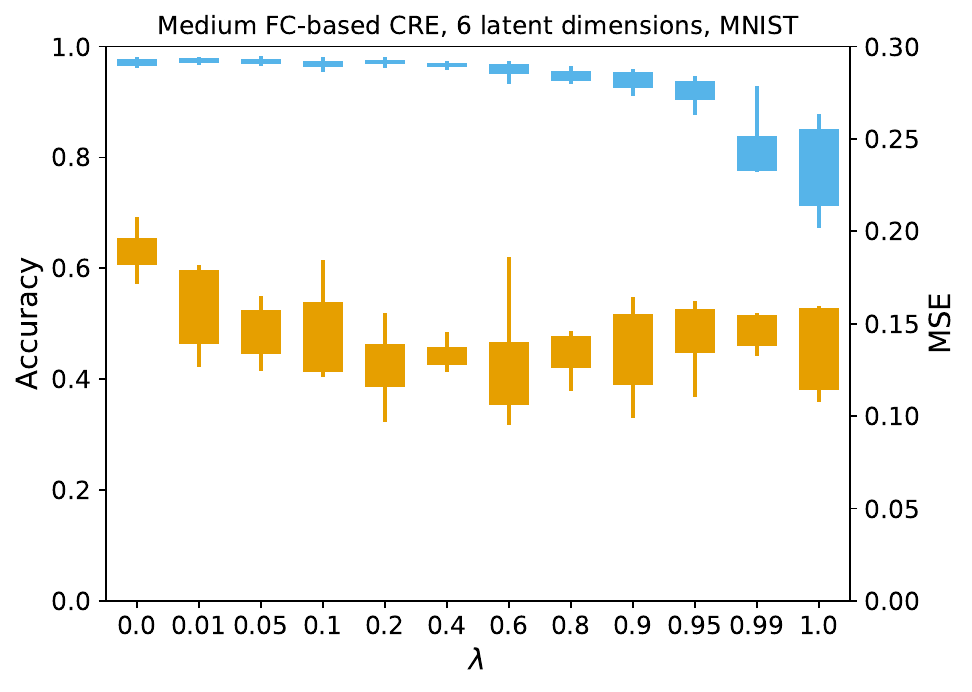}
        \includegraphics[width=1.0\linewidth]{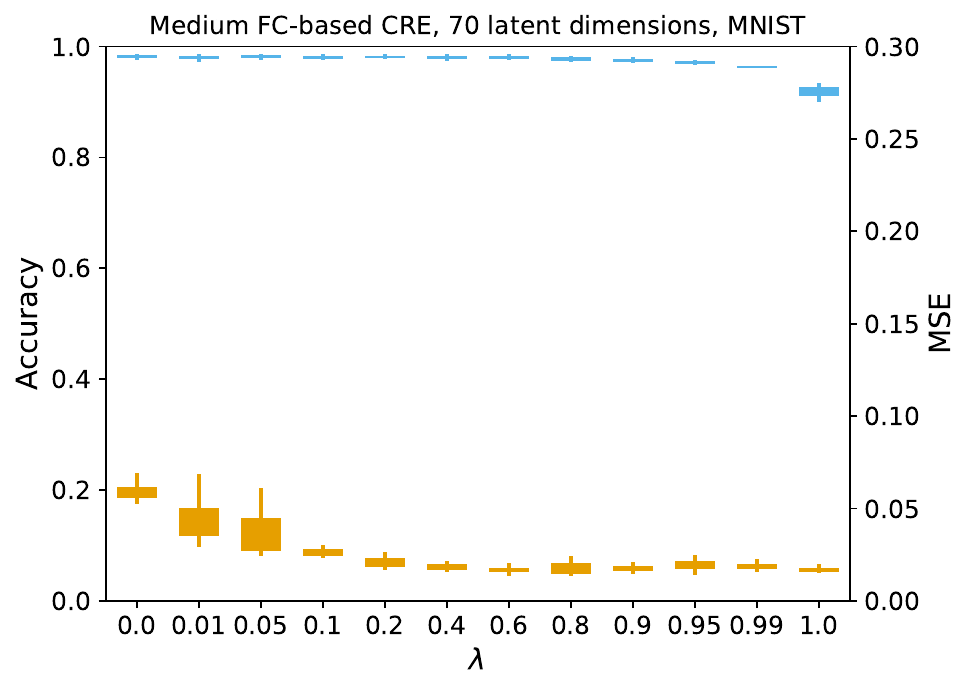}
          \includegraphics[width=1.0\linewidth]{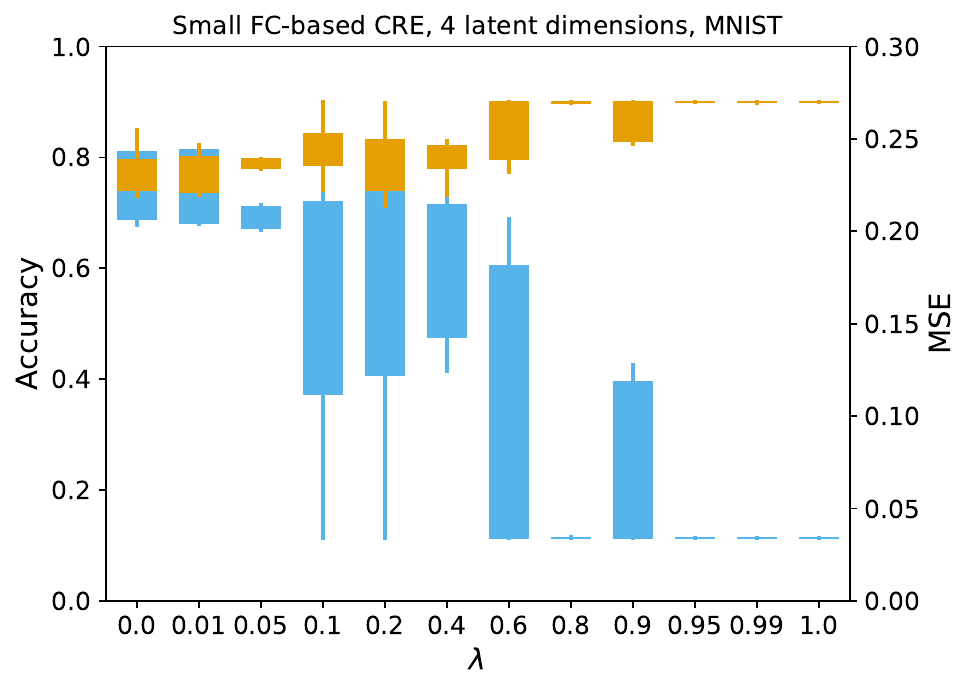}
        \includegraphics[width=1.0\linewidth]{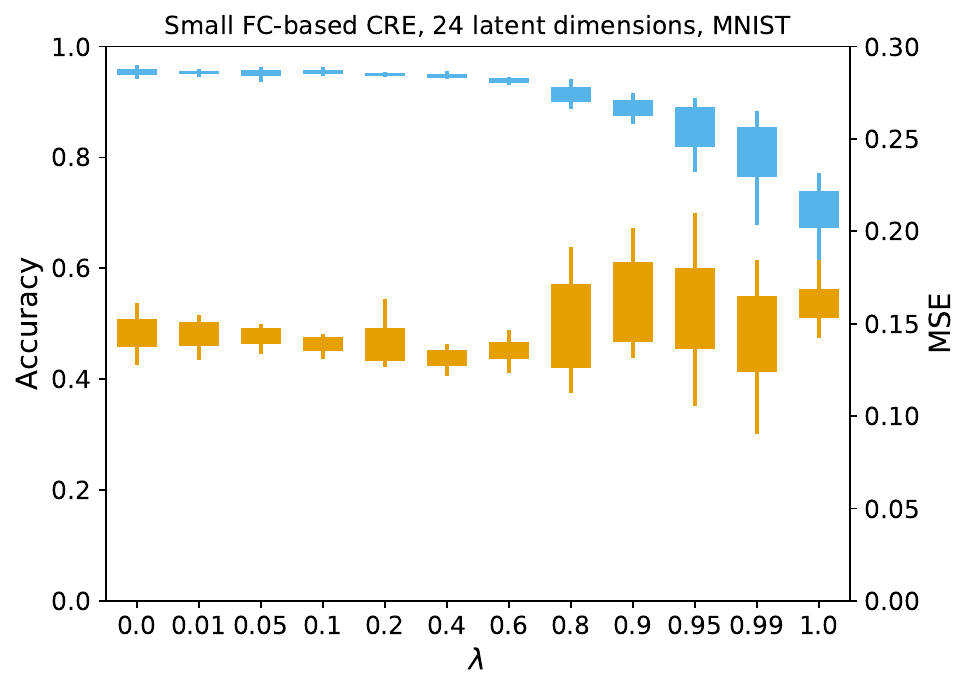}
    \end{subfigure}
    \caption{FC-based CREs on MNIST}
    \label{fig:fc_mnist}
\end{figure}

\begin{figure}[ht]
    \begin{subfigure}[t]{0.246\linewidth}
        \centering
        \includegraphics[width=1.0\linewidth]{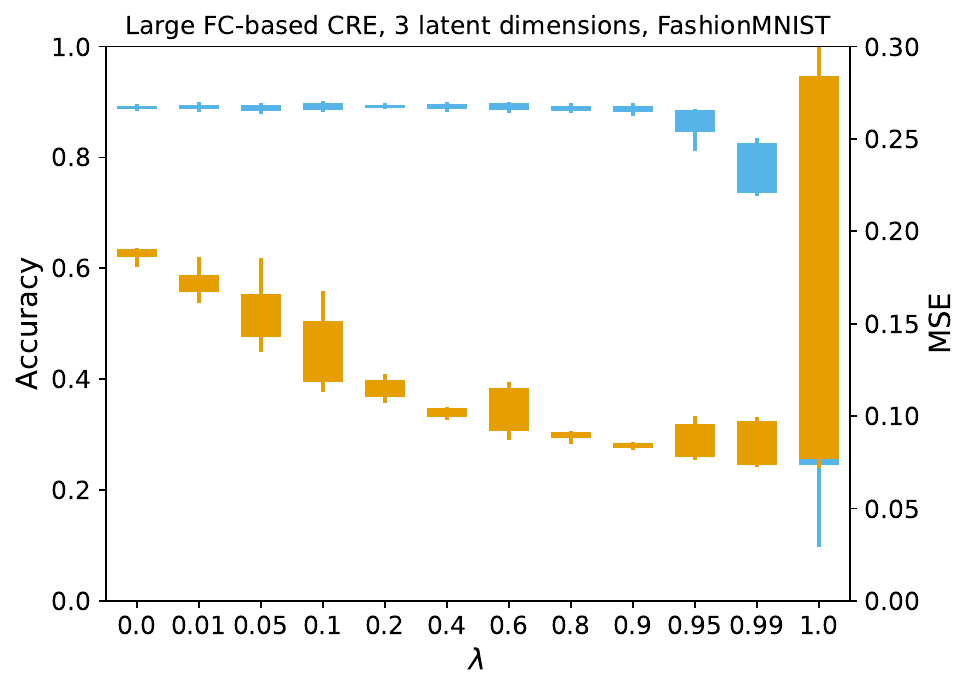}
        \includegraphics[width=1.0\linewidth]{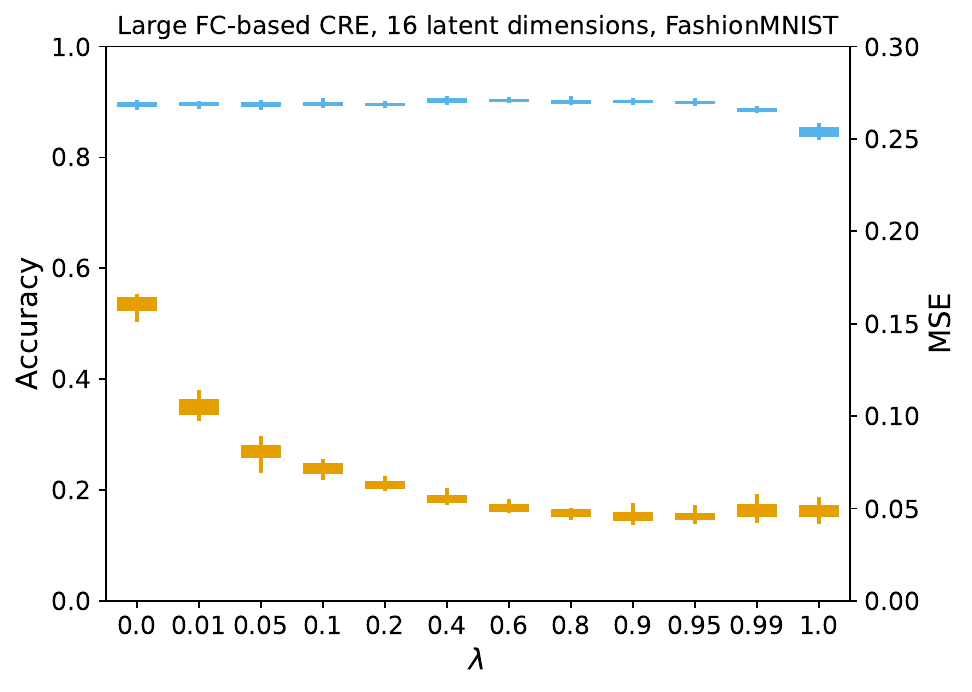}
        \includegraphics[width=1.0\linewidth]{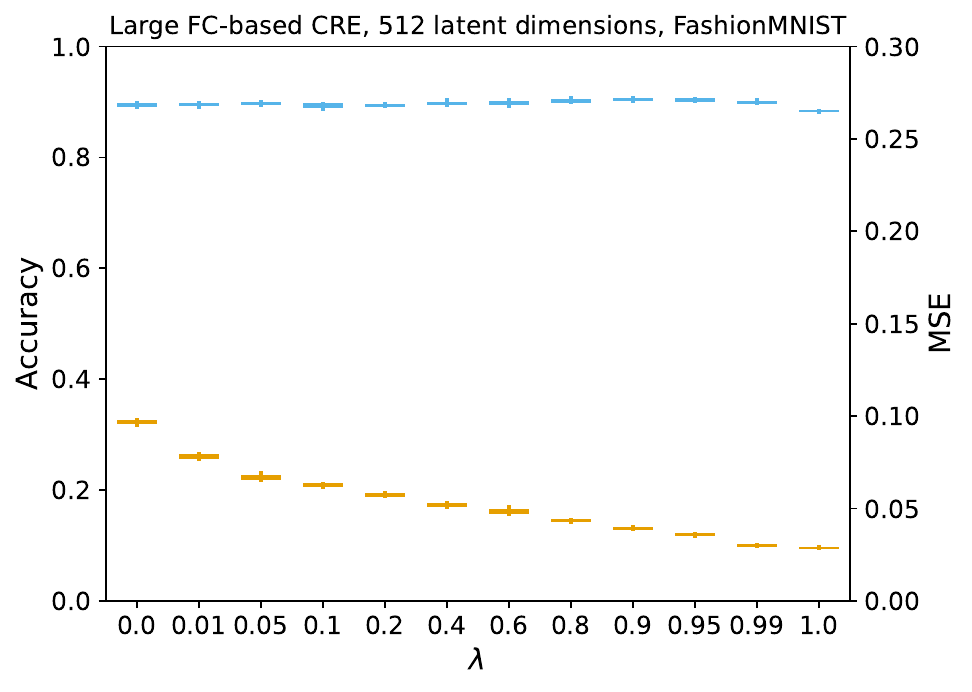}
        \includegraphics[width=1.0\linewidth]{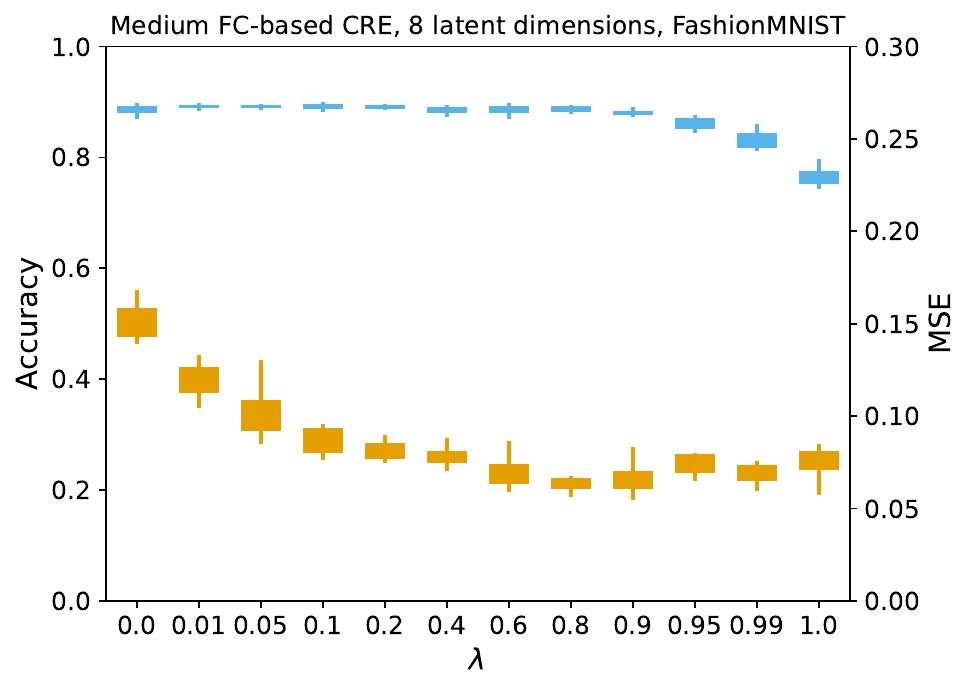}
        \includegraphics[width=1.0\linewidth]{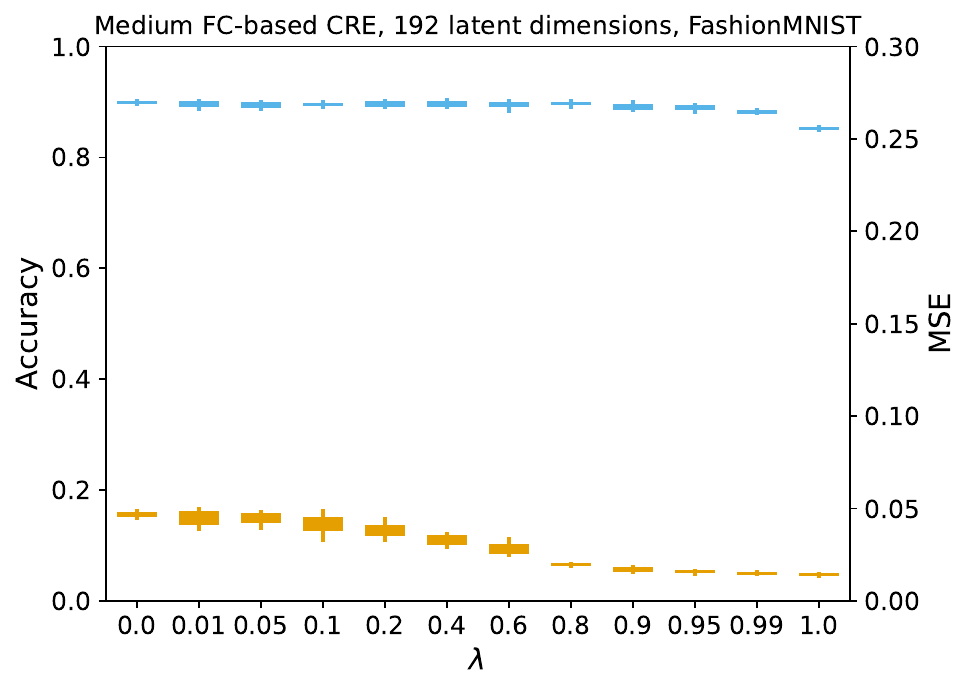}
         \includegraphics[width=1.0\linewidth]{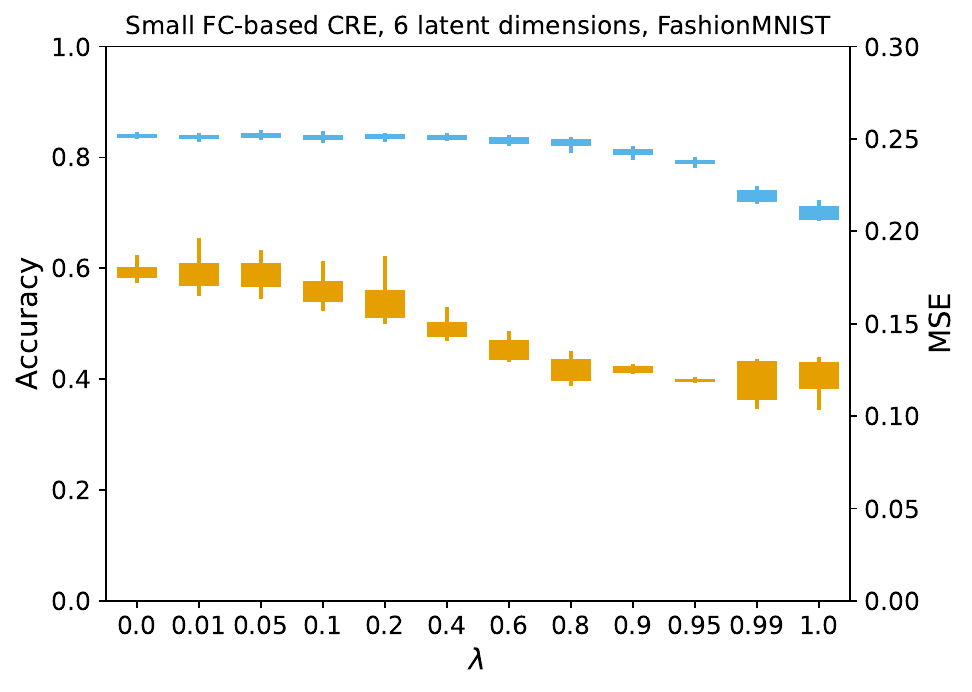}
        \includegraphics[width=1.0\linewidth]{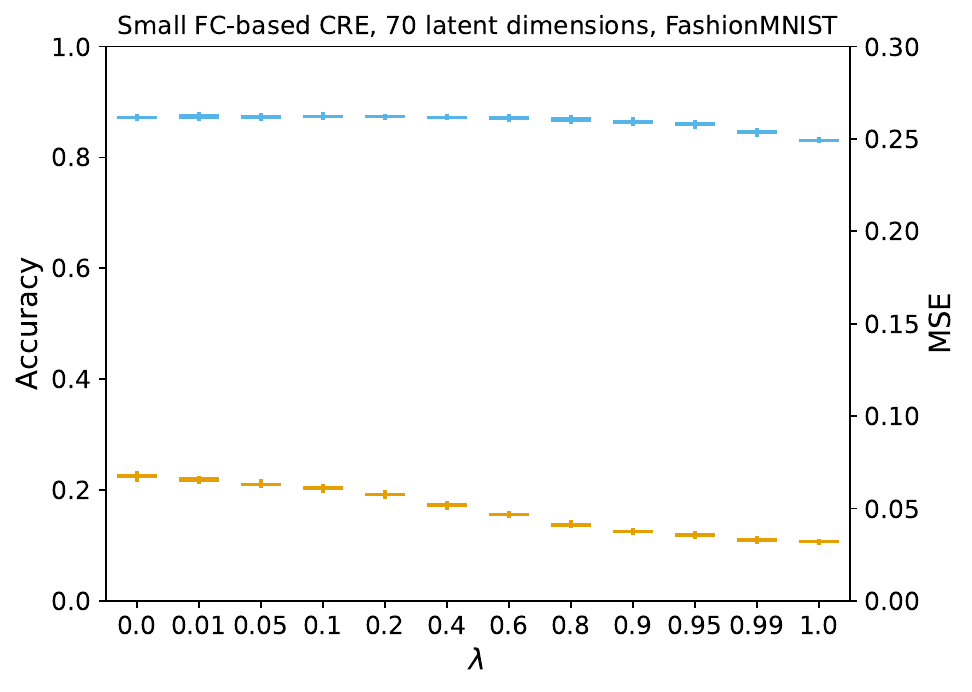}
    \end{subfigure}
    \begin{subfigure}[t]{0.246\linewidth}
        \includegraphics[width=1.0\linewidth]{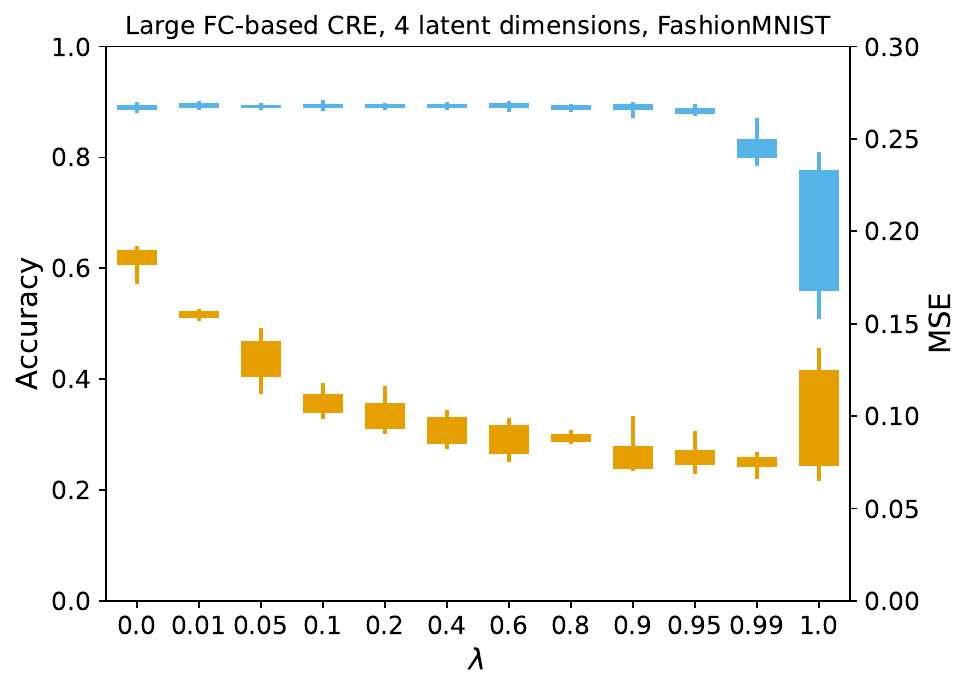}
        \includegraphics[width=1.0\linewidth]{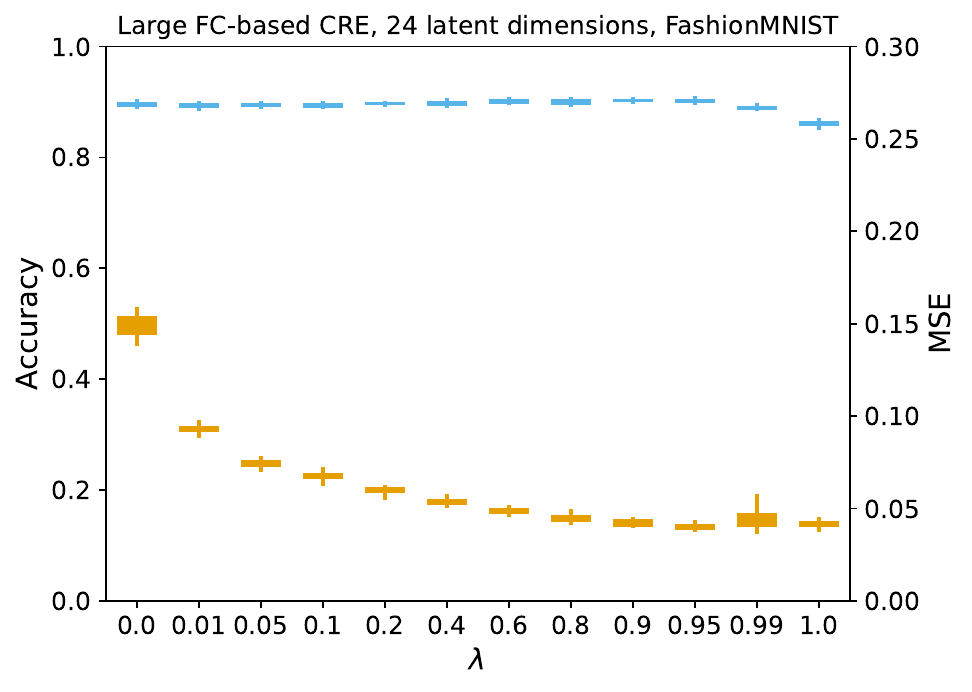}
        \includegraphics[width=1.0\linewidth]{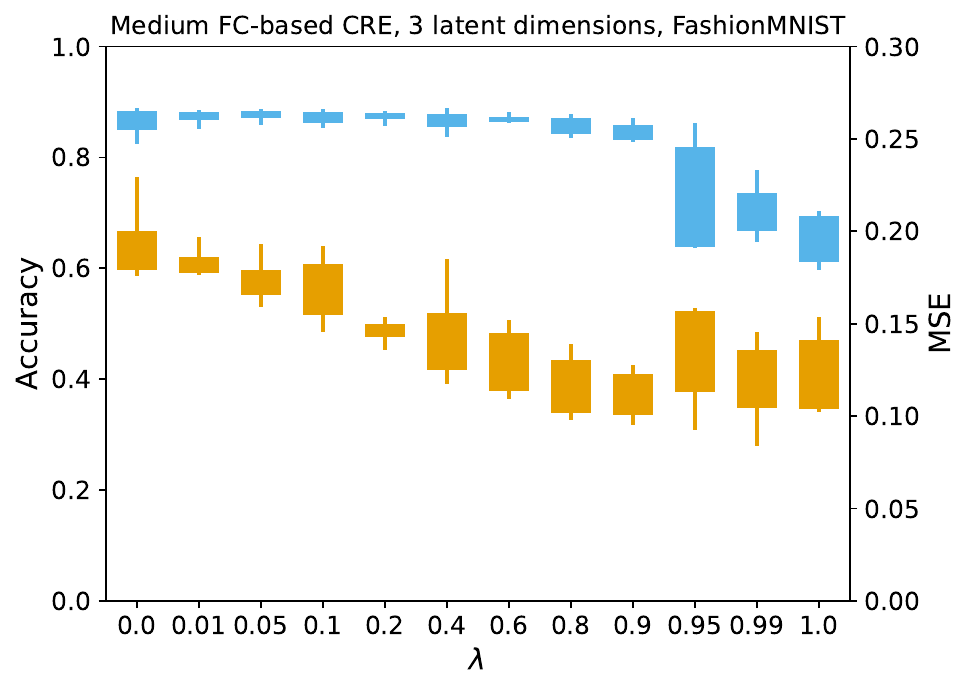}
        \includegraphics[width=1.0\linewidth]{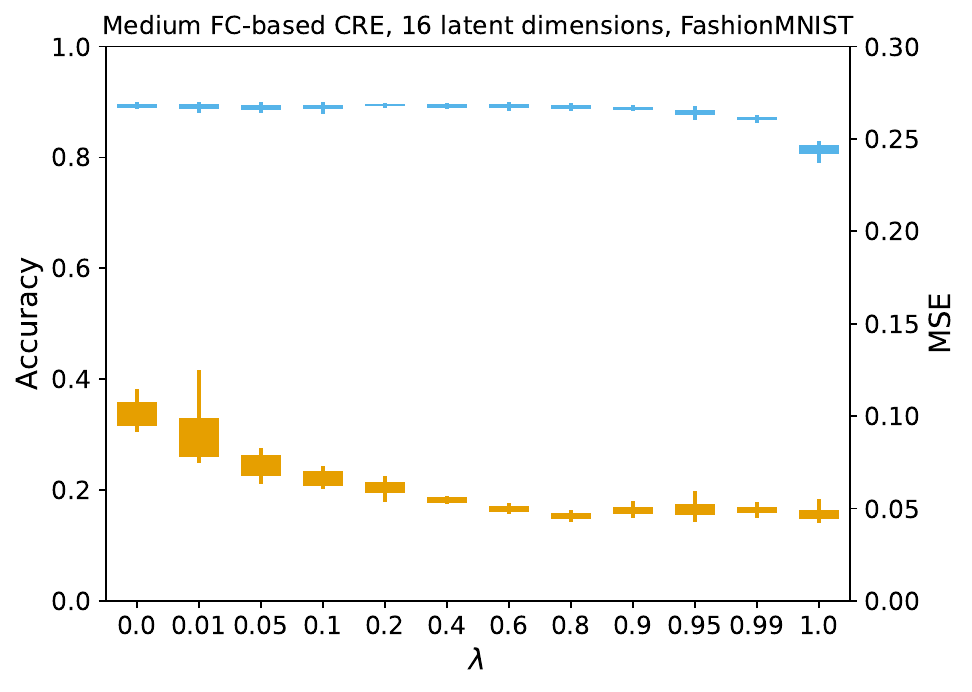}
        \includegraphics[width=1.0\linewidth]{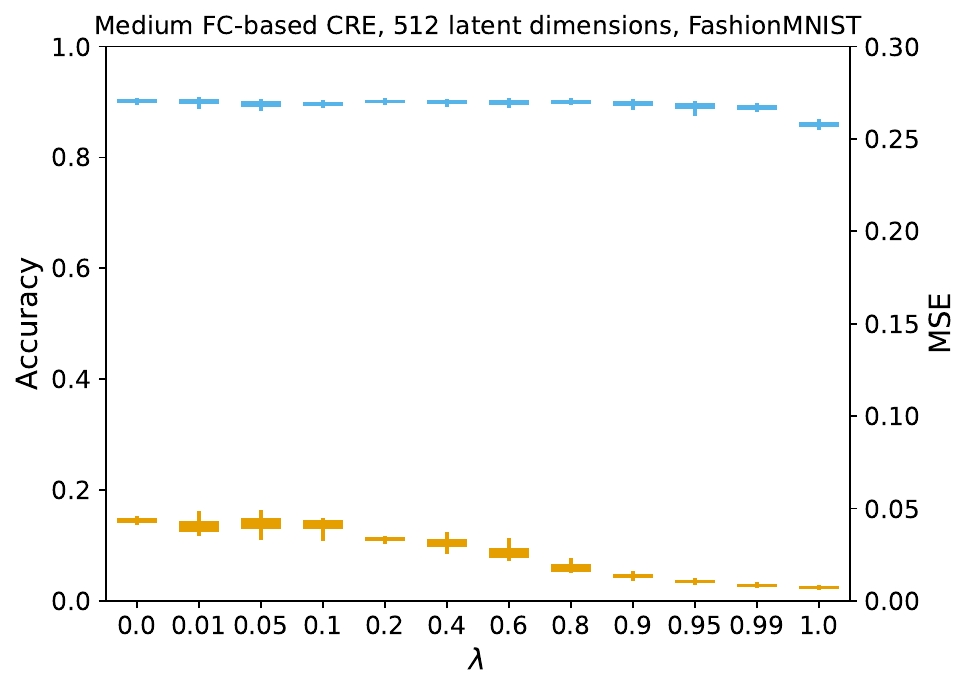}
        \includegraphics[width=1.0\linewidth]{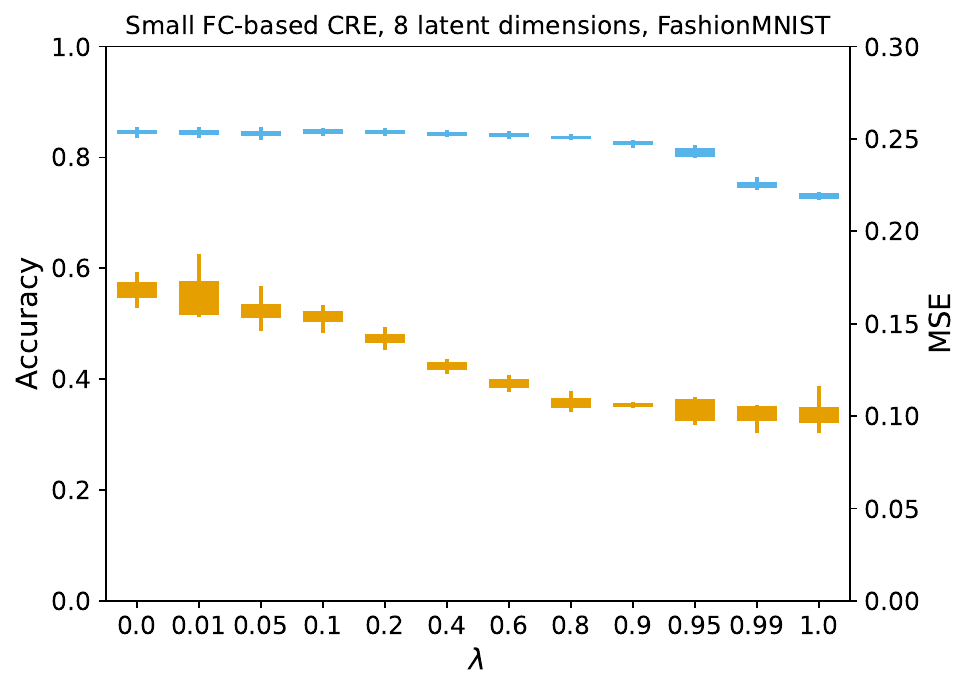}
        \includegraphics[width=1.0\linewidth]{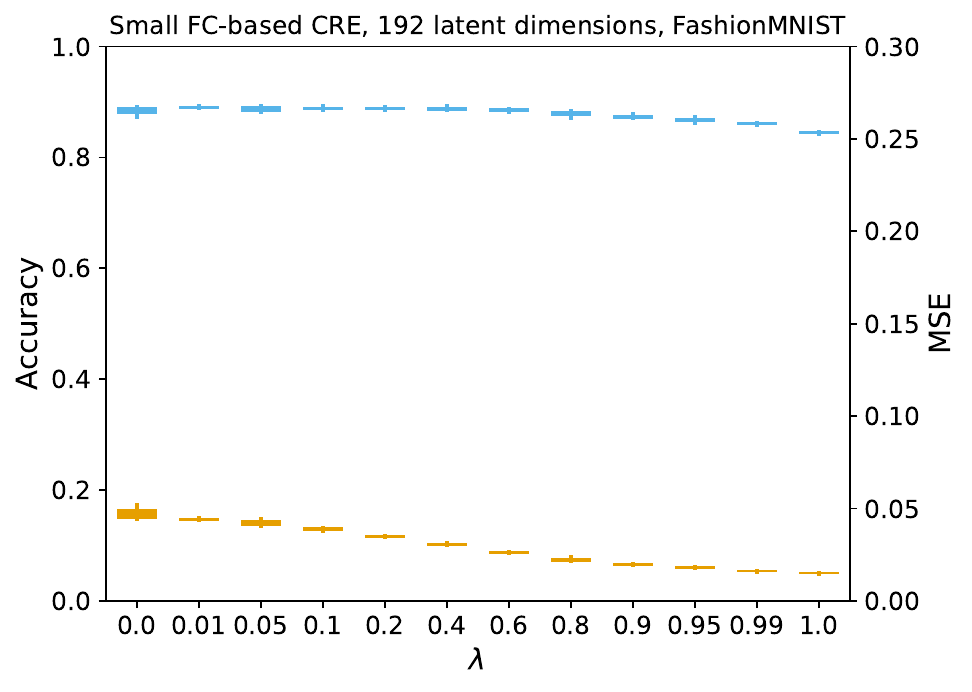}

    \end{subfigure}
    \begin{subfigure}[t]{0.246\linewidth}
        \centering
        \includegraphics[width=1.0\linewidth]{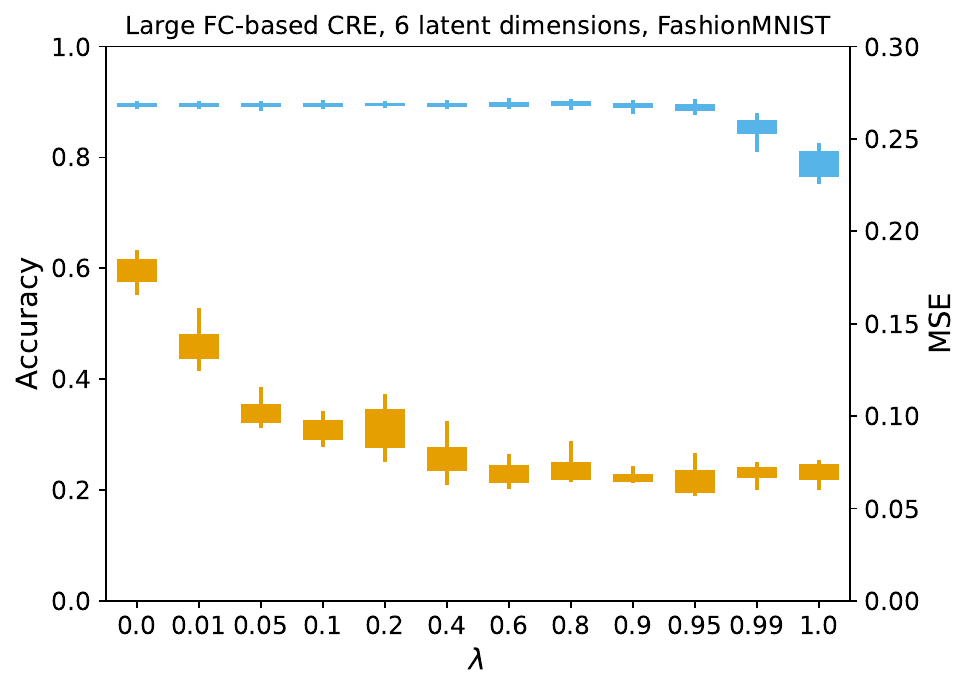}
        \includegraphics[width=1.0\linewidth]{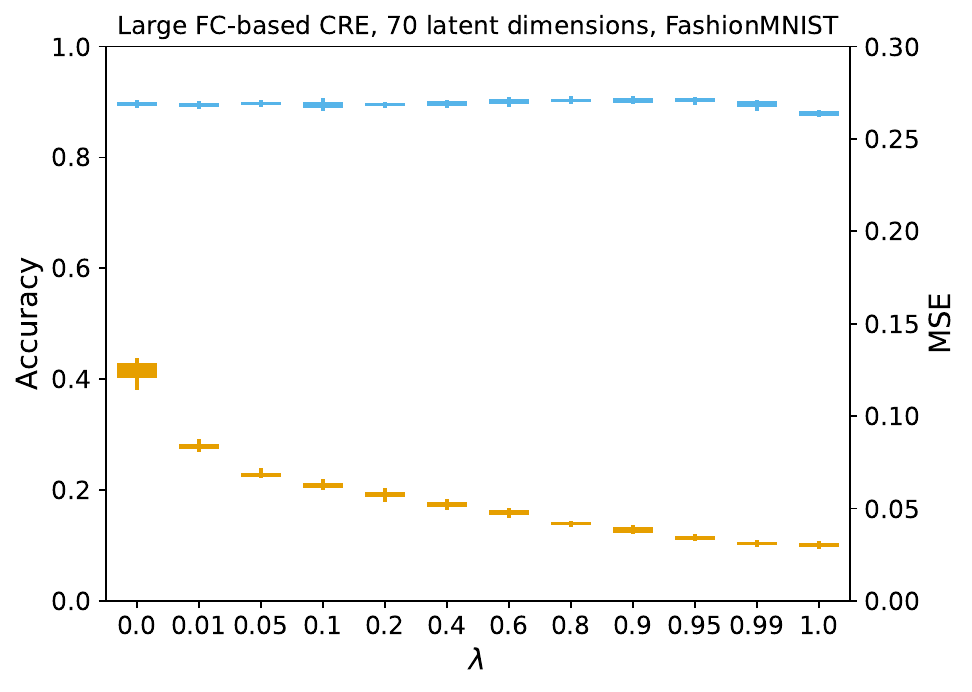}
        \includegraphics[width=1.0\linewidth]{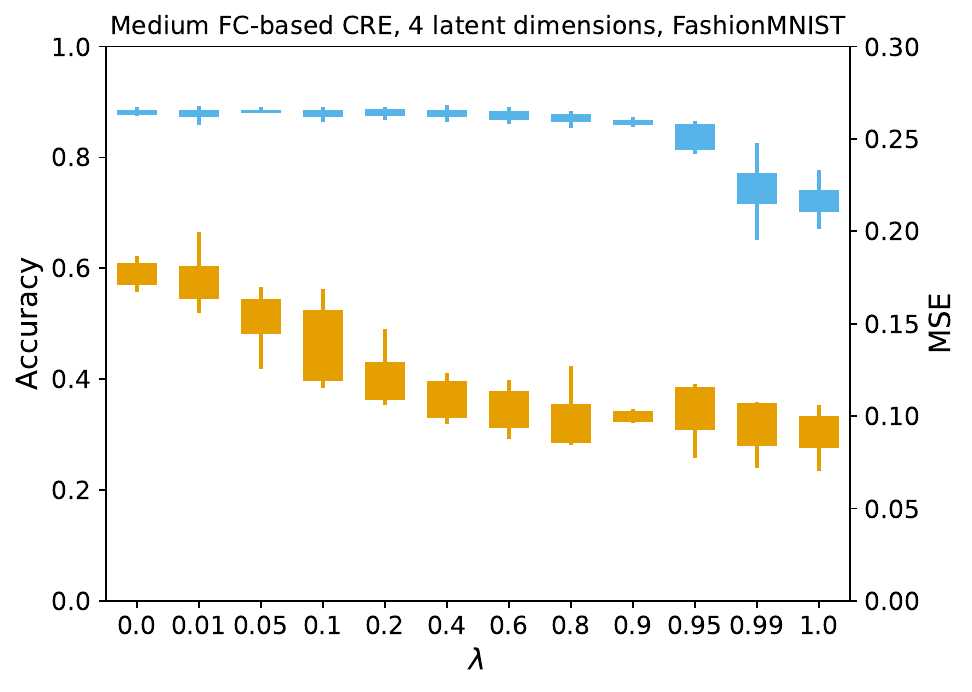}
        \includegraphics[width=1.0\linewidth]{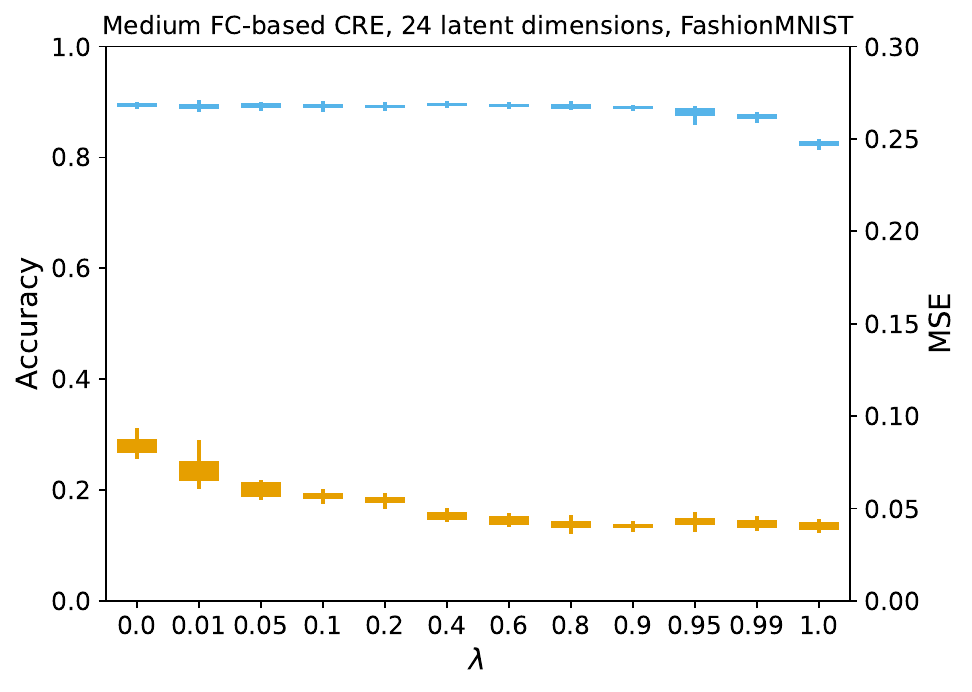}
        \includegraphics[width=1.0\linewidth]{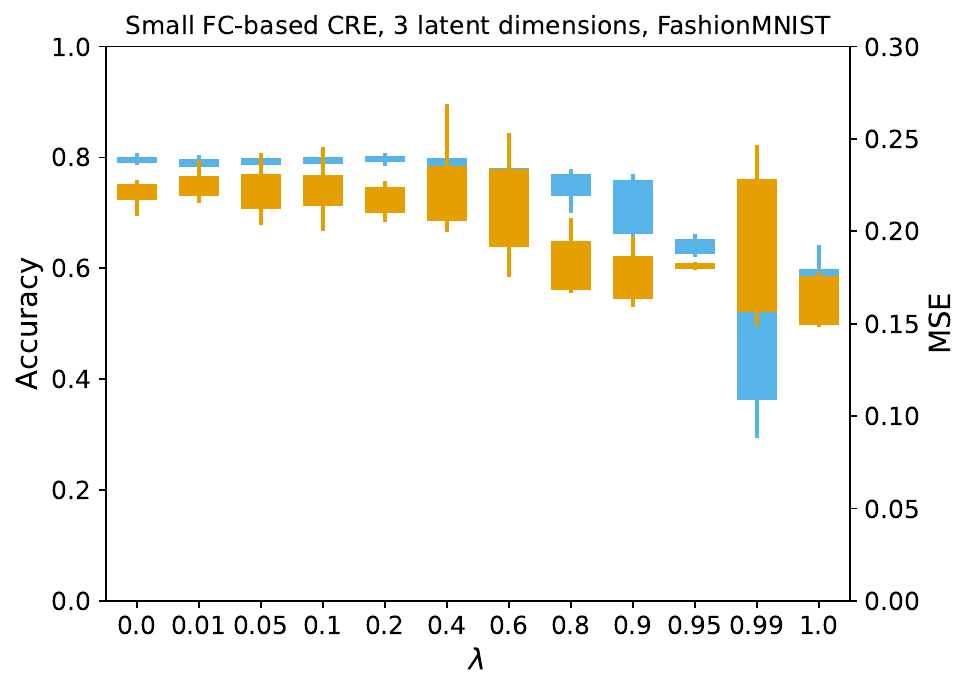}
        \includegraphics[width=1.0\linewidth]{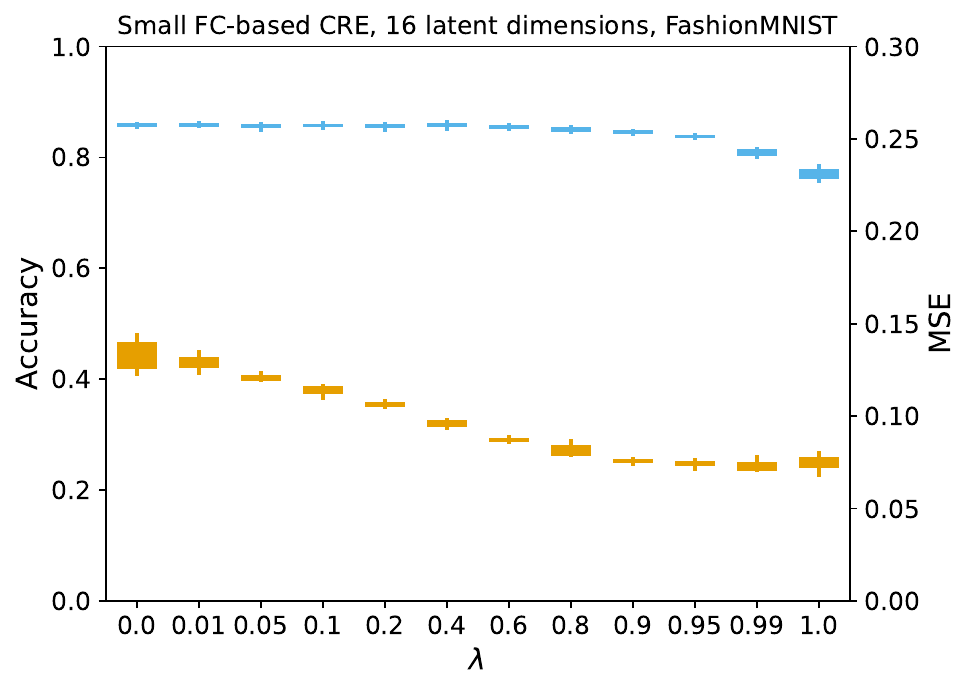}
        \includegraphics[width=1.0\linewidth]{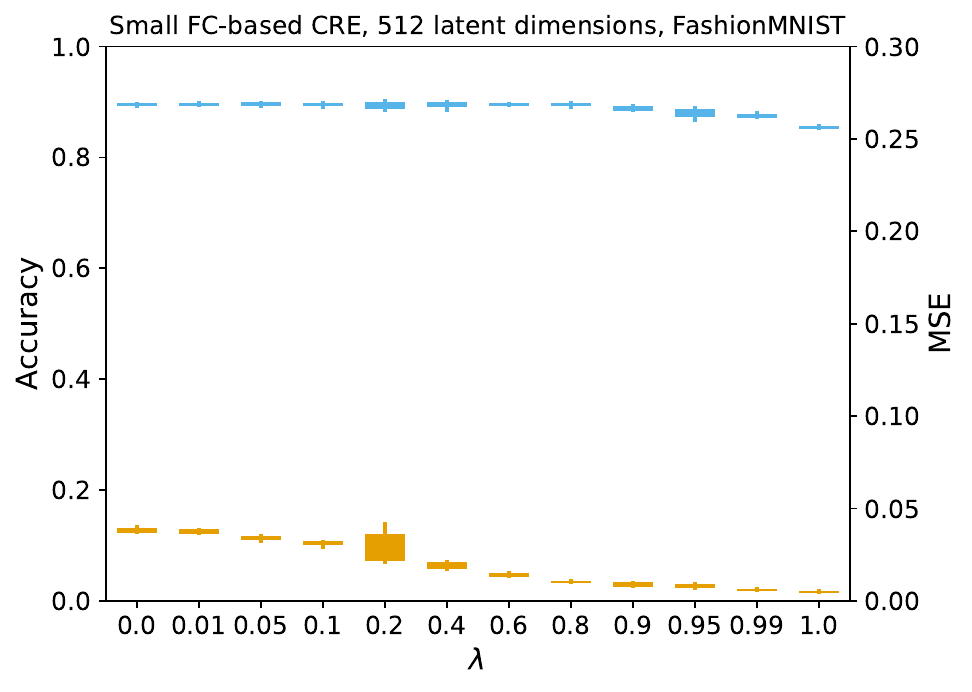}
    \end{subfigure}
    \begin{subfigure}[t]{0.246\linewidth}
        \centering
        \includegraphics[width=1.0\linewidth]{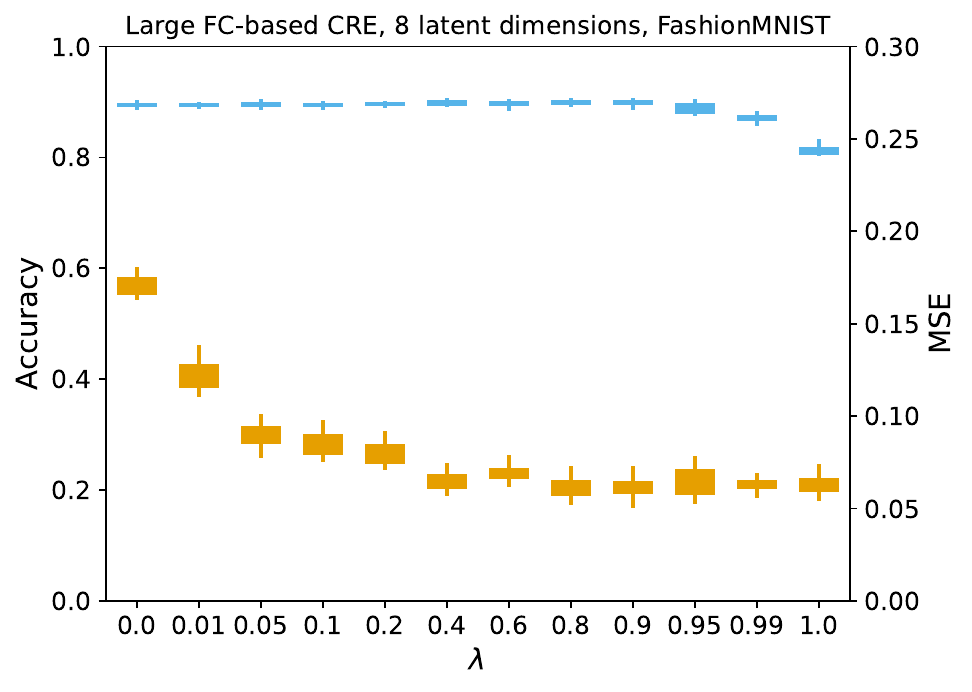}
        \includegraphics[width=1.0\linewidth]{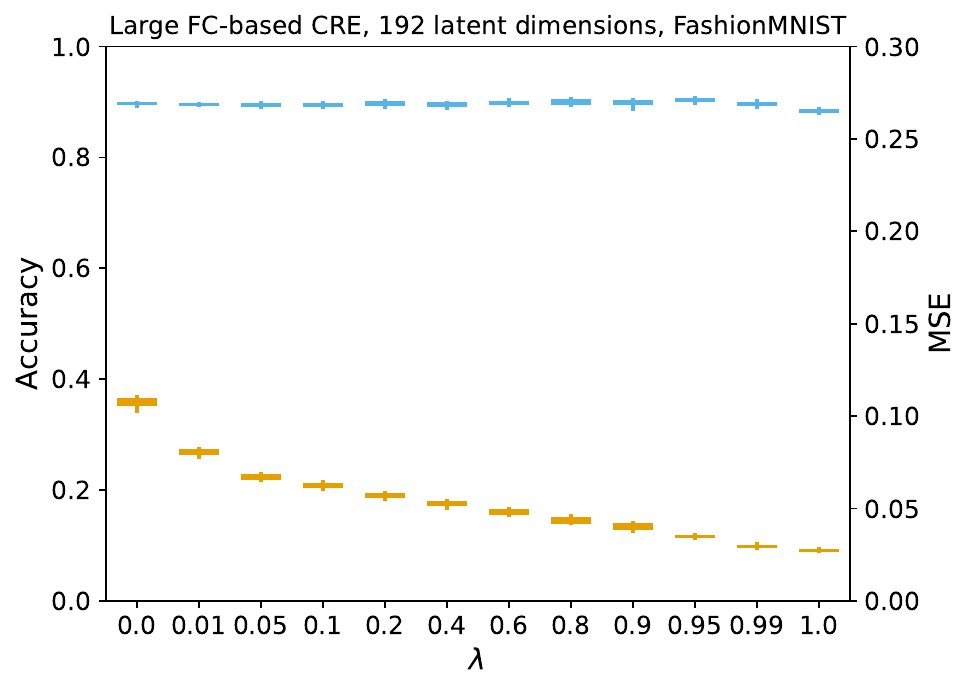}
        \includegraphics[width=1.0\linewidth]{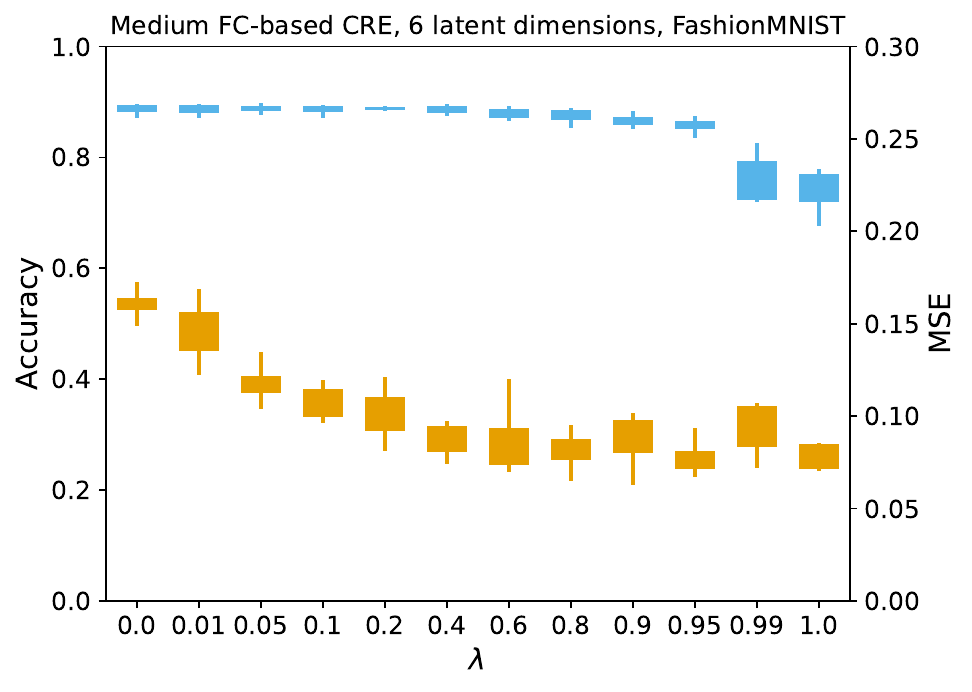}
        \includegraphics[width=1.0\linewidth]{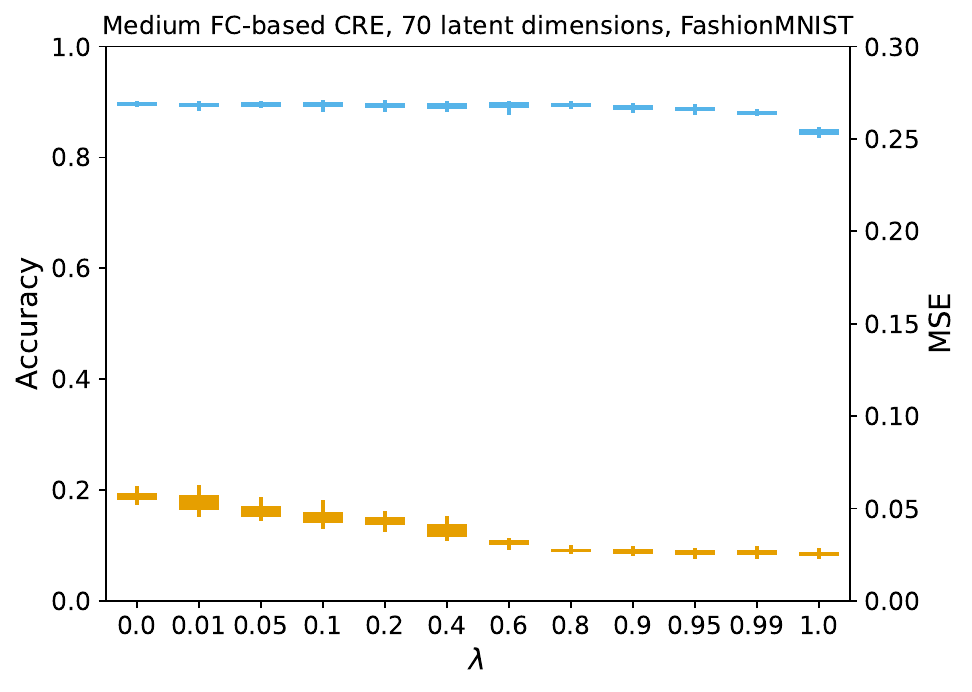}
          \includegraphics[width=1.0\linewidth]{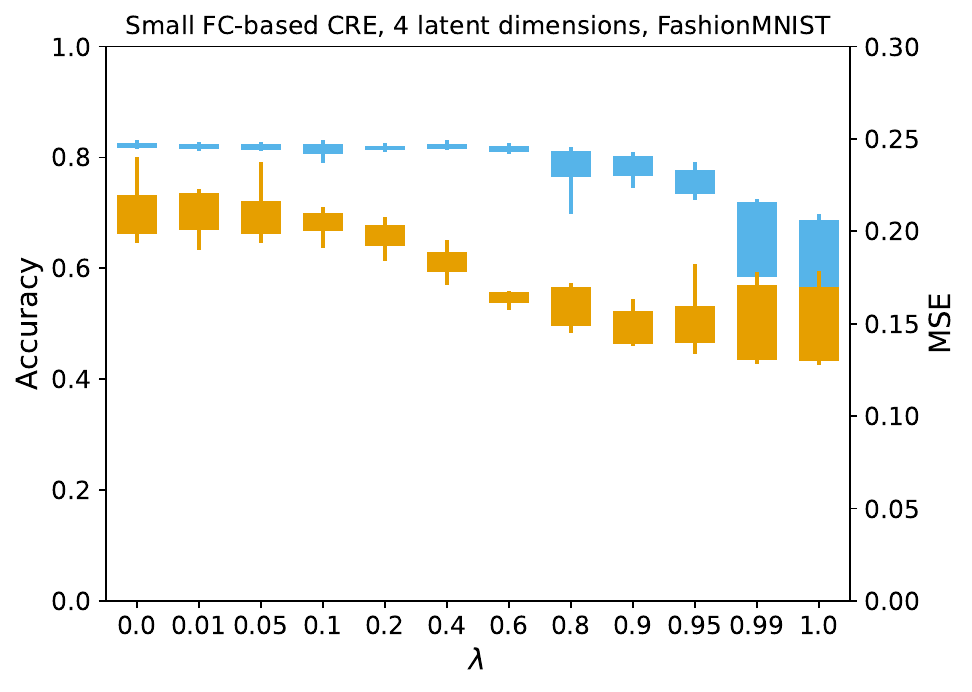}
        \includegraphics[width=1.0\linewidth]{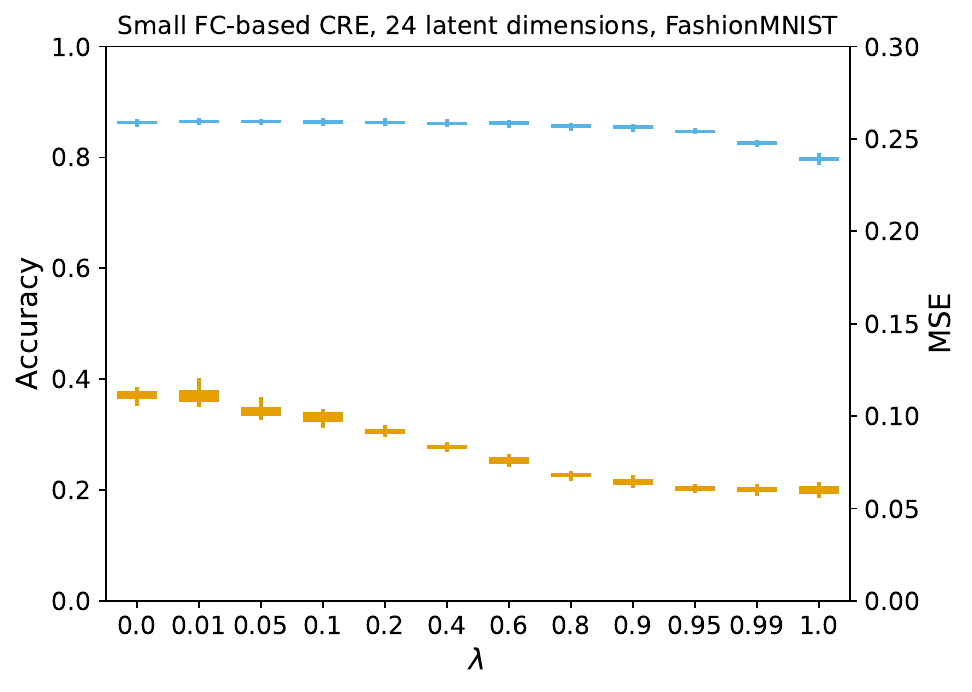}
    \end{subfigure}
    \caption{FC-based CREs on FashionMNIST}
    \label{fig:fc_fmnist}
\end{figure}

\begin{figure}[ht]
    \begin{subfigure}[t]{0.246\linewidth}
        \centering
        \includegraphics[width=1.0\linewidth]{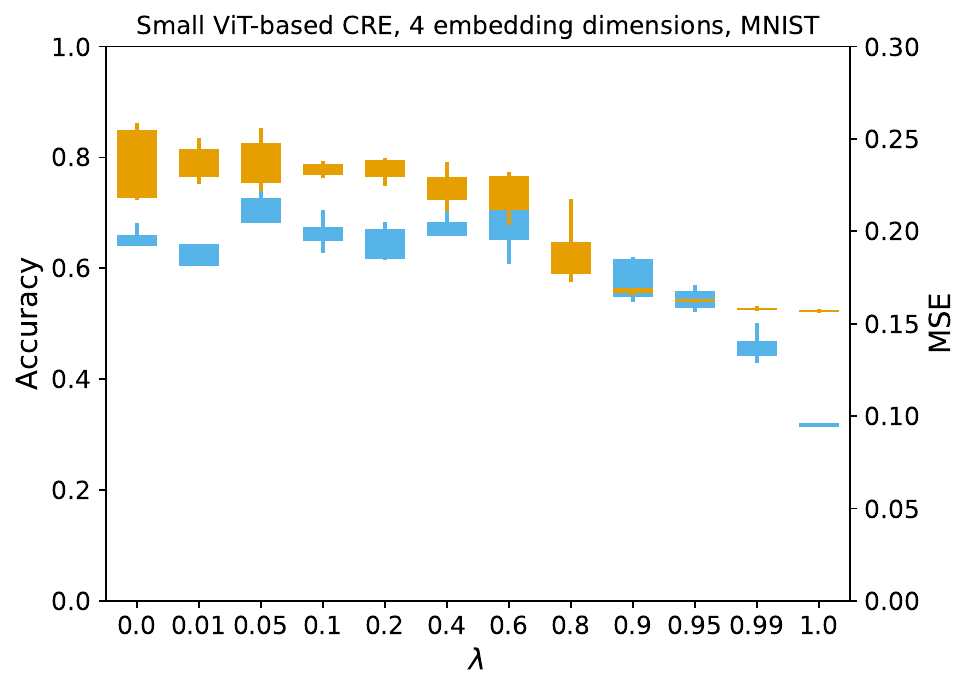}
        \includegraphics[width=1.0\linewidth]{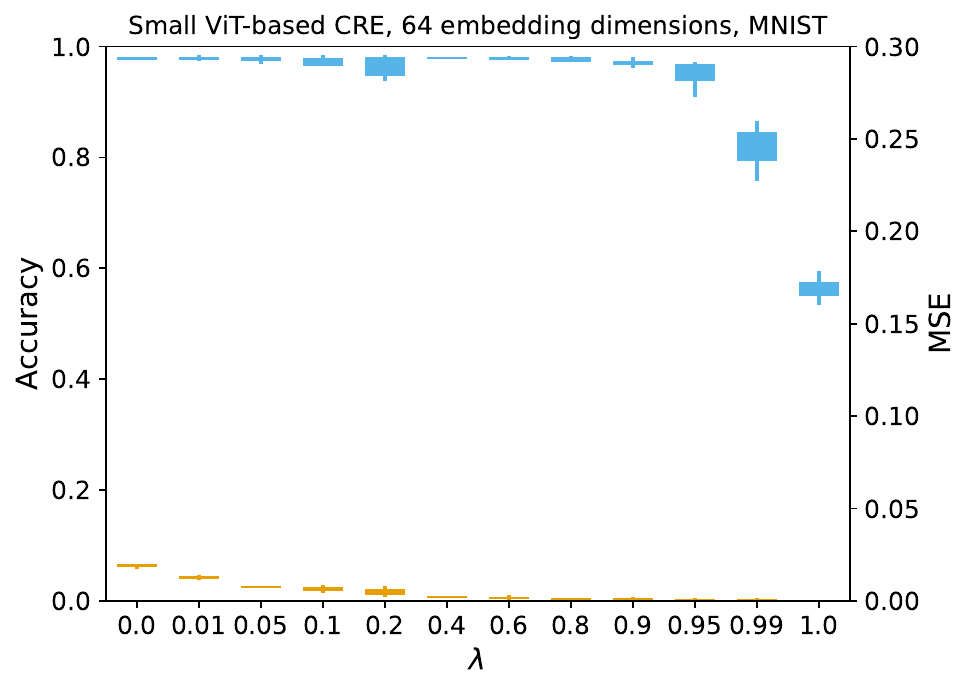}
        \includegraphics[width=1.0\linewidth]{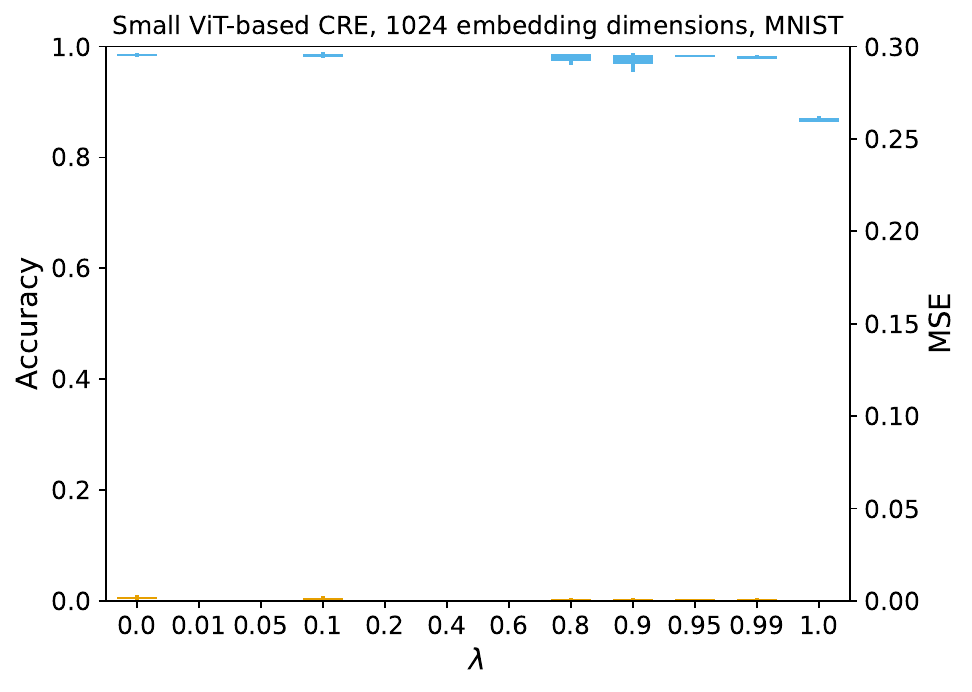}
    \end{subfigure}
    \begin{subfigure}[t]{0.246\linewidth}
        \includegraphics[width=1.0\linewidth]{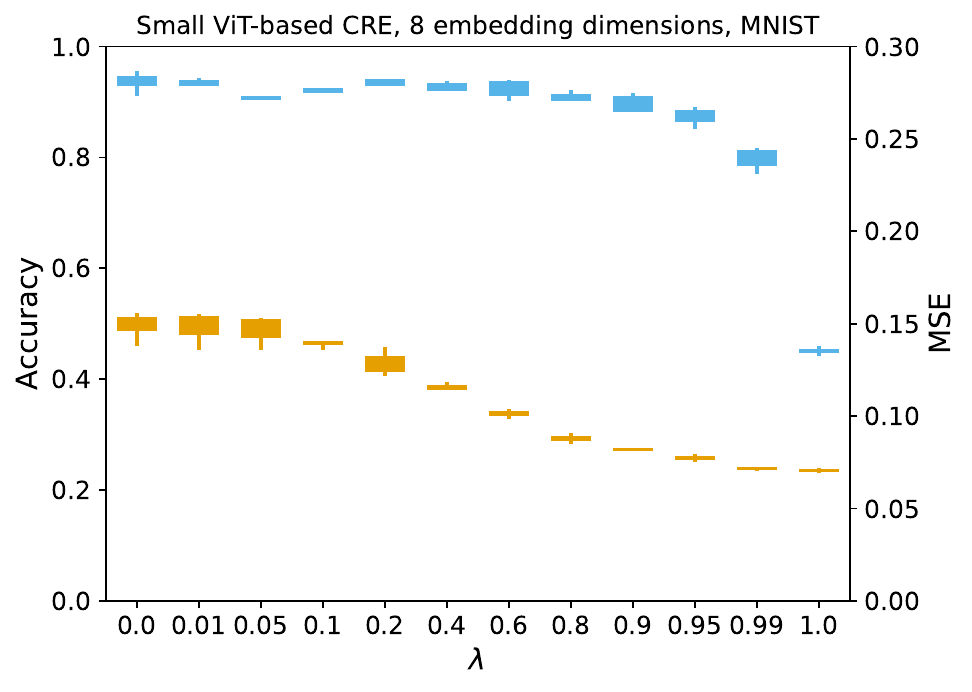}
        \includegraphics[width=1.0\linewidth]{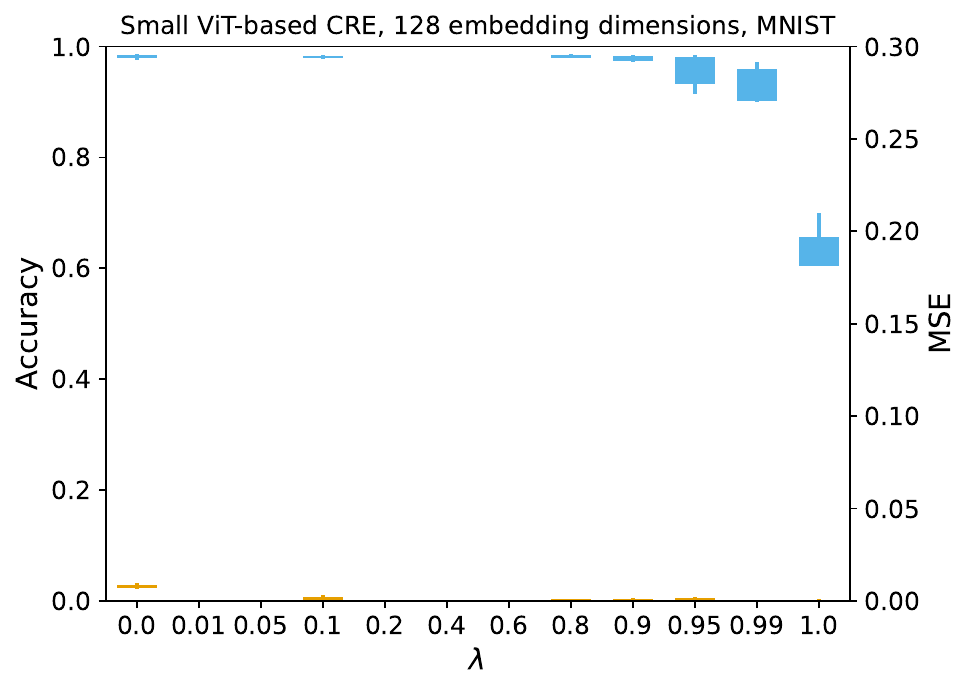}
        \includegraphics[width=1.0\linewidth]{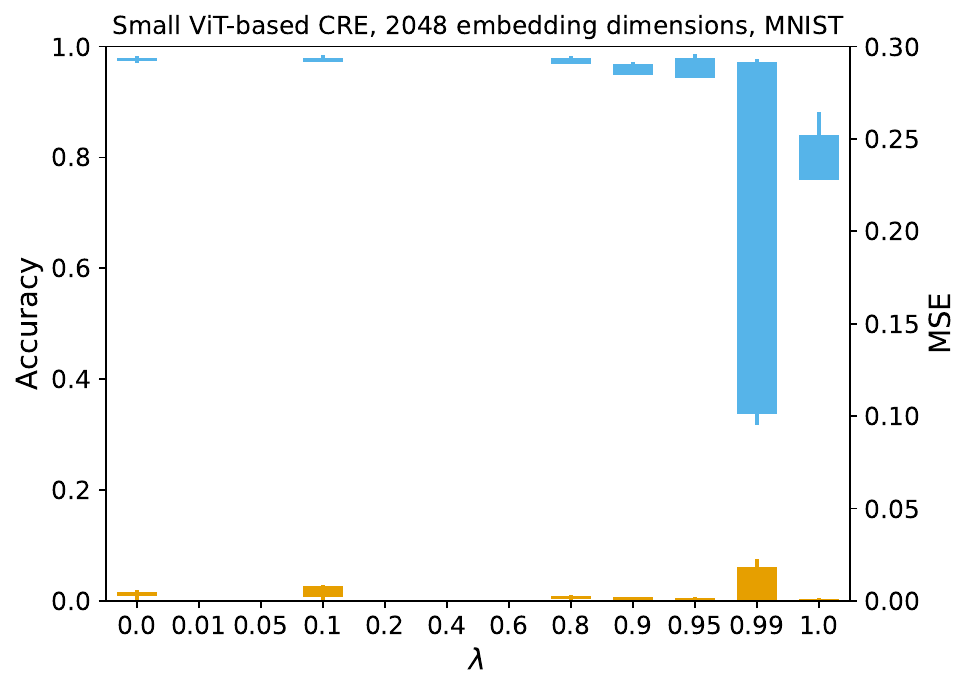}       
    \end{subfigure}
    \begin{subfigure}[t]{0.246\linewidth}
        \centering
        \includegraphics[width=1.0\linewidth]{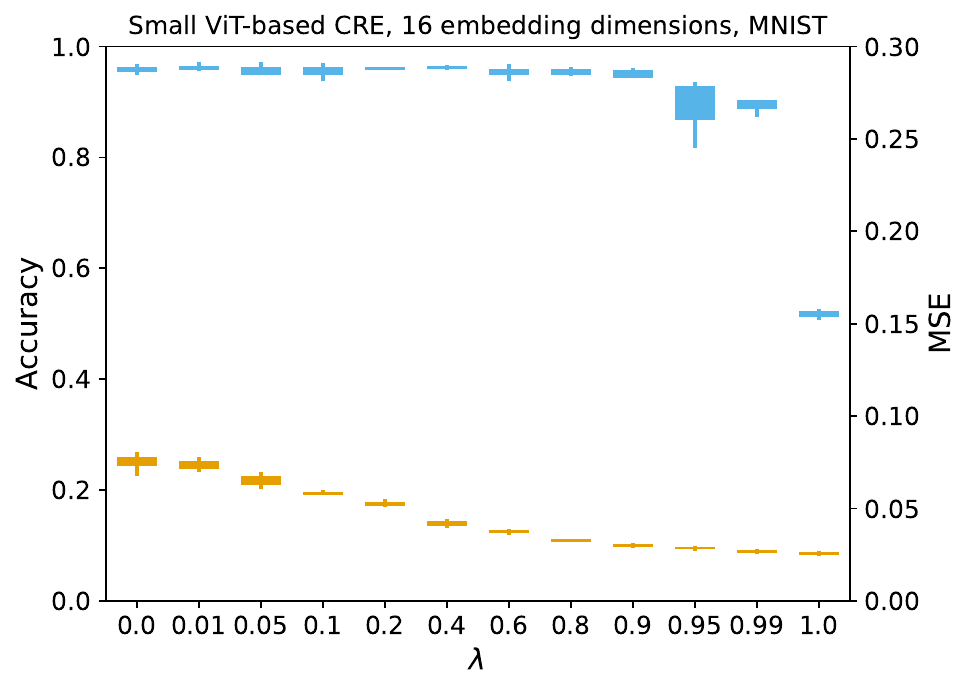}
        \includegraphics[width=1.0\linewidth]{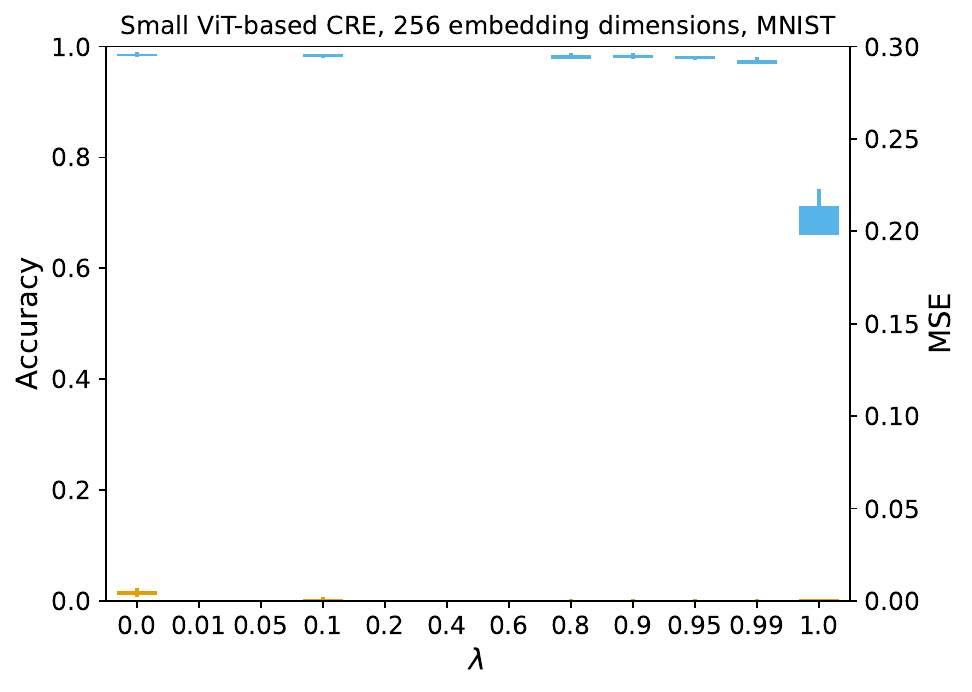}
    \end{subfigure}
    \begin{subfigure}[t]{0.246\linewidth}
        \centering
        \includegraphics[width=1.0\linewidth]{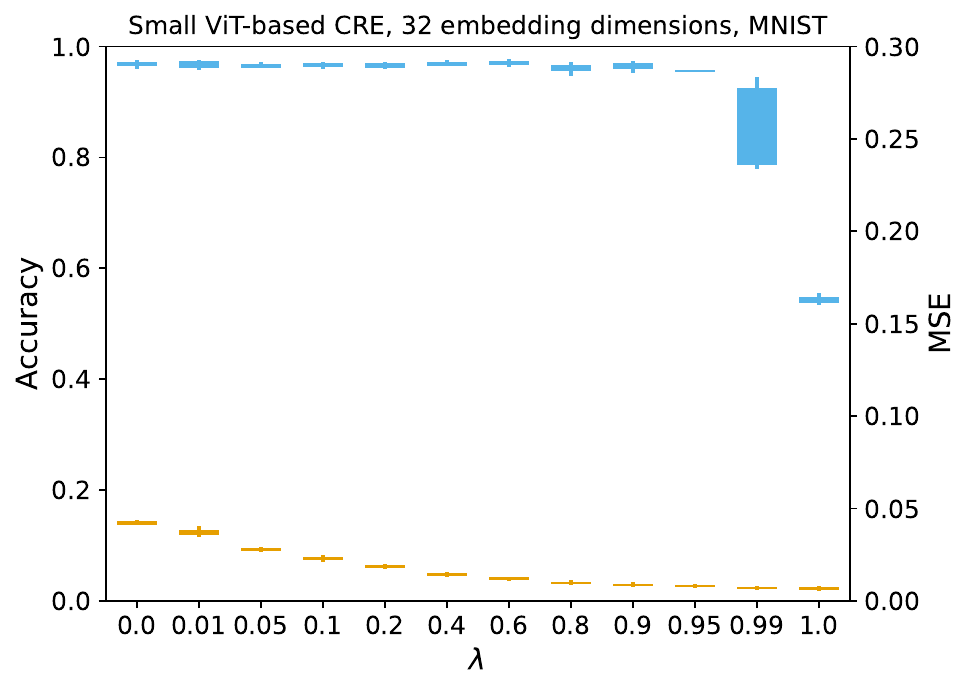}
        \includegraphics[width=1.0\linewidth]{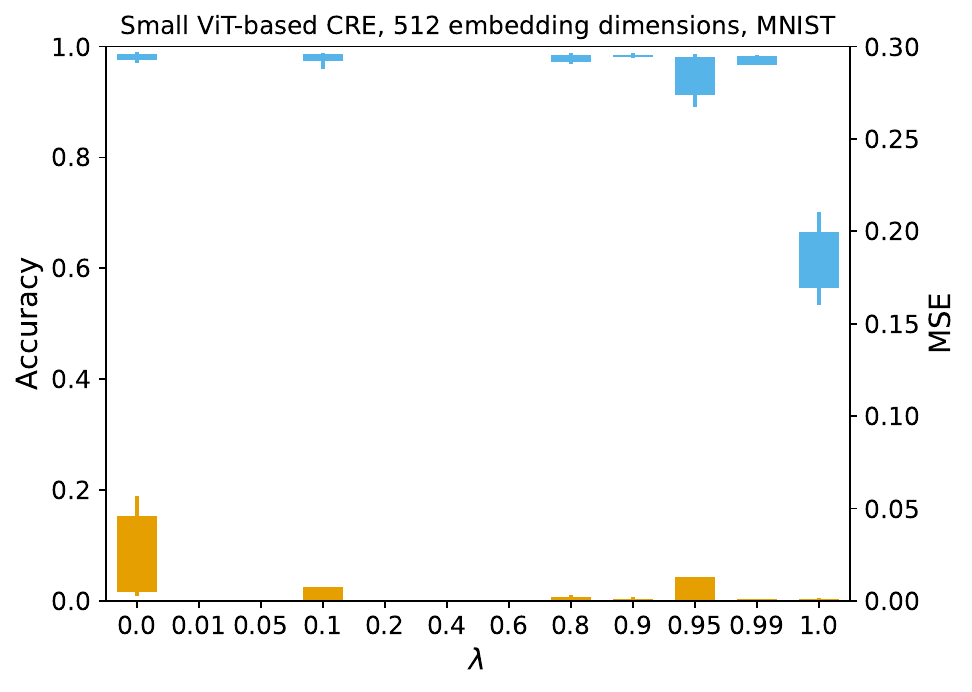}
    \end{subfigure}
    \caption{ViT-based CREs on MNIST}
    \label{fig:vit_mnist}

\end{figure}

\begin{figure}[ht]
    \begin{subfigure}[t]{0.246\linewidth}
        \centering
        \includegraphics[width=1.0\linewidth]{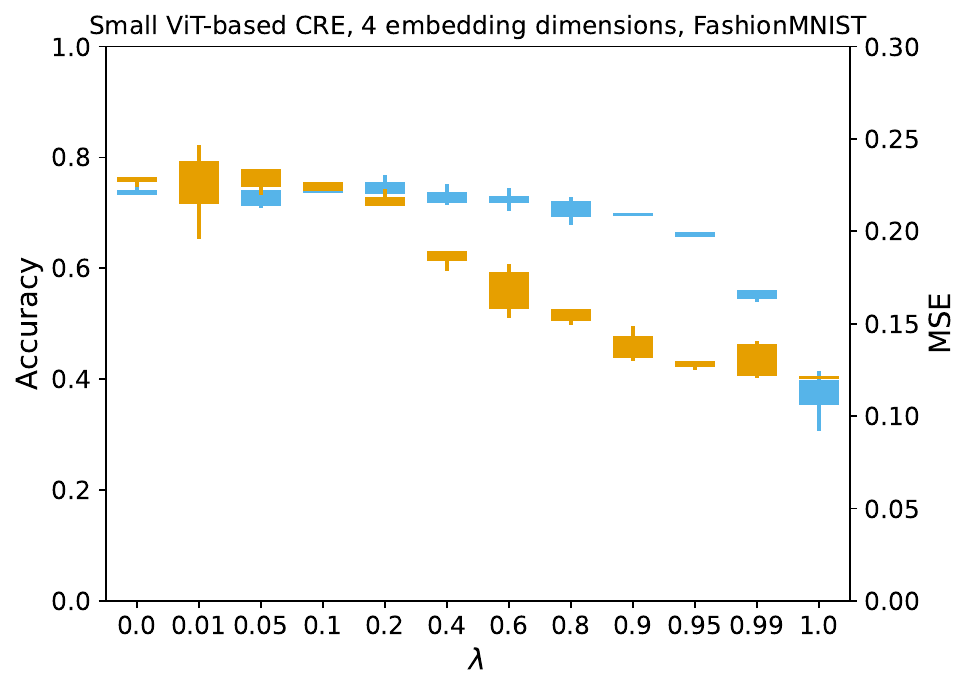}
        \includegraphics[width=1.0\linewidth]{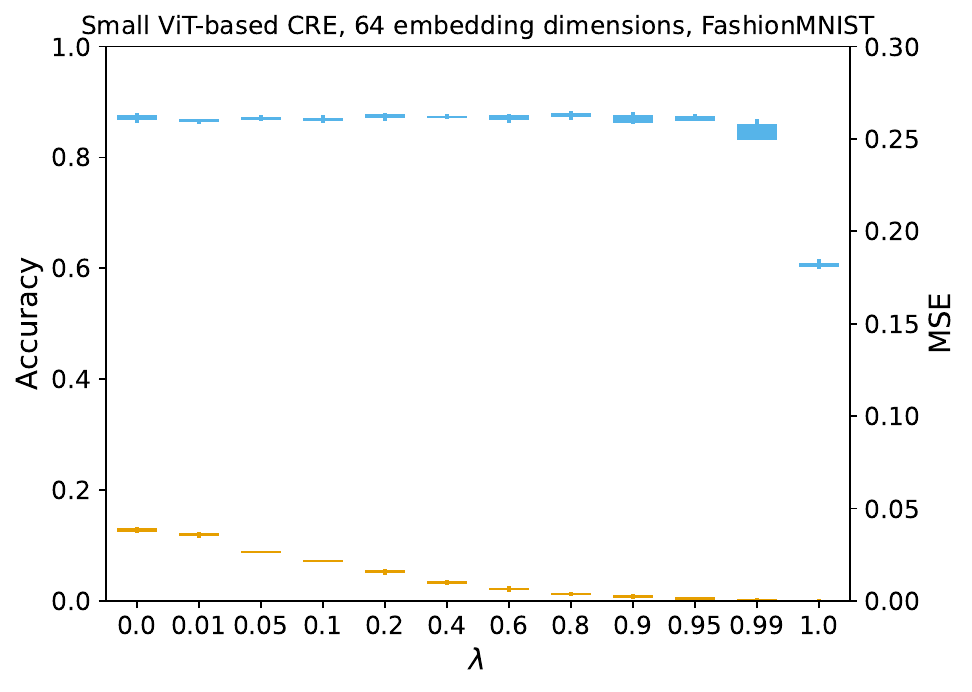}
        \includegraphics[width=1.0\linewidth]{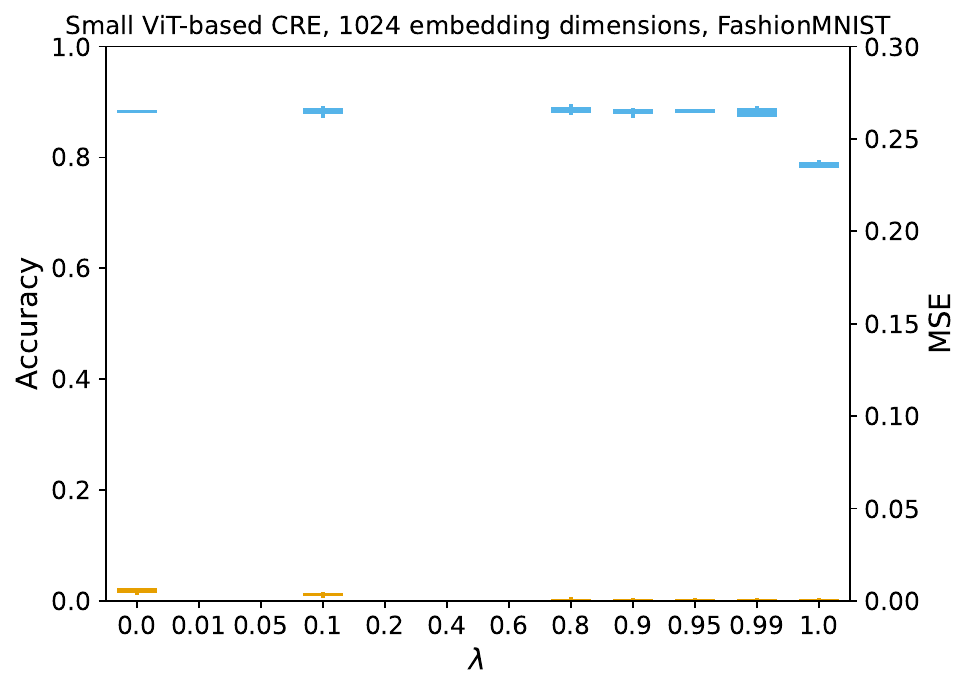}
    \end{subfigure}
    \begin{subfigure}[t]{0.246\linewidth}
        \includegraphics[width=1.0\linewidth]{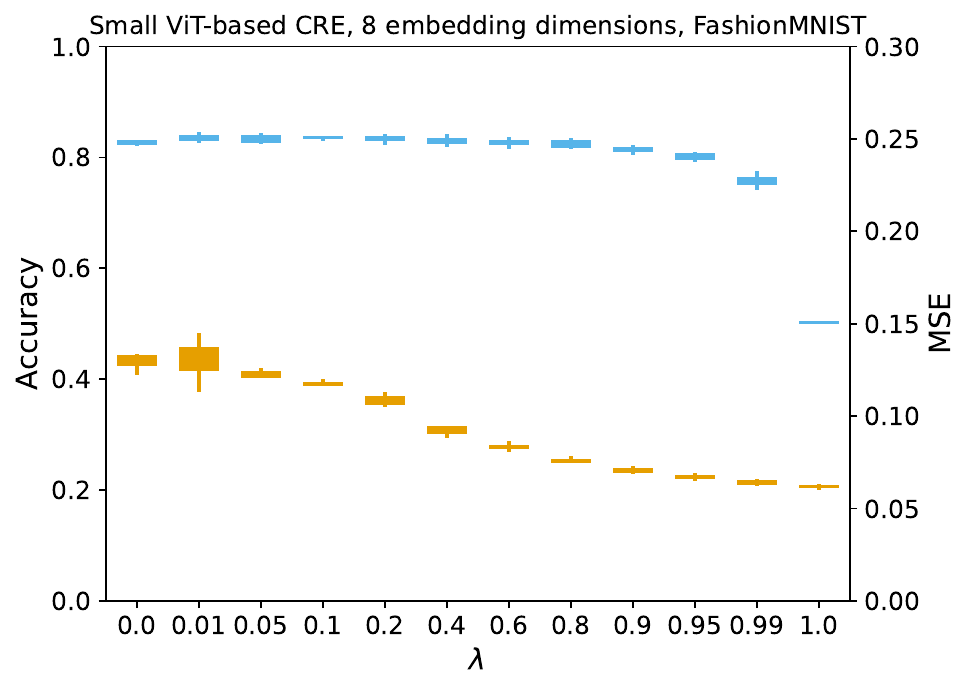}
        \includegraphics[width=1.0\linewidth]{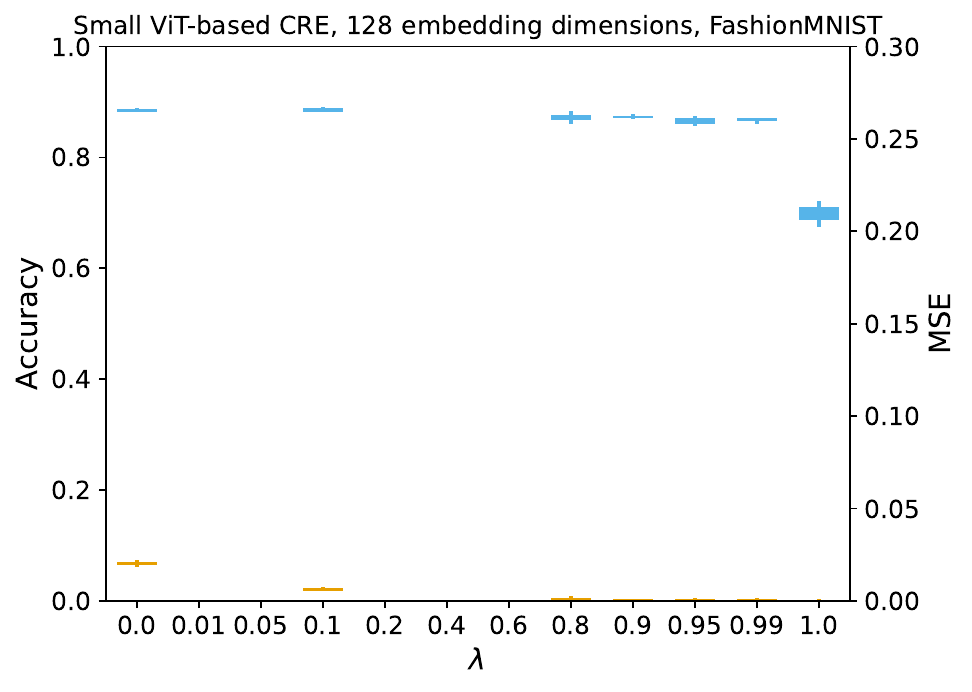}
        \includegraphics[width=1.0\linewidth]{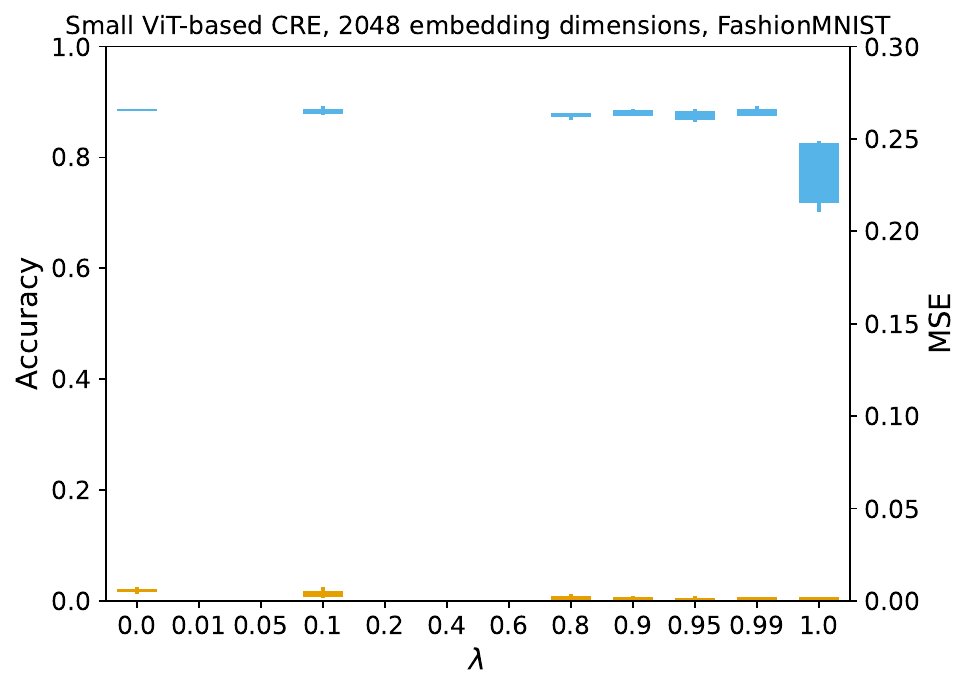}       
    \end{subfigure}
    \begin{subfigure}[t]{0.246\linewidth}
        \centering
        \includegraphics[width=1.0\linewidth]{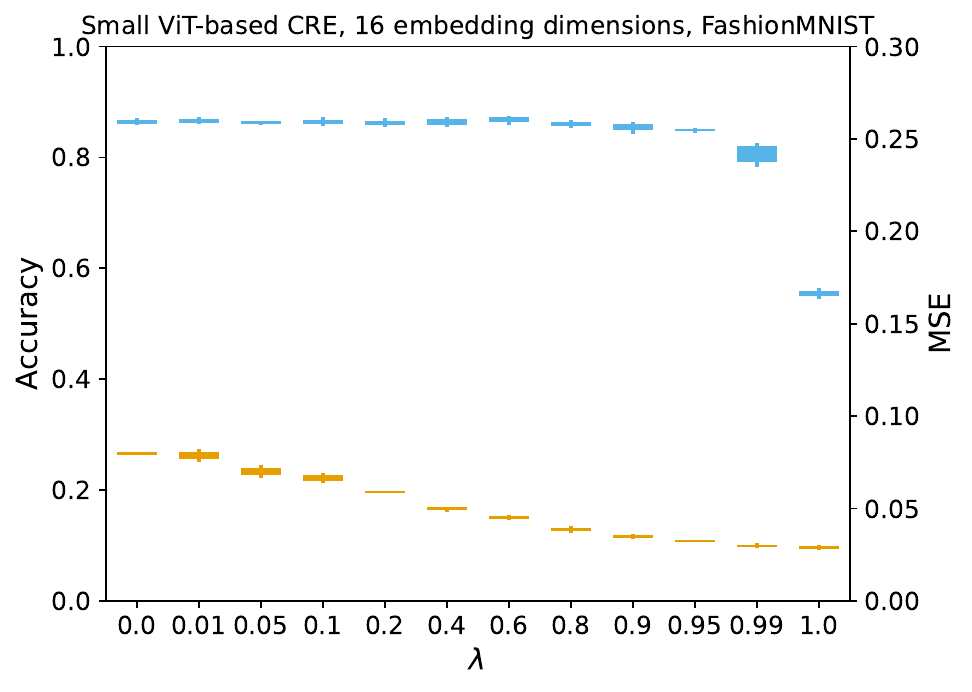}
        \includegraphics[width=1.0\linewidth]{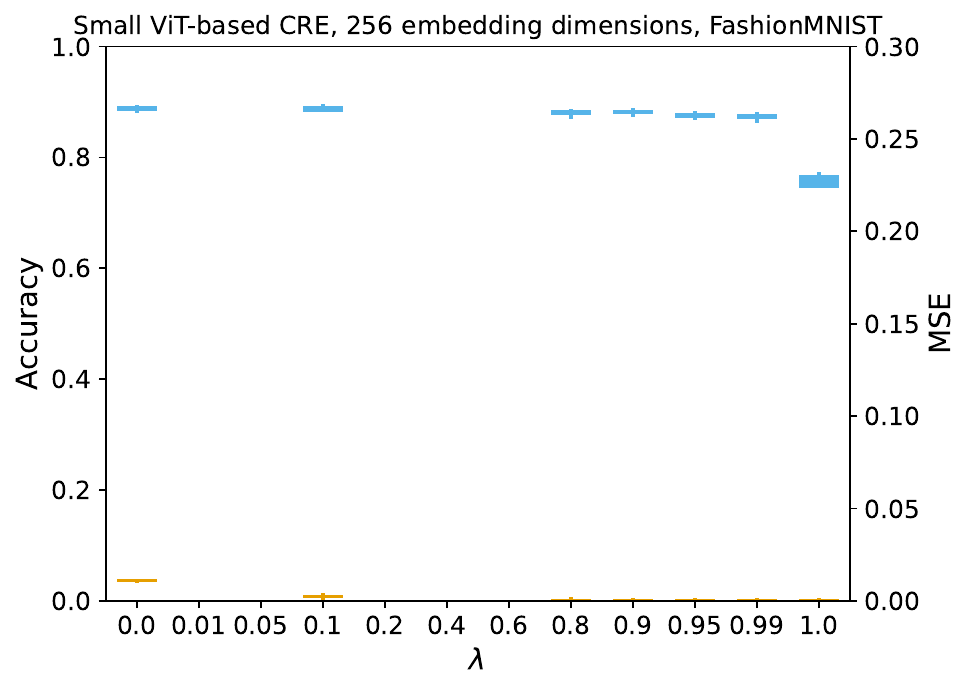}
    \end{subfigure}
    \begin{subfigure}[t]{0.246\linewidth}
        \centering
        \includegraphics[width=1.0\linewidth]{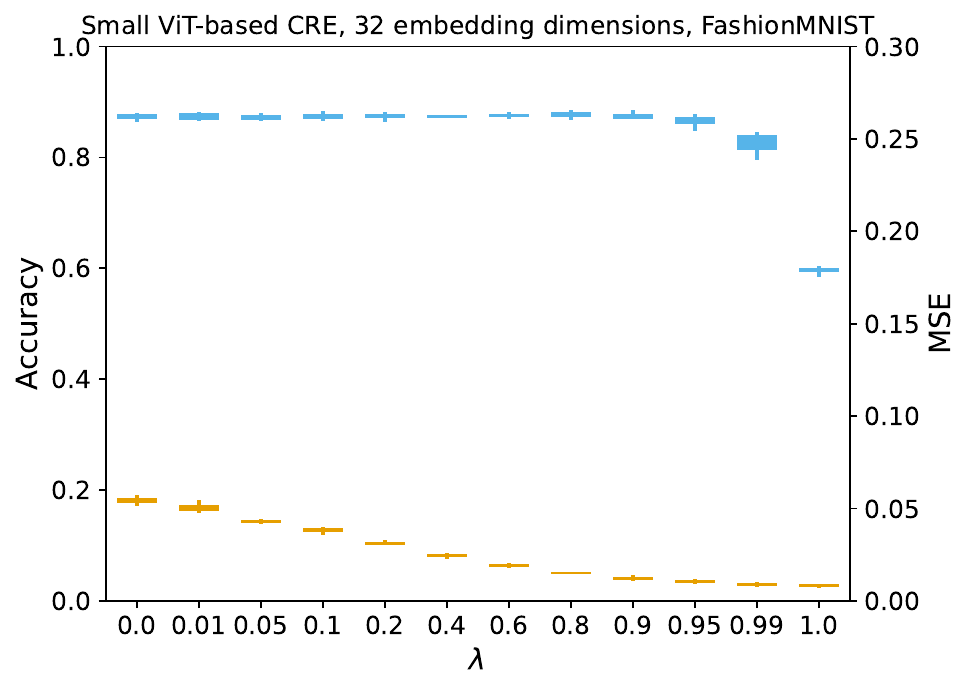}
        \includegraphics[width=1.0\linewidth]{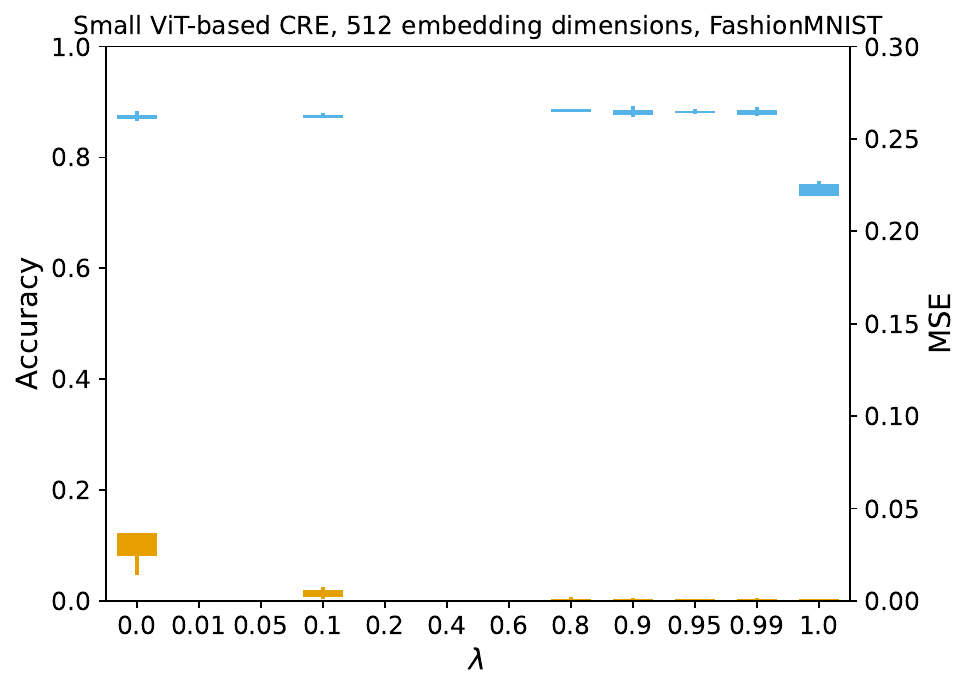}
    \end{subfigure}
    \caption{ViT-based CREs on FashionMNIST}
    \label{fig:vit_fmnist}
\end{figure}

\begin{figure}[ht]
    \begin{subfigure}[t]{0.246\linewidth}
        \centering
        \includegraphics[width=1.0\linewidth]{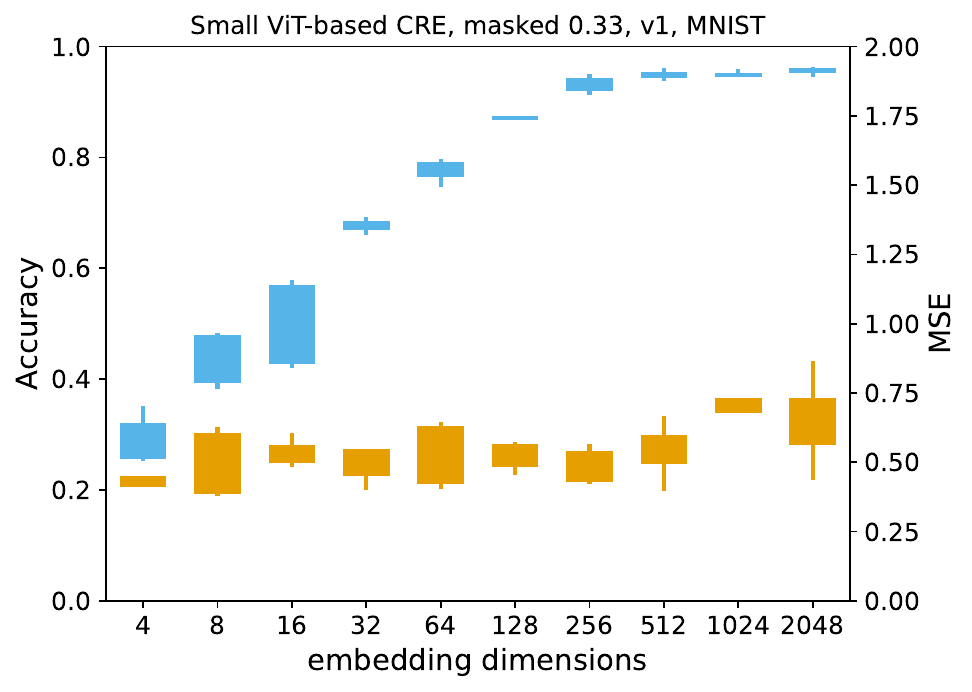}
        \includegraphics[width=1.0\linewidth]{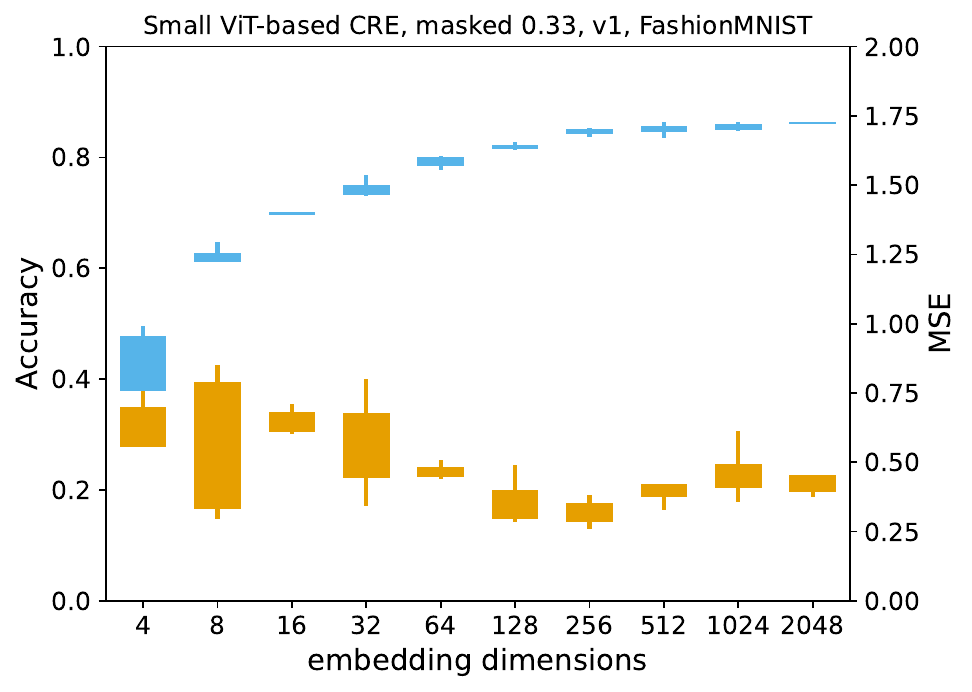}     
    \end{subfigure}
    \begin{subfigure}[t]{0.246\linewidth}
        \includegraphics[width=1.0\linewidth]{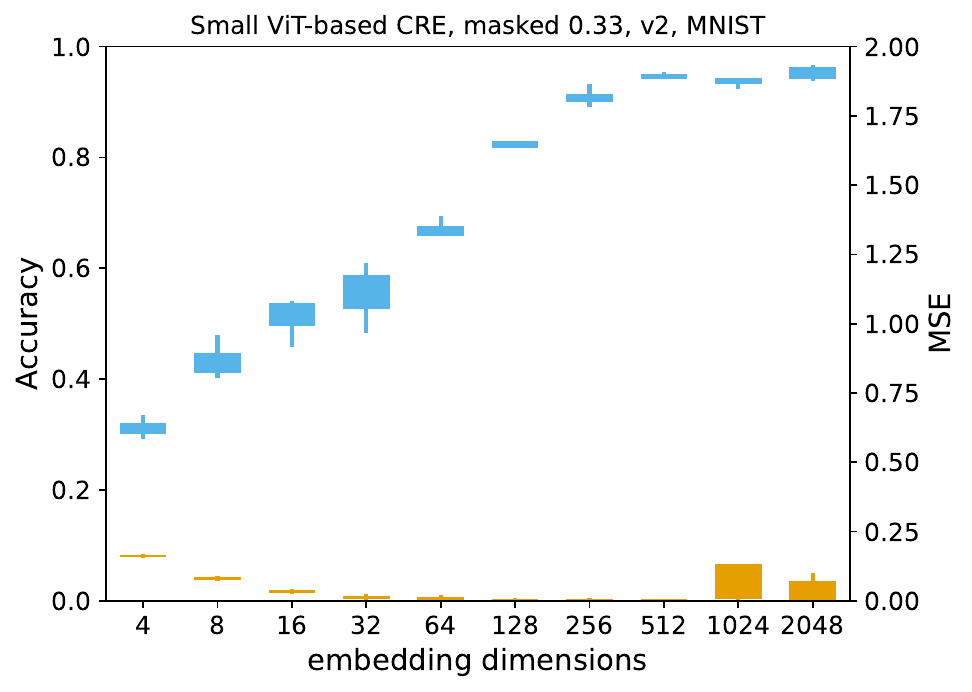}                \includegraphics[width=1.0\linewidth]{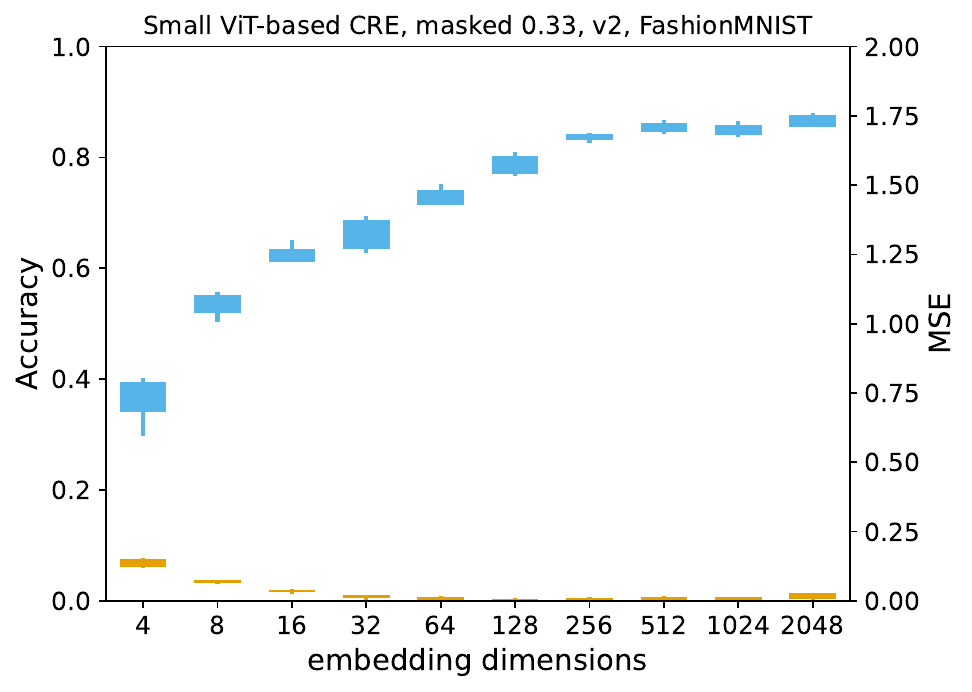}            
    \end{subfigure}
    \begin{subfigure}[t]{0.246\linewidth}
        \centering
        \includegraphics[width=1.0\linewidth]{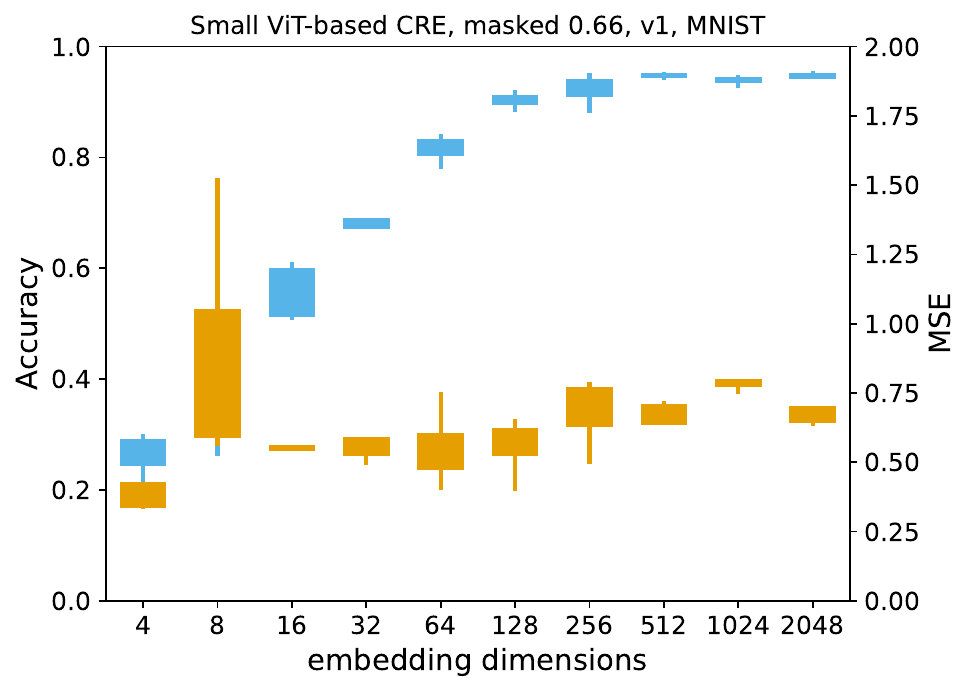}        \includegraphics[width=1.0\linewidth]{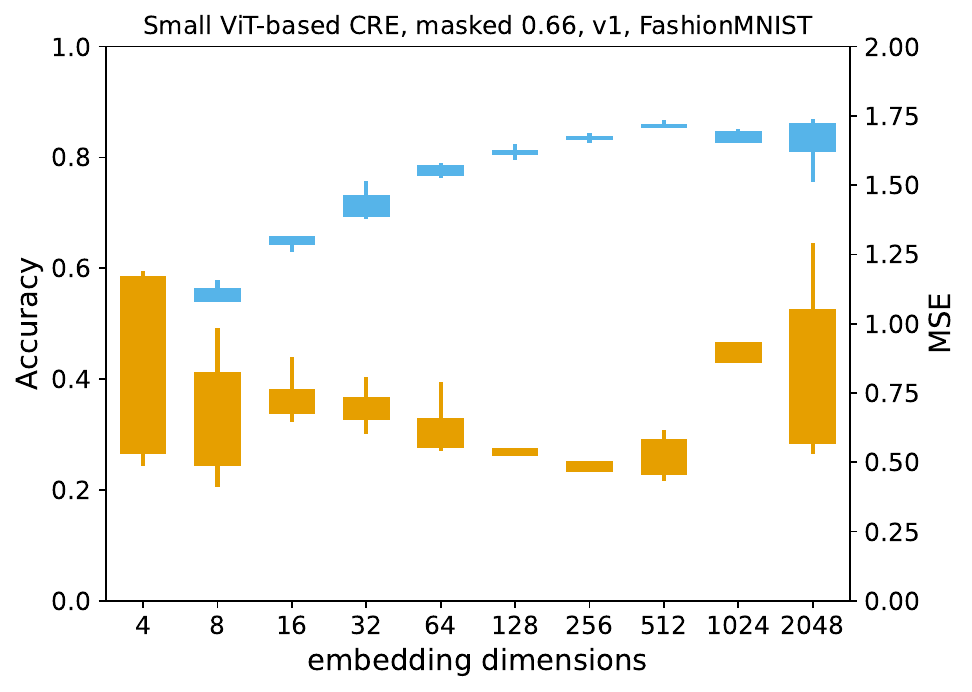}     
    \end{subfigure}
    \begin{subfigure}[t]{0.246\linewidth}
        \centering
        \includegraphics[width=1.0\linewidth]{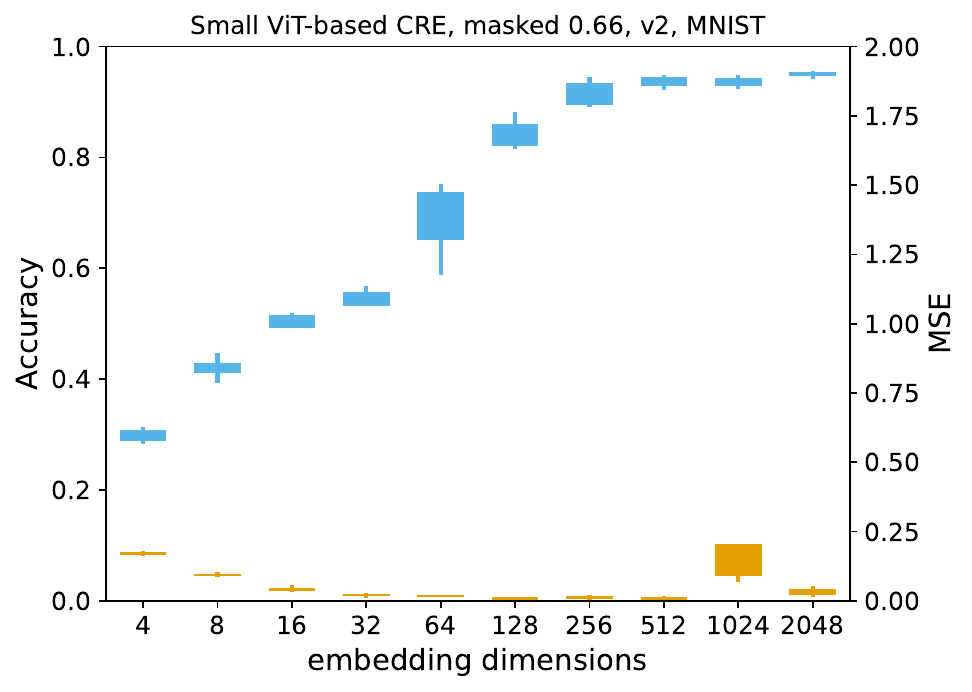}                \includegraphics[width=1.0\linewidth]{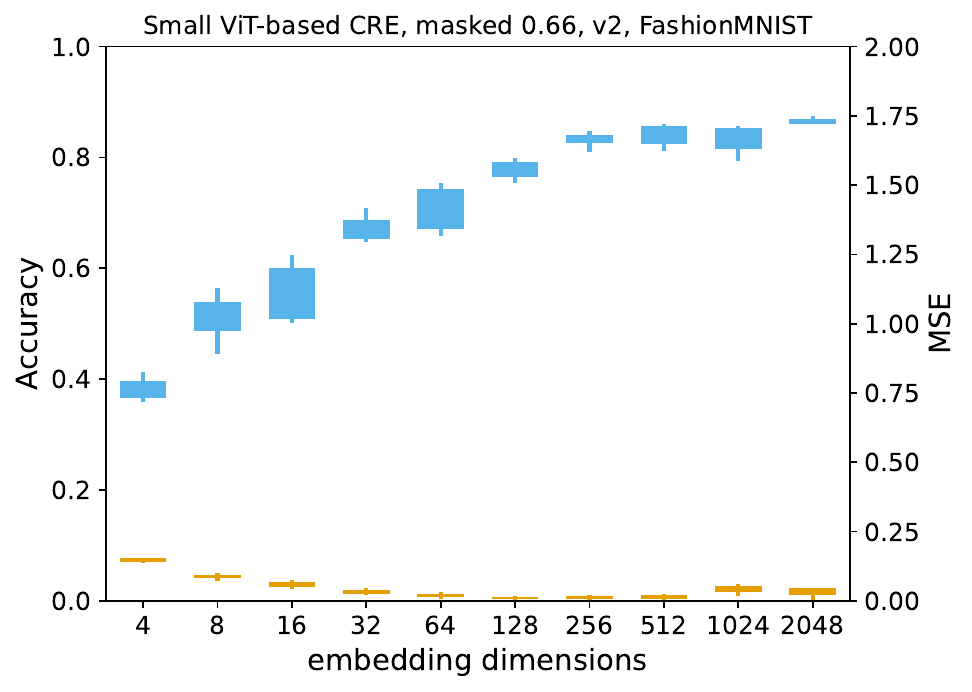}
    \end{subfigure}
    \caption{ViT-based CREs with masking on MNIST and FashionMNIST}
    \label{fig:vit_masked}
\end{figure}

\begin{figure}[ht]
    \begin{subfigure}[t]{0.246\linewidth}
        \centering
        \includegraphics[width=1.0\linewidth]{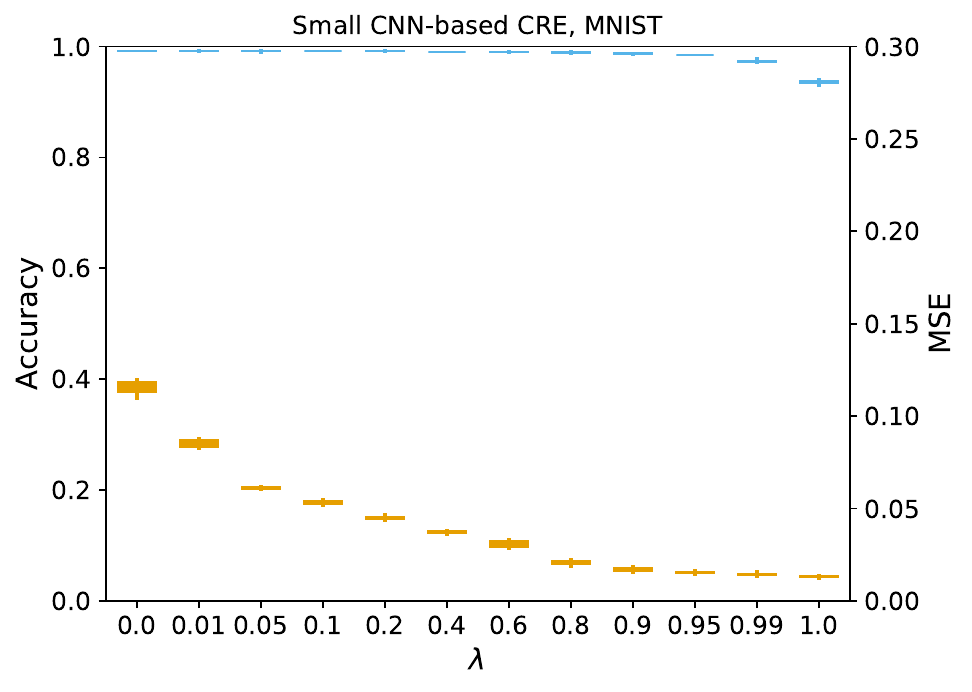}
    \end{subfigure}
    \begin{subfigure}[t]{0.246\linewidth}
        \includegraphics[width=1.0\linewidth]{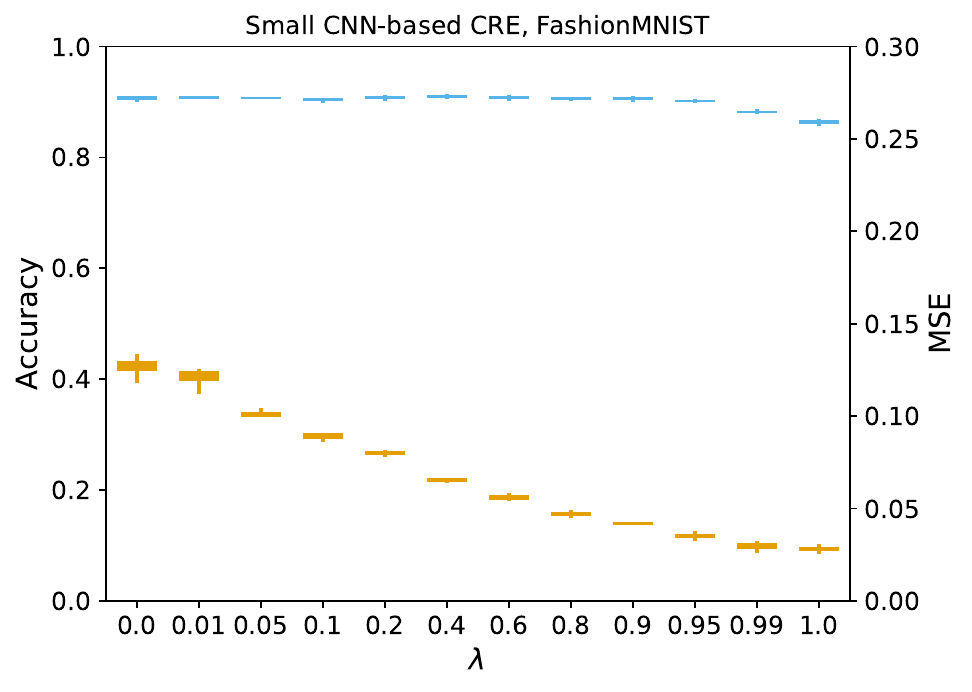}                
    \end{subfigure}
    \begin{subfigure}[t]{0.246\linewidth}
        \centering
        \includegraphics[width=1.0\linewidth]{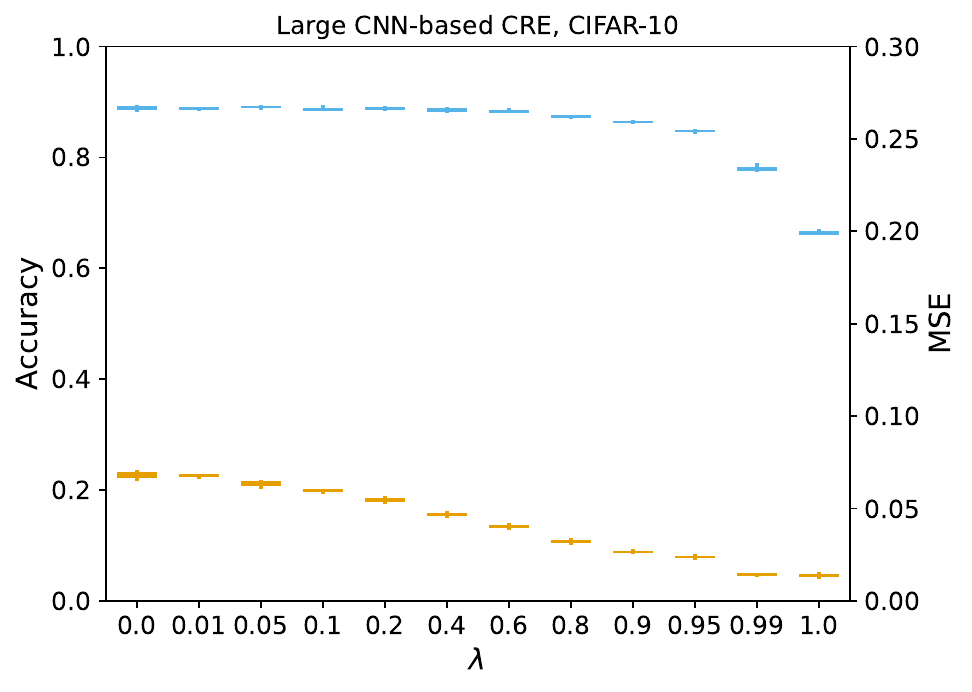}        
    \end{subfigure}
    \begin{subfigure}[t]{0.246\linewidth}
        \centering
        \includegraphics[width=1.0\linewidth]{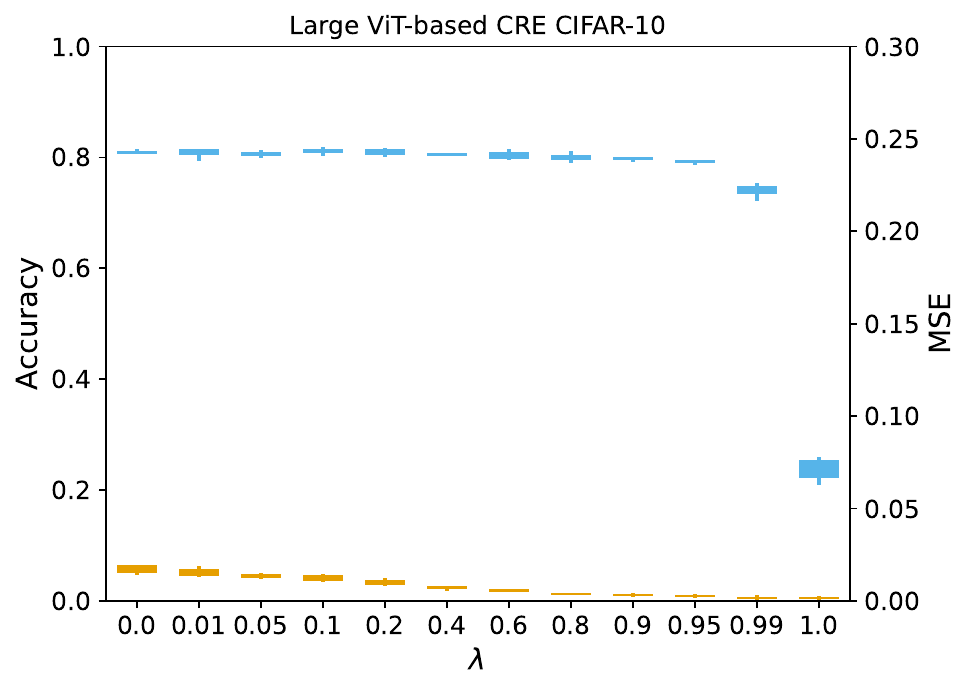}                
    \end{subfigure}
    \caption{CNN-based CREs on MNIST, FashionMNIST, CIFAR-10 and ViT-based CREs on CIFAR-10}
    \label{fig:misc}
\end{figure}

We provide the performance box plots of all CREs in Figures \ref{fig:fc_mnist}, \ref{fig:fc_fmnist}, \ref{fig:vit_mnist},  \ref{fig:vit_fmnist},  \ref{fig:vit_masked}, \ref{fig:misc}.

\subsection{Additional Analysis}
\begin{figure}[ht]
\centering
    \begin{subfigure}[t]{0.24\linewidth}
        \includegraphics[width=1.0\linewidth]{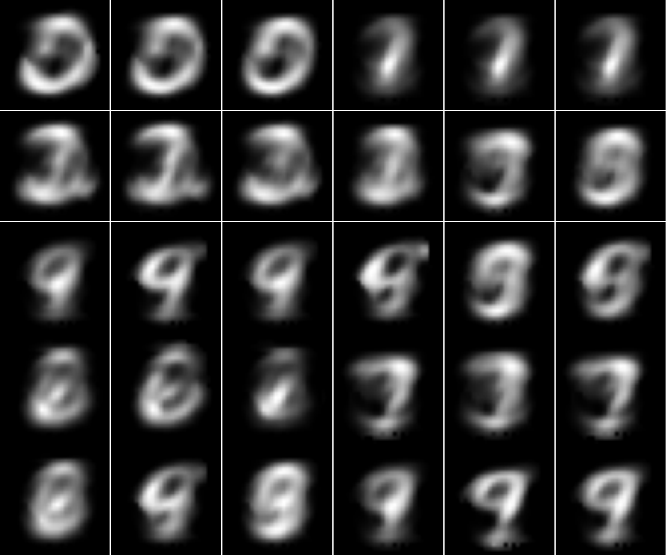}
    \end{subfigure}
    \begin{subfigure}[t]{0.24\linewidth}
        \includegraphics[width=1.0\linewidth]{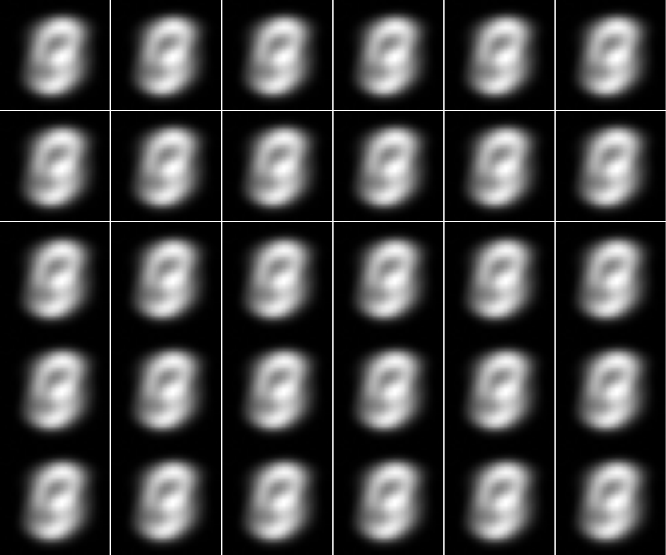}                
    \end{subfigure}
    \caption{Reconstructions on MNIST for small FC-based CREs with 6 latent dimensions. Left $\lambda=0$, right $\lambda=1$}
    \label{fig:recons_low}
\end{figure}

In our study involving small FC-CREs on the MNIST dataset, particularly those with a limited number of latent dimensions and generally low performance, we observed a notable decrease in reconstruction quality as the value of $\lambda$ approaches 1 (refer to \Cref{fig:fc_mnist}). Although these models exhibit only modest performance, the deviation from the trade-off effect, as discussed in the article's main body, presents an intriguing anomaly. This behavior can be understood by examining the reconstructions depicted in \Cref{fig:recons_low}. Both sets of reconstructions are relatively poor; at $\lambda=0$, the model appears to generate a distinct average image for each class, whereas, at $\lambda=1$, it produces a generalized average across all classes. The reconstructions suggest that class information might still offer some reconstructive value in scenarios where the architecture's resources are insufficient for the decoder to extract meaningful features. Notably, this phenomenon is unique to the MNIST dataset. We hypothesize that this unusual behavior is because the class labels in MNIST inherently carry more pixel-level information about the images than in other datasets. However, given its occurrence exclusively in low-performing models on the MNIST dataset, we treat this pattern as an outlier rather than a generalizable trend.

\end{document}